\documentclass[preprint]{elsarticle}
\usepackage{graphicx}
\usepackage{graphics}
\usepackage{subfigure}
\usepackage{amsmath}
\usepackage{amssymb}
\usepackage{multirow}
\usepackage{float}
\usepackage{bm}
\usepackage{setspace}
\usepackage{indentfirst}
\usepackage[ruled]{algorithm2e}
\usepackage{enumerate}
\usepackage{lineno}
\usepackage{amsmath}
\usepackage{blindtext}

\DeclareMathOperator*{\argmin}{arg\,min}

\newtheorem{lem}{Lemma}
\newtheorem{pro}{Proposition}
\newtheorem{thm}{Theorem}

\newenvironment{proof}{{\noindent \bf Proof.}\quad}{\hfill $\square$\par}

\begin{document}
	\begin{spacing}{2.0}
	\begin{frontmatter}
		\title{Joint Ranking SVM and Binary Relevance with Robust Low-Rank Learning for Multi-Label Classification}
	
		\author[address1,address2]{Guoqiang Wu}
		\ead{wuguoqiang16@mails.ucas.ac.cn}
		
		\author[address3]{Ruobing Zheng}
		\ead{zhengruobing@cnic.cn}
		
		\author[address2,address4,address5]{Yingjie Tian\corref{mycorrespondingauthor}}
		\cortext[mycorrespondingauthor]{Corresponding author}
		\ead{tyj@ucas.ac.cn}
		
		\author[address6]{Dalian Liu\corref{mycorrespondingauthor}}
		\ead{ldtdalian@buu.edu.cn}
		
		\address[address1]{School of Computer Science and Technology, University of Chinese Academy of Sciences, Beijing 100049, China}
		\address[address2]{Research Center on Fictitious Economy and Data Science, Chinese Academy of Sciences, Beijing 100190, China}
		\address[address3]{Computer Network Information Center, Chinese Academy of Sciences, Beijing 100190, China}
		\address[address4]{School of Economics and Management, University of Chinese Academy of Sciences, Beijing 100190, China}
		\address[address5]{Key Laboratory of Big Data Mining and Knowledge management, Chinese Academy of Sciences, Beijing 100190, China}
		\address[address6]{Department of Basic Course Teaching, Beijing Union University, Beijing 100101, China}
		
		\begin{abstract}
			Multi-label classification studies the task where each example belongs to multiple labels simultaneously. As a representative method, Ranking Support Vector Machine (Rank-SVM) aims to minimize the \emph{Ranking Loss} and can also mitigate the negative influence of the class-imbalance issue. However, due to its stacking-style way for thresholding, it may suffer \emph{error accumulation} and thus reduces the final classification performance. Binary Relevance (BR) is another typical method, which aims to minimize the \emph{Hamming Loss} and only needs one-step learning. Nevertheless, it might have the class-imbalance issue and doesn't take into account label correlations. To address the above issues, we propose a novel multi-label classification model, which joints Ranking support vector machine and Binary Relevance with robust Low-rank learning (RBRL). RBRL inherits the ranking loss minimization advantages of Rank-SVM, and thus overcomes the disadvantages of BR suffering the class-imbalance issue and ignoring the label correlations. Meanwhile, it utilizes the hamming loss minimization and one-step learning advantages of BR, and thus tackles the disadvantages of Rank-SVM including another thresholding learning step. Besides, a low-rank constraint is utilized to further exploit high-order label correlations under the assumption of low dimensional label space. Furthermore, to achieve nonlinear multi-label classifiers, we derive the kernelization RBRL. Two accelerated proximal gradient methods (APG) are used to solve the optimization problems efficiently. Extensive comparative experiments with several state-of-the-art methods illustrate a highly competitive or superior performance of our method RBRL.
		\end{abstract}
	
	
	\end{frontmatter}
	\section{Introduction}
		Traditional supervised single-label classification handles the task where each example is assigned to one class label. However, in many real-world classification applications, an example is often associated with a set of class labels. For instance, in text categorization, a document may belong to many labels such as ``religion" and ``politics". This brings the hot research interests of multi-label classification (MLC), which investigates the task where each example may be assigned to multiple class labels simultaneously. So far, MLC has witnessed its success in a wide range of research fields, such as function genomics \cite{clare2001knowledge,elisseeff2001kernel,zhang2006multilabel}, multimedia contents annotation \cite{boutell2004learning,qi2007correlative,trohidis2008multi}, and NLP (e.g., text categorization \cite{mccallum1999multi, schapire2000boostexter,ueda2003parametric}, and information retrieval \cite{yu2005multi,zhu2005multi, gopal2010multilabel}).
		
		As a representative method for MLC, Ranking Support Vector Machine (Rank-SVM) \cite{elisseeff2001kernel} aims to minimize the empirical \emph{Ranking Loss} while having a large margin and is enabled to cope with nonlinear cases with the kernel trick \cite{Kivinen2002Learning}. The class-imbalance issue usually occurs in MLC, which mainly includes two aspects \cite{zhang2015towards, xing2018multi}. On one hand, for a specific class label, the number of positive instances is greatly less than that of negative instances. On the other hand, for a specific instance, the number of relevant labels is usually less than that of irrelevant labels. Generally, the pairwise loss, which can be used to optimize imbalance-specific evaluation metrics such as the area under the ROC curve (AUC) and F-measure \cite{cortes2004auc, wu2017unified}, is more able to deal with the class-imbalance issue than the pointwise loss. Thus, Rank-SVM can tackle the second aspect of the class-imbalance issue in MLC by the minimization of the pairwise approximate ranking loss. Therefore, it can mitigate the negative influence of the class-imbalance issue in MLC. Nevertheless, apart from the first ranking learning step, it needs another thresholding learning step, which is a stacking-style way to set the thresholding function. Inevitably, each step has the estimation error. As a result, it may cause \emph{error accumulation} and eventually reduce the final classification performance for MLC. Therefore, it's better to find a way to train the model in only one step. Although there are some methods proposed to tackle this issue, such as calibrated Rank-SVM \cite{jiang2008calibrated} and Rank-SVMz \cite{xu2012efficient}, the basic idea of these methods is to introduce a virtual zero label for thresholding, which increases the number of the 	hypothesis parameter variables to raise the complexity of the model (i.e., the hypothesis set). Besides, while calibrated Rank-SVM \cite{jiang2008calibrated} makes the optimization problem more computationally complex, Rank-SVMz \cite{xu2012efficient} makes it train more efficiently. Moreover, there is little work to combine with Binary Relevance to address this issue.
		
		Binary Relevance (BR) \cite{boutell2004learning} is another typical method, which transforms the MLC task into many independent binary classification problems. It aims to optimize the \emph{Hamming Loss} and only needs one-step learning. Despite the intuitiveness of BR, it might have the class-imbalance issue, especially when the label cardinality (i.e. the average number of labels per example) is low and the label space is large. Besides, it doesn't take into account label correlations, which plays an important role to boost the performance for MLC. Recently, to facilitate the performance of BR, many regularization-based approaches \cite{xu2014learning, yu2014large, jing2015semi, xu2016local} impose a low-rank constraint on the parameter matrix to exploit the label correlations. However, little work has been done to consider the minimization of the \emph{Ranking Loss} to mitigate the negative influence of the class-imbalance issue and exploit the label correlations simultaneously. Moreover, these low-rank approaches are mostly linear models, which can't capture complex nonlinear relationships between the input and output.
		
		To address the above issues, in this paper we propose a novel multi-label classification model, which joints Ranking support vector machine and Binary Relevance with robust Low-rank learning (RBRL). Specifically, we incorporate the thresholding step into the ranking learning step of Rank-SVM via Binary Relevance, which makes it train the model in only one step. It can also be viewed as an extension of BR, which aims to additionally consider the minimization of the \emph{Ranking Loss} to boost the performance. Hence, it can enjoy the advantages of Rank-SVM and BR, and tackle the disadvantages of both. Besides, the low-rank constraint on the parameter matrix is employed to further exploit the label correlations. Moreover, to achieve nonlinear multi-label classifier, we derive the kernelization of the linear RBRL. What's more, to solve the objective functions for the linear and kernel RBRL efficiently, we use the accelerated proximal gradient methods (APG) with a fast convergence rate $\mathcal{O}(1/t^2)$, where $t$ is the number of iterations. 
		
		The contributions of this work are mainly summarized as follows:
		\begin{enumerate}[(1)]
			\item We present a novel multi-label classification model, which joints Rank-SVM and BR with robust low-rank learning.
			\item Different from existing low-rank approaches which are mostly linear models, we derive the kernelization RBRL to capture nonlinear relationships between the input and output.
			\item For the linear and kernel RBRL, we use two accelerated proximal gradient methods (APG) to efficiently solve the optimization problems with fast convergence. 
			\item Extensive experiments have confirmed the effectiveness of our approach RBRL over several state-of-the-art methods for MLC.
		\end{enumerate}
		
		The rest of this paper is organized as follows. In Section \ref{sec:related_work}, the related work about MLC is mainly reviewed. In Section \ref{sec:model}, the problem of formulation and the model RBRL are presented in detail. The corresponding optimization algorithms are proposed in Section \ref{sec:optimization}. In Section \ref{sec:experiments}, experimental results are presented. Finally, Section \ref{sec:conclusion} concludes this paper.
	
	\section{Related Work}
	\label{sec:related_work}
		For multi-label classification (MLC), each example may belong to multiple class labels simultaneously. During the past decade, a variety of approaches have been presented to deal with multi-label data in diverse domains. According to the popular taxonomy proposed in \cite{tsoumakas2006multi,tsoumakas2009mining}, the MLC approaches can be roughly grouped into two categories: algorithm adaption approaches and problem transformation approaches.
		
		Algorithm adaption approaches aim to modify traditional single-label classification algorithms to solve the MLC task directly. All single-label classification algorithms almost have been adapted to the MLC task. Rank-SVM \cite{elisseeff2001kernel} adapts a maximum margin strategy to MLC, which aims to minimize the \emph{Ranking Loss} while having a large margin and copes with nonlinear cases via the kernel trick. Multi-label Decision Tree (ML-DT) \cite{clare2001knowledge} derives from decision tree to solve the MLC task. CML \cite{ghamrawi2005collective} adapts the maximum entropy principle to deal with the MLC task. BP-MLL \cite{zhang2006multilabel} is an adaptation of neural networks for MLC. ML-kNN \cite{zhang2007ml} derives from the lazy learning technique kNN classifier. CNN-RNN \cite{wang2016cnn} adapts deep convolutional neural networks (CNNs) and recurrent neural networks (RNNs) for multi-label image classification. ML-FOREST \cite{wu2016mlforest} is an adaption of the tree ensemble method for MLC. $\text{ML}^2$ \cite{hou2016multi} adapts manifold learning to construct and exploit the label manifold for MLC. Linear discriminant analysis (LDA) is adapted by wMLDA \cite{xu2018weighted} for multi-label feature extraction.
		
		Problem transformation approaches aim to convert the MLC task to other well-established learning problems such as single-label classification and label ranking. BR \cite{boutell2004learning} transforms MLC into many independent binary classification problems. Calibrated Label Ranking (CLR) \cite{furnkranz2008multilabel} converts MLC task into the task of label ranking. Classifier Chain (CC) \cite{read2011classifier} converts MLC task into a chain of binary classification problems, which encodes the label correlations into feature representation and builds the classifiers based on a chaining order specified over the labels. Nevertheless, owing to the difficulty of the chain order determination, the ensemble learning \cite{zhou2012ensemble} technique can be used to construct the approach Ensemble of Classifier Chains (ECC) \cite{read2011classifier}. Label Powerset (LP) converts MLC task into a multi-class classification problem which views the subsets of label space as new classes. In view of the huge class space issue that LP may suffer, Random k-Labelsets (RAKEL) \cite{tsoumakas2011random} proposes to combine ensemble learning with LP to boost the performance.
		
		For MLC, labels may have inter-correlations between each other. As is well accepted, exploitation of label correlations plays an important role to boost the performance of MLC. Based on the order of label correlations that approaches consider \cite{zhang2014review}, the existing MLC approaches can be roughly categorized into three families: first-order, second-order and high-order. First-order approaches don't take into account the label correlations, of which the typical approaches mainly include ML-DT \cite{clare2001knowledge}, BR \cite{boutell2004learning}, and ML-kNN \cite{zhang2007ml}. Second-order approaches consider the correlations between label pairs, of which the typical approaches mainly include Rank-SVM \cite{elisseeff2001kernel}, CLR \cite{furnkranz2008multilabel}, BP-MLL \cite{zhang2006multilabel}, and CML \cite{ghamrawi2005collective}. High-order approaches take into account label correlations higher than second-order, of which the typical approaches mainly include RAKEL \cite{tsoumakas2011random} and ECC \cite{read2011classifier}. It may boost the performance more when an approach takes into account higher-order label correlations. Nevertheless, it would lead to greater computational cost and less scalability.
		
		Although Rank-SVM \cite{elisseeff2001kernel} has a good ability to minimize the \emph{Ranking Loss}, it may suffer the \emph{error accumulation} issue because of its stacking-style way to set the thresholding function, which also occurs in other ranking-based methods, such as BP-MLL \cite{zhang2006multilabel}. There are some approaches proposed to tackle this issue, such as calibrated Rank-SVM \cite{jiang2008calibrated} and Rank-SVMz \cite{xu2012efficient}. The core idea of these approaches is to introduce a virtual zero label for thresholding and train the model in one step. However, little work has been done to combine with Binary Relevance to address this issue. Besides, the mostly used Frank–-Wolfe method \cite{frank1956algorithm, jaggi2013revisiting} to solve the optimization problems has a convergence rate $\mathcal{O}(1/t)$, which is not efficient.
		
		Recently, some regularization-based approaches are proposed for MLC. To exploit the label correlations, many approaches \cite{xu2014learning, yu2014large, jing2015semi, xu2016local} often impose a low-rank constraint on the parameter matrix. Besides, the manifold regularization is employed by many approaches \cite{wang2009image, huang2015learning, huang2018joint, zhu2018multi} to exploit the label correlations. However, little work has been done to take into account the minimization of the \emph{Ranking Loss} to boost the performance. In addition, these methods are mostly linear models, which can't capture nonlinear relationships between the input and output.
		
		Besides, there are some other approaches for MLC. A strategy, called \emph{cross-coupling aggregation}, is proposed by COCOA \cite{zhang2015towards} to mitigate the class-imbalance issue and exploit the label correlations simultaneously. The label-specific features are employed by LIFT \cite{zhang2015lift} and JFSC \cite{huang2018joint} to boost the performance of MLC. The structural information of the feature space is employed by MLFE \cite{zhang2018feature} to enrich the labeling information. CPNL \cite{wu2018cost} presents a cost-sensitive loss to address the class-imbalance issue and boosts the performance by exploitations of negative and positive label correlations.
		
	\section{Problem Formulation}
	\label{sec:model}
		In this section, we first provide the basic linear Rank-SVM model. Then, we incorporate the thresholding step into the ranking learning problem via Binary Relevance. Next, the low-rank constraint is utilized to further exploit the label correlations. Afterward, we present the kernelization of the linear model. Finally, we compare the proposed method RBRL with other variant methods of Rank-SVM.
	
	\subsection{Preliminary}
		For a matrix $\mathbf{A}$, $\mathbf{A}^\top$ is its transpose, $\mathbf{a}_i$ and $\mathbf{a}^j$ are the $i$th row and $j$th column of $\mathbf{A}$, $\| \mathbf{a}_i\|$ (or $\| \mathbf{a}_i\|_2$) is the vector $l_2$-norm, $\| \mathbf{A} \|_1$ is the matrix $l_1$-norm and $\|\mathbf{A}\|_F$ (or $\|\mathbf{A}\|$) denotes the Frobenius norm. $Tr(\cdot)$ is the trace operator for a matrix. $Rank(\cdot)$ denotes the rank of a matrix. $\| \mathbf{A} \|_* = Tr(\sqrt{\mathbf{A}^\top \mathbf{A}}) = \sum_i \sigma_i(\mathbf{A})$ is the trace norm (or nuclear norm), where $\sigma_i(\mathbf{A})$ is the $i$th largest singular value of $\mathbf{A}$. For two matrices $\mathbf{A}$ and $\mathbf{B}$, $\mathbf{A} \circ \mathbf{B}$ denotes the Hadamard (element-wise) product. $[\![\pi ]\!]$ equals 1 when the proposition $\pi$ holds, and 0 otherwise. Denote a function $g: \mathbb{R} \rightarrow \mathbb{R}$, $\forall \ \mathbf{A} \in \mathbb{R}^{n \times m}$, define $g(\mathbf{A}): \mathbb{R}^{n \times m} \rightarrow \mathbb{R}^{n \times m}$, where $(g(\mathbf{A}))_{ij} = g(\mathbf{A}_{ij})$. 
		
		Given a training set $\mathcal{D} = (\mathbf{X}, \mathbf{Y})$, where $\mathbf{X} = [\mathbf{x}_1;...; \mathbf{x}_n] \in \mathbb{R}^{n \times m}$ is the input matrix, $\mathbf{Y} = [\mathbf{y}_1;...;\mathbf{y}_n] \in {\{-1, 1\}}^{n \times l}$ is the label matrix, $\mathbf{x}_i \in \mathbb{R}^m$ is an $m$-dimensional real-valued instance vector, $\mathbf{y}_i \in \{-1, 1\}^l$ is the label vector of $\mathbf{x}_i$, $n$ is the number of the training samples, and $l$ is the number of potential labels. Besides, $y_{ij} = 1$ (or $-1$) indicates the $j$th label of the $i$th instance is relevant (or irrelevant). The goal of MLC is to learn a multi-label classifier $H: \mathbb{R}^m \rightarrow {\{-1, 1\}}^l$.
	
	\subsection{Rank-SVM}
		We begin with the basic linear Rank-SVM \cite{elisseeff2001kernel} for MLC. For an instance $\mathbf{x} \in \mathbb{R}^{m}$, its real-valued prediction is obtained by $\mathbf{f} = \mathbf{x} \mathbf{W} + \mathbf{b}$, where  $\mathbf{b} = [b_1, b_2, ..., b_l] \in \mathbb{R}^l$ is the bias and $\mathbf{W} = [ \mathbf{w}^1,\mathbf{w}^2,...,\mathbf{w}^l ] \in \mathbb{R}^{m \times l}$ is the parameter matrix. For simplicity, $b_j$ can be absorbed into $\mathbf{w}^j$ when appending $1$ to each instance $\mathbf{x}$ as an additional feature. Rank-SVM aims to minimize the \emph{Ranking Loss} while having a large margin, where the ranking learning step can be formulated as follows.
		\begin{equation}
		\label{equation:model1}
		\begin{aligned}
		& \min \limits_\mathbf{W} \ \frac{1}{2} \sum_{j=1}^{l} \| \mathbf{w}^j \|^2 + \lambda_2 \sum_{i=1}^{n} \frac{1}{|Y_i^+||Y_i^-|} \sum_{p \in Y_i^+} \sum_{q \in Y_i^-} \xi_{pq}^i \\
		& s.t. \ \langle \mathbf{w}^p, \mathbf{x}_i \rangle - \langle \mathbf{w}^q, \mathbf{x}_i \rangle \geq 1 - \xi_{pq}^i, \ (p, q) \in Y_i^{+}  \times Y_i^{-} \\
		& \qquad \xi_{pq}^i \geq 0, \ i = 1,...,n \\ 
		\end{aligned}
		\end{equation}
		where $Y_i^{+}$ (or $Y_i^{-}$) denotes the index set of relevant (or irrelevant) labels associated with the instance $\mathbf{x}_i$, $| \ |$ denotes the cardinality of a set, and $\lambda_2$ is a tradeoff hyper-parameter which controls the model complexity. Note that it will additionally regularize the bias term $b_j$ when absorbing $b_j$ into $\mathbf{w}^j$, which is different from the original optimization problem that doesn't regularize $b_j$. However, regularizing the bias usually does not make a significant difference to the sample complexity \cite{shalev2014understanding}. Besides, it has good performance in practice \cite{huang2018joint, wu2018unified, wu2018cost}. 
		
		Besides, Rank-SVM needs another thresholding learning step. First, based on the learned $\mathbf{f}_i = [f_{i1}, f_{i2},...,f_{il}] \in \mathbb{R}^{l}$ for each instance $\mathbf{x}_i$ in the training set, it finds the ideal thresholding value $t(\mathbf{x}_i)$ according to the following rule. 
		\begin{equation}
		t(\mathbf{x}_i) = \argmin_t \sum_{j=1}^{l} \Big \{ [\![ j \in Y_i^{+} ]\!] [\![ f_{ij} \leq t ]\!] + [\![ j \in Y_i^{-} ]\!] [\![ f_{ij} \geq t ]\!] \Big \}
		\end{equation}
		When the obtained threshold is not unique and the optimal values are a segment, Rank-SVM chooses the middle of this segment. Then, the \emph{threshold based} method \cite{elisseeff2001kernel} can be formalized as a regression problem $T: \mathbb{R}^l \rightarrow \mathbb{R}$, where a linear least square method is used.
		
		When in prediction for each test instance $\mathbf{x}$, we first achieve $\mathbf{f} =  \mathbf{x} \mathbf{W}$, then get the threshold value $t(\mathbf{x}) = T(\mathbf{f})$ and finally obtain the multi-label classifier result $h(\mathbf{x}) = sign([\![ \mathbf{f} > t(\mathbf{x}) ]\!])$, where $sign(x)$ returns $1$ when $x > 0$ and $-1$ otherwise. Notably, the second thresholding step has an implicit presumption that when in training and test, its input has the same data distribution, especially for the label ranking order \cite{jiang2008calibrated}. It's obvious that this assumption probably doesn't hold. Specifically, when in prediction, the output of the first ranking step naturally has the test error bias relative to the training error, especially for the ranking loss. Besides, in the following thresholding step, the regression model inevitably has the test error and its input also has the bias because of the first step. Thus, the obtained thresholding value would have a large error.  Since the first step has the ranking loss error bias and the thresholding value also has the error bias, the final multi-label classifier would suffer the \emph{error accumulation} issue.
	
	\subsection{Thresholding via Binary Relevance}
		To tackle the above issue, we aim to incorporate the thresholding step into the ranking learning optimization problem, which can be formulated as follows.
		\begin{equation}
		\label{equation:model2_1}
		\begin{split}
		& \min \limits_\mathbf{W} \ \sum_{i=1}^{n} \sum_{j=1}^{l} \Big \{ [\![ j \in Y_i^{+} ]\!] [\![ \langle \mathbf{w}^j, \mathbf{x}_i \rangle \leq t(\mathbf{x}_i) ]\!] + [\![ j \in Y_i^{-} ]\!] [\![ \langle \mathbf{w}^j, \mathbf{x}_i \rangle \geq t(\mathbf{x}_i) ]\!] \Big \} \\ 
		& \qquad + \frac{\lambda_1}{2} \sum_{j=1}^{l} \| \mathbf{w}^j \|^2 + \lambda_2 \sum_{i=1}^{n} \frac{1}{|Y_i^+||Y_i^-|} \sum_{p \in Y_i^+} \sum_{q \in Y_i^-} \xi_{pq}^i \\
		& s.t. \ \langle \mathbf{w}^p, \mathbf{x}_i \rangle - \langle \mathbf{w}^q, \mathbf{x}_i \rangle \geq 1 - \xi_{pq}^i, \ (p, q) \in Y_i^{+}  \times Y_i^{-} \\
		& \qquad \xi_{pq}^i \geq 0, \ i = 1,...,n \\ 
		\end{split}
		\end{equation}
		
		However, the $t(\mathbf{x}_i)$ depends on the learned parameter $\mathbf{W}$ and the corresponding instance $\mathbf{x}_i$, which makes it difficult to optimize. Thus, we aim to fix the threshold value and set $t(\mathbf{x}_i) = 0, i=1,...,n$ for all the instances for simplicity. Besides, the surrogate least squared hinge loss $loss(y, f(\mathbf{x}))= max(0, 1 - y f(\mathbf{x}))^2 = (| 1 - y f(\mathbf{x}) |_+)^2$ is employed to approximate the thresholding $0-1$ loss. Moreover, we change the hinge-like ranking loss to least squared hinge-like ranking loss to make it smooth for efficient optimization. Furthermore, we replace the value 1 to 2 in the first constraint condition of the optimization problem, which makes it compatible with the label tags (i.e., -1 or 1) and the first term of Eq.\eqref{equation:model2_1}. Therefore, the problem becomes as follows.
		\begin{equation}
		\label{equation:model2_2}
		\begin{split}
		& \min \limits_\mathbf{W} \ \frac{1}{2} \sum_{i=1}^{n} \sum_{j=1}^{l} max(0, 1 - y_{ij} \langle \mathbf{w}^j, \mathbf{x}_i \rangle )^2 + \frac{\lambda_1}{2} \sum_{j=1}^{l} \| \mathbf{w}^j \|^2 \\ 
		& \qquad + \frac{\lambda_2}{2} \sum_{i=1}^{n} \frac{1}{|Y_i^+||Y_i^-|} \sum_{p \in Y_i^+} \sum_{q \in Y_i^-} {\xi_{pq}^i}^2 \\
		& s.t. \ \langle \mathbf{w}^p, \mathbf{x}_i \rangle - \langle \mathbf{w}^q, \mathbf{x}_i \rangle \geq 2 - \xi_{pq}^i, \ (p, q) \in Y_i^{+}  \times Y_i^{-} \\
		& \qquad \xi_{pq}^i \geq 0, \ i = 1,...,n \\ 
		\end{split}
		\end{equation}
		
		Obviously, the constrained optimization problem Eq.\eqref{equation:model2_2} can be equivalently transformed into the following unconstrained optimization problem.
		\begin{equation}
		\label{equation:model2_3}
		\begin{split}
		& \min \limits_\mathbf{W} \ \frac{1}{2} \| (|\mathbf{E} -\mathbf{Y} \circ (\mathbf{XW})|_{+})^2 \|_1 + \frac{\lambda_1}{2} \| \mathbf{W} \|_F^2 + \\
		& \qquad \frac{\lambda_2}{2} \sum_{i=1}^{n} \frac{1}{|Y_i^+||Y_i^-|} \sum_{p \in Y_i^+} \sum_{q \in Y_i^-} max(0, 2 - \langle \mathbf{w}^p - \mathbf{w}^q, \mathbf{x}_i \rangle)^2
		\end{split}
		\end{equation}
		where $\mathbf{E} = \{ 1 \}^{n \times l}$ denotes the matrix with each element equal to $1$, and $\circ$ denotes the Hadamard (element-wise) product of matrices.
		
		It can be easily observed that the first two items of Eq.\eqref{equation:model2_3} (i.e., $\lambda_2 = 0$) actually construct the linear Binary Relevance (BR) \cite{boutell2004learning, wu2018unified}, where the base learner is the least squared hinge Support Vector Machine. Thus, the model can also be viewed as an extension of BR, which aims to additionally consider the minimization of the \emph{Ranking Loss} to boost the performance.
		
		Note that the loss function of the first and third term of Eq.\eqref{equation:model2_3} can be other forms of surrogate loss functions, such as exponential(-like) loss function. Here we adopt least squared hinge(-like) loss because it's not only convex and smooth for efficient optimization but also has good performance in practice \cite{wu2018unified, wu2018cost}. 
	
	\subsection{Robust Low-Rank Learning} 
		Recently, to exploit the high-order label correlations, many approaches \cite{xu2014learning, yu2014large, jing2015semi, xu2016local} have imposed a low-rank constraint on the parameter matrix under the assumption of low dimensional label space, which has shown good performance. Therefore, we also impose the low-rank constraint on $\mathbf{W}$ as follows.
		\begin{equation}
		\label{equation:linear_model_rank}
		\begin{split}
		& \min \limits_\mathbf{W} \ \frac{1}{2} \| (|\mathbf{E} -\mathbf{Y} \circ (\mathbf{XW})|_{+})^2 \|_1 + \frac{\lambda_1}{2} \| \mathbf{W} \|_F^2 + \\
		& \qquad \frac{\lambda_2}{2} \sum_{i=1}^{n} \frac{1}{|Y_i^+||Y_i^-|} \sum_{p \in Y_i^+} \sum_{q \in Y_i^-} max(0, 2 - \langle \mathbf{w}^p - \mathbf{w}^q, \mathbf{x}_i \rangle)^2 \\
		& s.t. \ Rank(\mathbf{W}) \leq k
		\end{split}
		\end{equation}
		
		Obviously, the constrained optimization problem Eq.\eqref{equation:linear_model_rank} can be equivalently transformed into the following unconstrained optimization problem.
		\begin{equation}
		\label{equation:linear_model_1}
		\begin{split}
		& \min \limits_\mathbf{W} \ \frac{1}{2} \| (|\mathbf{E} -\mathbf{Y} \circ (\mathbf{XW})|_{+})^2 \|_1 + \frac{\lambda_1}{2} \| \mathbf{W} \|_F^2 + \lambda_3 Rank(\mathbf{W}) \\
		& \qquad \frac{\lambda_2}{2} \sum_{i=1}^{n} \frac{1}{|Y_i^+||Y_i^-|} \sum_{p \in Y_i^+} \sum_{q \in Y_i^-} max(0, 2 - \langle \mathbf{w}^p - \mathbf{w}^q, \mathbf{x}_i \rangle)^2
		\end{split}
		\end{equation}
		
		Due to the noncontinuity and nonconvexity of the matrix rank function, the above optimization problem is NP-hard. Similar to previous work, we employ the convex trace norm to approximate the matrix rank as follows.
		\begin{equation}
		\label{equation:linear_model}
		\begin{split}
		& \min \limits_\mathbf{W} \ \frac{1}{2} \| (|\mathbf{E} -\mathbf{Y} \circ (\mathbf{XW})|_{+})^2 \|_1 + \frac{\lambda_1}{2} \| \mathbf{W} \|_F^2 + \lambda_3  \| \mathbf{W} \|_* \\
		& \qquad \frac{\lambda_2}{2} \sum_{i=1}^{n} \frac{1}{|Y_i^+||Y_i^-|} \sum_{p \in Y_i^+} \sum_{q \in Y_i^-} max(0, 2 - \langle \mathbf{w}^p - \mathbf{w}^q, \mathbf{x}_i \rangle)^2
		\end{split}
		\end{equation}
		Note that the combination of the Frobenius norm and the trace norm leads to the matrix elastic net (MEN) \cite{li2012error} regularizer, which has shown good performance for exploring inter-target correlations \cite{zhen2018multi} in multi-target regression.
	
	\subsection{Kernelization}
	
		The model in Eq.\eqref{equation:linear_model_rank} (or Eq.\eqref{equation:linear_model}) remains a linear multi-label classifier and thus is less able to handle nonlinear relationships between the input and output effectively. Here we aim to utilize kernel methods to achieve nonlinear multi-label classifiers. However, it's not easy to apply the classical linear representer theorem \cite{scholkopf2002learning, Dinuzzo2012The} to get the kernelization due to the low-rank constraint. Thus, we derive a specific linear representer theorem as shown in Theorem \ref{theorem:linear_representer} for the kernelization of the linear model in Eq.\eqref{equation:linear_model_rank}.
		
		Suppose that $\phi(\cdot)$ is a feature mapping function, which maps $\mathbf{x}$ from $\mathbb{R}^m$ to a Hilbert space $\mathcal{H}$. Thus, the optimization problem \eqref{equation:linear_model_rank} becomes
		\begin{equation}
		\label{equation:kernel_model_rank}
		\begin{split}
		& \min \limits_\mathbf{W} \ \frac{1}{2} \| (|\mathbf{E} -\mathbf{Y} \circ (\Phi(\mathbf{X}) \mathbf{W})|_{+})^2 \|_1 + \frac{\lambda_1}{2} \| \mathbf{W} \|_F^2 + \\
		& \qquad \frac{\lambda_2}{2} \sum_{i=1}^{n} \frac{1}{|Y_i^+||Y_i^-|} \sum_{p \in Y_i^+} \sum_{q \in Y_i^-} max(0, 2 - \langle \mathbf{w}^p - \mathbf{w}^q, \phi(\mathbf{x}_i) \rangle)^2 \\
		& s.t. \ Rank(\mathbf{W}) \leq k
		\end{split}
		\end{equation}
		where $\Phi(\mathbf{X}) = [\phi(\mathbf{x}_1);...;\phi(\mathbf{x}_n)]$ denotes the input data matrix in the Hilbert space.
		\begin{thm}
			\label{theorem:linear_representer}
			Suppose that $\phi(\cdot)$ is a mapping from $\mathbb{R}^m$ to a Hilbert space $\mathcal{H}$. If Eq.\eqref{equation:kernel_model_rank} has a minimizer, there exsits an optimal $\mathbf{W}$ that admits a linear representer theorem of the form
			\begin{equation}
			\label{equation:linear_representer}
			\mathbf{W} = \Phi(\mathbf{X})^\top \mathbf{A}, \ s.t. \ Rank(\mathbf{A}) \leq k
			\end{equation}
			where $\mathbf{A} = [\alpha_1,...,\alpha_l] \in \mathbb{R}^{n \times l},\ \alpha_i \in \mathbb{R}^{n}$.
		\end{thm}
		Please see the proof in \ref{appendix:1} for details.
			
		\emph{Remark.} Important theoretical guarantees are provided by Theorem \ref{theorem:linear_representer} to obtain nonlinear multi-label classifiers. Through specifying the kernel matrix, the RBRL can flexibly handle linear or nonlinear relationships between the input and output.
		
		The linear representer theorem will show great power when the feature mapping space $\mathcal{H}$ is a reproducing kernel Hilbert space (RKHS). Thus, we further suppose $\phi(\cdot)$ maps $\mathbf{x}$ to some high or even infinite dimensional RKHS, where the corresponding kernel function $k(\cdot,\cdot)$ satisfies $k(\mathbf{x}_i, \mathbf{x}_j) = \langle \phi(\mathbf{x}_i), \phi(\mathbf{x}_j) \rangle$.
		
		Since $\| \mathbf{W} \|_F^2 = Tr(\mathbf{W}^\top \mathbf{W})$, plugging \eqref{equation:linear_representer} into \eqref{equation:kernel_model_rank} gives rise to the following optimization problem.
		\begin{equation}
		\label{equation:kernel_model_original}
		\begin{split}
		& \min \limits_\mathbf{A} \ \frac{1}{2} \| (|\mathbf{E} -\mathbf{Y} \circ (\Phi(\mathbf{X}) \Phi(\mathbf{X})^\top \mathbf{A})|_{+})^2 \|_1 + \frac{\lambda_1}{2} Tr(\mathbf{A}^\top \Phi(\mathbf{X}) \Phi(\mathbf{X})^\top \mathbf{A}) + \\ 
		& \qquad \frac{\lambda_2}{2} \sum_{i=1}^{n} \frac{1}{|Y_i^+||Y_i^-|} \sum_{p \in Y_i^+} \sum_{q \in Y_i^-} max(0, 2 - \langle \Phi(\mathbf{X})^\top \alpha_p - \Phi(\mathbf{X})^\top \alpha_q, \phi(\mathbf{x}_i) \rangle )^2 \\
		& s.t. \ Rank(\mathbf{A}) \leq k
		\end{split}
		\end{equation}
		
		Define $\mathbf{K} = \Phi(\mathbf{X}) \Phi(\mathbf{X})^\top \in \mathbb{R}^{n \times n}$ to be the kernel matrix (or Gram matrix) in the RKHS. Similarly, we first transform the above problem to the unconstrained problem and then replace the rank regularization with the trace norm. Consequently, the kernel RBRL is established as follows.
		\begin{equation}
		\label{equation:kernel_model}
		\begin{split}
		& \min \limits_\mathbf{A} \ \frac{1}{2} \| (|\mathbf{E} -\mathbf{Y} \circ (\mathbf{KA})|_{+})^2 \|_1 + \frac{\lambda_1}{2} Tr(\mathbf{A}^\top \mathbf{KA})  + \lambda_3 \| \mathbf{A} \|_* + \\
		& \qquad \frac{\lambda_2}{2} \sum_{i=1}^{n} \frac{1}{|Y_i^+||Y_i^-|} \sum_{p \in Y_i^+} \sum_{q \in Y_i^-} max(0, 2 - \langle \alpha_p - \alpha_q, \mathbf{K}^i \rangle )^2
		\end{split}
		\end{equation}
		where $\mathbf{K}^i = [k(\mathbf{x}_1, \mathbf{x}_i),...,k(\mathbf{x}_n, \mathbf{x}_i)] \in \mathbb{R}^n$, which is the $i$-th column of the kernel matrix $\mathbf{K}$.
		Note that we can also derive the linear representer theorem as the Theorem 4 in \cite{argyriou2008convex} to directly transform the problem Eq.\eqref{equation:linear_model} to Eq.\eqref{equation:kernel_model}. Nevertheless, we don't derive the kernelization in this way because the trace norm is only one way to approximate the matrix rank and there are also other ways to approximate it, such as factorization based approaches \cite{yu2014large}.
		
		Based on the learned parameter $\mathbf{A}$, for an unseen instance $\mathbf{x}_t$, its real-valued prediction is obtained by $\mathbf{f}_t = \mathbf{K}_t^\top \mathbf{A}$, where $\mathbf{K}_t = [k(\mathbf{x}_1, \mathbf{x}_t),...,k(\mathbf{x}_n, \mathbf{x}_t)] \in \mathbb{R}^n$.
	
	\subsection{Comparison with other variant approaches of Rank-SVM}
	
		Here we aim to compare our proposed method RBRL with other classical variant approaches of Rank-SVM. Besides, we focus on two representative methods, i.e., calibrated Rank-SVM \cite{jiang2008calibrated} and Rank-SVMz \cite{xu2012efficient}, which both introduce a virtual zero label for thresholding to tackle the \emph{error accumulation} issue of Rank-SVM. Note that, for the convenience of discussion, the bias terms are both incorporated into the weight parameters due to the little difference between models as mentioned before. Besides, linear models are given for comparison and analysis because the classical linear representer theorem \cite{Kivinen2002Learning, Dinuzzo2012The} can be directly applied to achieve kernel models although the original literature derives the kernel models by use of the dual optimization problems.
		
		To incorporate another virtual zero label into the original ranking learning stage, calibrated Rank-SVM \cite{jiang2008calibrated} is formulated as follows.
		\begin{equation}
		\label{equation:calibrated_ranksvm}
		\begin{aligned}
			& \min \limits_{\mathbf{w}^j, j=0,1,...,l} \frac{1}{2} \sum_{j=0}^{l} \| \mathbf{w}^j \|^2 + C \sum_{i=1}^{n} \frac{1}{|Y_i^+||Y_i^-|} \Big \{ \sum_{p \in Y_i^+} \sum_{q \in Y_i^-} \xi_{pq}^i + \sum_{p \in Y_i^+} \xi_{p0}^i + \sum_{q \in Y_i^-} \xi_{0q}^i \Big \} \\
			& \qquad s.t. \quad \langle \mathbf{w}^p, \mathbf{x}_i \rangle - \langle \mathbf{w}^q, \mathbf{x}_i \rangle \geq 2 - \xi_{pq}^i, \ (p, q) \in Y_i^{+}  \times Y_i^{-} \\
			& \qquad \qquad \ \langle \mathbf{w}^p, \mathbf{x}_i \rangle - \langle \mathbf{w}^0, \mathbf{x}_i \rangle \geq 1 - \xi_{p0}^i, \ p \in Y_i^+ \\
			& \qquad \qquad \ \langle \mathbf{w}^0, \mathbf{x}_i \rangle - \langle \mathbf{w}^q, \mathbf{x}_i \rangle \geq 1 - \xi_{0q}^i, \ q \in Y_i^- \\
			& \qquad \qquad \ \xi_{pq}^i \geq 0, \ \xi_{p0}^i \geq 0, \ \xi_{0q}^i \geq 0
		\end{aligned}
		\end{equation}
		where the first term in the braces aims to minimize the ranking loss, and the last two terms in the braces corresponds to the minimization of the hamming loss. 
		
		To address high computational complexity and error accumulation issue of Rank-SVM, Rank-SVMz \cite{xu2012efficient} also introduces a zero label, which is formulated as follows.
		\begin{equation}
		\label{equation:ranksvmz}
		\begin{aligned}
			& \min \limits_{\mathbf{w}^j, j=0,1,...,l} \frac{1}{2} \sum_{j=0}^{l} \| \mathbf{w}^j \|^2 + C \sum_{i=1}^{n} \Big \{ \frac{1}{|Y_i^+|} \sum_{p \in Y_i^+} \xi_{ip} + \frac{1}{|Y_i^-|} \sum_{q \in Y_i^-} \xi_{iq}  \Big\} \\
			& \qquad s.t. \quad y_{ik} ( \langle \mathbf{w}^k, \mathbf{x}_i \rangle - \langle \mathbf{w}^0, \mathbf{x}_i \rangle ) \geq 1 - \xi_{ik} \\
			& \qquad \qquad \ \xi_{ik} \geq 0, \ k = 1,...,l, \ i = 1,...,n 
		\end{aligned}
		\end{equation}
		where the second term in the objective function aims to minimize the ranking loss. 
		
		From the above discussions, it can be observed that, compared with our method RBRL, the main difference is that calibrated Rank-SVM and Rank-SVMz both introduce another zero label parameter (i.e., $\mathbf{w}^0$) and thus increases the complexity of the model (i.e., the hypothesis set), which generally makes it harder to control the generalization error. Specifically, the hypothesis set of RBRL can be expressed as $\mathcal{H} = \{ \mathbf{W} = [\mathbf{w}^1,...,\mathbf{w}^l] \in \mathbb{R}^{m \times l}:  \mathbf{x} \rightarrow sign(\mathbf{xW}) \}$, whereas for calibrated Rank-SVM and Rank-SVMz, the hypothesis sets are both $\mathcal{H} = \{ \mathbf{W} = [\mathbf{w}^0,...,\mathbf{w}^l] \in \mathbb{R}^{m \times (l+1)}:  \mathbf{x} \rightarrow sign(\mathbf{x}[\mathbf{w}^1 - \mathbf{w}^0,...,\mathbf{w}^l - \mathbf{w}^0] \}$. Besides, Rank-SVMz doesn't explicitly minimize the hamming loss. Furthermore, RBRL additionally utilizes low-rank learning to further exploit the label correlations.
		
	\section{Optimization}
		\label{sec:optimization}
		
		In this section, we concentrate on the optimization algorithms for the RBRL. We use two accelerated proximal gradient methods (APG) to solve the linear and kernel RBRL respectively. Then, the convergence and computational time complexity of the algorithms are analyzed.
		
		\subsection{Algorithms}
			The problems Eq.\eqref{equation:linear_model} and Eq.\eqref{equation:kernel_model} are both convex but nonsmooth due to the trace norm. We seek to solve them by the accelerated proximal gradient methods (APG) \cite{nesterov2005smooth, ji2009accelerated} which have a fast convergence rate $\mathcal{O}(1/t^2)$. Specifically, a general APG aims to solve the following nonsmooth convex problem. 
			\begin{equation}
			\label{euation:convex_problem_nonsmooth}
			\min \limits_{\mathbf{W} \in \mathcal{H}} \ \mathcal{F}(\mathbf{W}) = f(\mathbf{W}) + g(\mathbf{W})
			\end{equation}
			where $f: \mathcal{H} \rightarrow \mathbb{R}$ is a convex and smooth function, $g: \mathcal{H} \rightarrow \mathbb{R}$ is a convex and typically nonsmooth function, and the gradient function $\nabla{f(\cdot)}$ is \emph{Lipschitz} continuous, i.e., $\forall \ \mathbf{W}_1, \mathbf{W}_2 \in \mathcal{H}, \|\nabla{f(\mathbf{W}_1)} - \nabla{f(\mathbf{W}_2)} \| \leq L_f \| \Delta \mathbf{W}\|$, where $\Delta \mathbf{W} = \mathbf{W}_1 - \mathbf{W}_2$ and $L_f$ is the \emph{Lipschitz} constant.
			
			\subsubsection{Linear Model}
			\label{sec:algorithm_linear}
				Here we solve the problem Eq.\eqref{equation:linear_model} for the linear RBRL via the APG. For the convenience of discussion, the objective function in Eq.\eqref{equation:linear_model} (or Eq.\eqref{equation:kernel_model}) is denoted as $\mathcal{F}(\cdot)$. In addition, we denote the last term in Eq.\eqref{equation:linear_model} corresponding to the minimization of ranking loss as follows.
				\begin{equation}
				\label{equation:ranking_loss_function}
				f_r(\mathbf{W}) = \frac{1}{2} \sum_{i=1}^{n} \frac{1}{|Y_i^+||Y_i^-|} \sum_{p \in Y_i^+} \sum_{q \in Y_i^-} max(0, 2 - \langle \mathbf{w}^p - \mathbf{w}^q, \mathbf{x}_i \rangle)^2
				\end{equation}
				Thus, the objective function of Eq.\eqref{equation:linear_model} is split as follows.			
				\begin{equation}
				\label{equation:convex_problem_split1}
				\begin{split}
				& f(\mathbf{W}) = \frac{1}{2} \| (|\mathbf{E} -\mathbf{Y} \circ (\mathbf{XW})|_{+})^2 \|_1 + \frac{\lambda_1}{2} \| \mathbf{W} \|_F^2 + \lambda_2 f_r(\mathbf{W}) \\
				& g(\mathbf{W}) = \lambda_3 \| \mathbf{W} \|_*
				\end{split}
				\end{equation}
				
				In what follows, we first compute the gradient function of $f(\mathbf{W})$ and then compute its \emph{Lipschitz} constant.
				
				Firstly, we can obtain the gradient of $f(\mathbf{W})$ w.r.t. $\mathbf{W}$ as follows.
				\begin{equation}
				\label{equation:model_gradient}
				\begin{split}
				& \nabla_\mathbf{W}{f(\mathbf{W})} = \mathbf{X}^\top (|\mathbf{E} -\mathbf{Y} \circ (\mathbf{XW})|_{+} \circ (-\mathbf{Y}) ) + \lambda_1 \mathbf{W} + \lambda_2 \nabla_\mathbf{W}{f_r(\mathbf{W})}
				\end{split}
				\end{equation}
				where $\nabla_\mathbf{W}{f_r(\mathbf{W})} = [\frac{\partial f_r}{\partial {\mathbf{w}^1}},\frac{\partial f_r}{\partial {\mathbf{w}^2}},...,\frac{\partial f_r}{\partial {\mathbf{w}^l}}]$
				and $for \ j=1,...,l$
				\begin{equation}
				\begin{aligned}
				& \frac{\partial f_r}{\partial {\mathbf{w}^j}} = \sum_{i=1}^{n} \frac{1}{|Y_i^+||Y_i^-|} \Big \{ [\![ j \in Y_i^{+} ]\!] \sum_{q \in Y_i^-} max(0, 2 - \langle {\mathbf{w}^j} - \mathbf{w}^q, \mathbf{x}_i \rangle ) (-\mathbf{x}_i) \\ 
				& \qquad \qquad \qquad \qquad \ + [\![ j \in Y_i^{-} ]\!] \sum_{p \in Y_i^+} max(0, 2 - \langle \mathbf{w}^p - {\mathbf{w}^j}, \mathbf{x}_i \rangle) \mathbf{x}_i \Big \} 
				\end{aligned}			
				\end{equation}
				
				From the above discussion, it can be observed that the main difficulties of the computation of the \emph{Lipschitz} constant for $\nabla_\mathbf{W}{f(\mathbf{W})}$ in Eq.\eqref{equation:model_gradient} lie in the third term $\nabla_\mathbf{W}{f_r(\mathbf{W})}$. Consequently, in what follows, we first deal with the term $\nabla_\mathbf{W}{f_r(\mathbf{W})}$ and compute its \emph{Lipschitz} constant.
				
				We first provide a lemma and then a proposition for the computation of the \emph{Lipschitz} constant for $\nabla_\mathbf{W}{f_r(\mathbf{W})}$.
				\begin{lem}
					$\forall \ a, b \in \mathbb{R}$, the following inequality always holds.
					\begin{equation}
					| |a|_+ - |b|_+ | \leq |a - b|
					\end{equation}
					where $|x|_+ = max(0, x)$.
				\end{lem}
				
				\begin{lem}
				\label{lemma:2}
					$\forall \ a_1, a_2, ... , a_n \in \mathbb{R}^m$, the following inequality always holds.
					\begin{equation}
						\| a_1 + a_2 + ... + a_n \|^2 \leq n \sum_{i=1}^{n} \| a_i \|^2
					\end{equation}
				\end{lem}
				Please see the proof in \ref{appendix:2} for details.		
				
				\begin{pro}
				\label{proposition:linear_lipschitz_constant_part}
					The Lipschitz constant of $\nabla_\mathbf{W}{f_r(\mathbf{W})}$ w.r.t. $\mathbf{W}$ in Eq.\eqref{equation:ranking_loss_function} is
					\begin{equation}
					L_{f_r} = \sqrt{max\{ A_j B_j \}_{j=1,...,l}}
					\end{equation}
					where $l$ is the number of the labels, $B_j = \sum_{i=1}^{n} \frac{ ( [\![ j \in Y_i^{+} ]\!] |Y_i^-| + [\![ j \in Y_i^{-} ]\!] |Y_i^+| ) \| \mathbf{x}_i \|^4 }{|Y_i^+|^2 |Y_i^-|^2}$, and $A_j = \sum_{i=1}^{n} ([\![ j \in Y_i^{+} ]\!] |Y_i^-| + [\![ j \in Y_i^{-} ]\!] |Y_i^+| ), j = 1,...,l$.
				\end{pro}
				Please see the proof in \ref{appendix:3} for details.
				
				Finally, a theorem is given to compute the \emph{Lipschitz} constant of $\nabla_\mathbf{W}{f(\mathbf{W})}$ in Eq.\eqref{equation:model_gradient}. 
				\begin{thm}
				\label{theorem:linear_lipschitz_constant}
					The Lipschitz constant of $\nabla_\mathbf{W}{f(\mathbf{W})}$ w.r.t. $\mathbf{W}$ in Eq.\eqref{equation:model_gradient} is
					\begin{equation}
					\label{equation:linear_lipschitz_constant}
					L_f = \sqrt{ 3 (\| \mathbf{X} \|_F^2)^2 + 3 \lambda_1^2 + 3 (\lambda_2 L_{f_r})^2 }
					\end{equation}
				\end{thm}	
				Please see the proof in \ref{appendix:4} for details.

				Besides, the singular value thresholding (SVT) operator \cite{cai2010singular} is as follows.
				\begin{equation}
				\textbf{prog}_\epsilon (\mathbf{W}) =  \mathbf{U} \mathbf{\Sigma}_{\epsilon} \mathbf{V}^\top
				\end{equation}
				where $\mathbf{W}$ has the singular value decomposition (SVD) $\mathbf{W} = \mathbf{U \Sigma V}^\top$ in which $\mathbf{U}$ and $\mathbf{V}$ are unitary matrices and $\mathbf{\Sigma}$ is the diagonal matrix with real numbers on the diagonal, and $\mathbf{\Sigma}_{\epsilon}$ is a diagonal matrix with $(\mathbf{\Sigma}_{\epsilon})_{ii} = max(0, \mathbf{\Sigma}_{ii} - \epsilon)$.
				
				In addition, the main iterations in the APG are as follows.
				\begin{eqnarray}
				\mathbf{W}_t & \longleftarrow & \textbf{prox}_{(\lambda_3 / L_f)} ( \mathbf{G}_t - \frac{1}{L_f} \nabla_{\mathbf{G}_{t}}{f(\mathbf{G}_{t}) )} \\
				b_{t+1} & \longleftarrow & \frac{1 + \sqrt{1 + 4 b_t^2}}{2} \\
				\mathbf{G}_{t+1} & \longleftarrow & \mathbf{W}_t + \frac{b_t - 1}{b_{t+1}} (\mathbf{W}_t - \mathbf{W}_{t-1})
				\end{eqnarray}
				
				Moreover, Algorithm \ref{code:linear_model} summarizes the detailed APG to solve Eq.\eqref{equation:linear_model}.
				\begin{algorithm}[!htb]
					\label{code:linear_model}
					\caption{Accelerated Proximal Gradient method for Eq.\eqref{equation:linear_model}} 
					\LinesNumbered 
					\KwIn{$\mathbf{X} \in \mathbb{R}^{n \times m}, \mathbf{Y} \in \{-1, 1\}^{n \times l}$, tradeoff hyperparameter $\lambda_1, \lambda_2, \lambda_3$} 
					\KwOut{$ \mathbf{W}^* \in \mathbb{R}^{m \times l} $} 
					Initialize $t = 1$, $b_1 = 1$;\\
					Initialize $\mathbf{G}_1 = \mathbf{W}_0 \in \mathbb{R}^{m \times l}$ as zero matrix;\\
					Compute $L_f$ according to Eq.\eqref{equation:linear_lipschitz_constant};\\
					\While{{\rm (\ref{equation:linear_model})} not converge}{
						Compute the gradient of $\nabla_{\mathbf{G}_{t}}{f(\mathbf{G}_{t})}$ via Eq.\eqref{equation:model_gradient};\\
						$\mathbf{W}_t \longleftarrow \textbf{prox}_{(\lambda_3 / L_f)} (\mathbf{G}_t - \frac{1}{L_f} \nabla_{\mathbf{G}_{t}}{f(\mathbf{G}_{t})})$;\\
						$b_{t+1} \longleftarrow \frac{1 + \sqrt{1 + 4 b_t^2}}{2}$;\\
						$\mathbf{G}_{t+1} \longleftarrow \mathbf{W}_t + \frac{b_t - 1}{b_{t+1}} (\mathbf{W}_t - \mathbf{W}_{t-1})$;\\
						$t = t + 1$;
					}
					$\mathbf{W}^* = \mathbf{W}_{t-1}$;
				\end{algorithm}
		
			\subsubsection{Kernel Model}
				Here we solve the problem Eq.\eqref{equation:kernel_model} for the kernel RBRL via the APG. For the convenience of discussion, we denote the last term in Eq.\eqref{equation:kernel_model} as follows.
				\begin{equation}
				\label{equation:ranking_loss_function_kernel}
				f_r(\mathbf{A}) = \frac{1}{2} \sum_{i=1}^{n} \frac{1}{|Y_i^+||Y_i^-|} \sum_{p \in Y_i^+} \sum_{q \in Y_i^-} max(0, 2 - \langle \alpha_p - \alpha_q, \mathbf{K}^i \rangle )^2
				\end{equation}
				Similarly, we split the objective function of Eq.\eqref{equation:kernel_model} as follows.
				\begin{equation}
				\label{equation:convex_problem_split}
				\begin{split}
				& f(\mathbf{A}) = \frac{1}{2} \| (|\mathbf{E} -\mathbf{Y} \circ (\mathbf{KA})|_{+})^2 \|_1 + \frac{\lambda_1}{2} Tr(\mathbf{A}^{\top} \mathbf{K} \mathbf{A}) + \lambda_2 f_r(\mathbf{A}) \\
				& g(\mathbf{A}) = \lambda_3 \| \mathbf{A} \|_*
				\end{split}
				\end{equation}
				
				In what follows, we first provide the gradient of $f(\mathbf{A})$ and then compute its \emph{Lipschitz} constant.
				
				Firstly, we can obtain the gradient of $f(\mathbf{A})$ w.r.t. $\mathbf{A}$ as follows.
				\begin{equation}
				\label{equation:kernel_model_gradient}
				\begin{split}
				& \nabla_\mathbf{A}{f(\mathbf{A})} = \mathbf{K}^\top (|\mathbf{E} -\mathbf{Y} \circ (\mathbf{KA})|_{+} \circ (-\mathbf{Y}) ) + \lambda_1 \mathbf{KA} + \lambda_2 \nabla_\mathbf{A}{f_r(\mathbf{A})} 
				\end{split}
				\end{equation}
				where $\nabla_\mathbf{A}{f_r(\mathbf{A})} = [\frac{\partial f_r}{\partial {\alpha_1}},\frac{\partial f_r}{\partial {\alpha_2}},...,\frac{\partial f_r}{\partial {\alpha_l}}]$
				and $for \ j=1,...,l$
				\begin{equation}
				\begin{aligned}
				& \frac{\partial f_r}{\partial {\alpha_j}} = \sum_{i=1}^{n} \frac{1}{|Y_i^+||Y_i^-|} \Big \{ [\![ j \in Y_i^{+} ]\!] \sum_{q \in Y_i^-} max(0, 2 - \langle \alpha_j - \alpha_q, \mathbf{K}^i \rangle) (-\mathbf{K}^i) \\ 
				& \qquad \qquad \qquad \qquad \ + [\![ j \in Y_i^{-} ]\!] \sum_{p \in Y_i^+} max(0, 2 - \langle \alpha_p - \alpha_j, \mathbf{K}^i \rangle) \mathbf{K}^i \Big \} 
				\end{aligned}			
				\end{equation}
				
				Then, similar to Section \ref{sec:algorithm_linear} for the optimization of the linear RBRL, we provide a proposition and a theorem to compute the \emph{Lipschitz} constant of $\nabla_\mathbf{A}{f(\mathbf{A})}$ in Eq.\eqref{equation:kernel_model_gradient} and leave out the proof here owing to lack of space.
				\begin{pro}
					The Lipschitz constant of $\nabla_\mathbf{A}{f_r(\mathbf{A})}$ w.r.t. $\mathbf{A}$ in Eq.\eqref{equation:ranking_loss_function_kernel} is
					\begin{equation}
					L_{f_r} = \sqrt{max\{ A_j B_j \}_{j=1,...,l}}
					\end{equation}
					where $l$ is the number of the labels, $B_j = \sum_{i=1}^{n} \frac{ ( [\![ j \in Y_i^{+} ]\!] |Y_i^-| + [\![ j \in Y_i^{-} ]\!] |Y_i^+| ) \| \mathbf{K}^i \|^4 }{|Y_i^+|^2 |Y_i^-|^2}$, and $A_j = \sum_{i=1}^{n} ([\![ j \in Y_i^{+} ]\!] |Y_i^-| + [\![ j \in Y_i^{-} ]\!] |Y_i^+| ), j = 1,...,l$.
				\end{pro}
				\begin{thm}
					The Lipschitz constant of $\nabla_\mathbf{A}{f(\mathbf{A})}$ w.r.t. $\mathbf{A}$ in Eq.\eqref{equation:kernel_model_gradient} is
					\begin{equation}
					\label{equation:kernel_lipschitz_constant}
					L_f = \sqrt{ 3 (\| \mathbf{K} \|_F^2)^2 + 3 \| \lambda_1 \mathbf{K} \|_F^2 + 3 (\lambda_2 L_{f_r})^2 }
					\end{equation}
				\end{thm}
				
				Moreover, Algorithm \ref{code:kernel_model} summarizes the detailed APG to solve Eq.\eqref{equation:kernel_model}.
				\begin{algorithm}[!htb]
					\label{code:kernel_model}
					\caption{Accelerated Proximal Gradient method for Eq.\eqref{equation:kernel_model}} 
					\LinesNumbered 
					\KwIn{$\mathbf{X} \in \mathbb{R}^{n \times m}, \mathbf{Y} \in \{-1, 1\}^{n \times l}$, the kernel matrix $\mathbf{K} \in \mathbb{R}^{n \times n}$, tradeoff hyper-parameters $\lambda_1, \lambda_2, \lambda_3$} 
					\KwOut{$ \mathbf{A}^* \in \mathbb{R}^{n \times l} $} 
					Initialize $t = 1$, $b_1 = 1$;\\
					Initialize $\mathbf{G}_1 = \mathbf{A}_0 \in \mathbb{R}^{n \times l}$ as zero matrix;\\
					Compute $L_f$ according to Eq.\eqref{equation:kernel_lipschitz_constant};\\
					\While{{\rm (\ref{equation:kernel_model})} not converge}{
						Compute the gradient of $\nabla_{\mathbf{G}_{t}}{f(\mathbf{G}_{t})}$ via Eq.\eqref{equation:kernel_model_gradient};\\
						$\mathbf{A}_t \longleftarrow \textbf{prox}_{(\lambda_3 / L_f)} (\mathbf{G}_t - \frac{1}{L_f} \nabla_{\mathbf{G}_{t}}{f(\mathbf{G}_{t})})$;\\
						$b_{t+1} \longleftarrow \frac{1 + \sqrt{1 + 4 b_t^2}}{2}$;\\
						$\mathbf{G}_{t+1} \longleftarrow \mathbf{A}_t + \frac{b_t - 1}{b_{t+1}} (\mathbf{A}_t - \mathbf{A}_{t-1})$;\\
						$t = t + 1$;
					}
					$\mathbf{A}^* = \mathbf{A}_{t-1}$;
				\end{algorithm}
	
		\subsection{Convergence and Computational Complexity Analysis}
	
			First we analyze the convergence property of the optimization parts in Algorithm \ref{code:linear_model} (or Algorithm \ref{code:kernel_model}). Thanks to the superiority of the APG, it can guarantee Eq.\eqref{equation:linear_model} (or Eq.\eqref{equation:kernel_model}) to converge to a global optimum and have the convergence rate such that $\mathcal{F}(\mathbf{W}_t) - \mathcal{F}(\mathbf{W}^*) \leq \mathcal{O}(1/t^2)$ (or $\mathcal{F}(\mathbf{A}_t) - \mathcal{F}(\mathbf{A}^*) \leq \mathcal{O}(1/t^2)$), where $\mathbf{W}^*$ (or $\mathbf{A}^*$) is an optimal solution of Eq.\eqref{equation:linear_model} (or Eq.\eqref{equation:kernel_model}). In other words, to get an $\epsilon$ tolerance accurate solution, it needs $\mathcal{O}(1/\sqrt{\epsilon})$ iteration rounds.
			
			Then, the computational time complexity of Algorithm \ref{code:linear_model} is analyzed as follows. In step 3, the computation of the \emph{Lipschitz} constant needs $\mathcal{O}(nm^2 + nl^2)$. In each iteration, step 5, which needs $\mathcal{O}(nm^2 + m^2l + nml^2)$, is the most expensive. Regularly, for an $m \times l$ matrix, the computation of the singular value decomposition (SVD) requires $\mathcal{O}(ml{\rm min}(m,l))$ \cite{golub1996matrix}. Since it usually meets $m \gg l$ in multi-label classification, step 6 requires $\mathcal{O}(ml^2)$. Besides, it needs $\mathcal{O}(ml)$ in step 8. Consequently, for Algorithm \ref{code:linear_model}, the overall computational time complexity is $\mathcal{O}((nm^2 + m^2l + nml^2)/\sqrt{\epsilon})$ to get an $\epsilon$ tolerance accurate solution.
			
			Here the computational time complexity of Algorithm \ref{code:kernel_model} is analyzed. In step 3, the computation of the \emph{Lipschitz} constant leads to $\mathcal{O}(n^3 + nl^2)$. In each iteration, step 5, which needs $\mathcal{O}(n^3 + n^2l^2)$, is the most expensive. It leads to $\mathcal{O}(nl^2)$ in step 6. Besides, it needs $\mathcal{O}(nl)$ in step 8. Hence, for Algorithm \ref{code:kernel_model}, the overall computational time complexity is $\mathcal{O}((n^3 + n^2l^2)/\sqrt{\epsilon})$ to get an $\epsilon$ tolerance accurate solution.
			
			Furthermore, based on the computational complexity analysis, it can be observed that when the number of instances (i.e. $n$) or labels (i.e. $l$) is large, a stochastic variant of our method might be a better choice for efficient training. 
				
	\section{Experiments}
	\label{sec:experiments}
	\subsection{Datasets and settings}
		In our experiments, we conduct comparative experiments on 10 widely-used multi-label benchmark datasets, which covers a broad range of sizes and domains. Different datasets might demonstrate diverse label correlations and input-output relationships, which poses great challenges for multi-label classification (MLC) approaches. The statistics of the experimental datasets are summarized in Table \ref{tab:multi_label_dataset}. For each dataset, 60\% is randomly split for training, and the rest 40\% is for testing. Besides, ten independent experiments are repeated for the reduction of statistical variability.
		\begin{table}[!htb]
			\scriptsize
			\renewcommand{\arraystretch}{1.0}
			\centering
			\caption{Statistics of the experimental datasets. (``Cardinality" indicates the average number of labels per example. ``Density" normalizes the ``Cardinality" by the number of possible labels. ``URL" indicates the source URL of the dataset.)}
			\label{tab:multi_label_dataset}
			\begin{tabular}{lcccccll}
				\hline
				Dataset & \#Instance & \#Feature & \#Label & Cardinality & Density & Domain & URL\footnotemark[1] \\
				\hline
				emotions & 593  & 72  & 6  & 1.869 & 0.311 & music & URL 1\\
				image    & 2000 & 294 & 5  & 1.240 & 0.248 & image & URL 2\\
				scene    & 2407 & 294 & 6  & 1.074 & 0.179 & image & URL 1\\
				yeast    & 2417 & 103 & 14 & 4.237 & 0.303 & biology & URL 1\\
				enron   & 1702 & 1001 & 53  & 3.378 & 0.064 & text & URL 1\\
				arts     & 5000 & 462 & 26 & 1.636 & 0.063 & text & URL 1\\
				education& 5000 & 550 & 33 & 1.461 & 0.044 & text & URL 1\\
				recreation& 5000& 606 & 22 & 1.423 & 0.065 & text & URL 1\\
				science  & 5000 & 743 & 40 & 1.451 & 0.036 & text & URL 1\\
				business & 5000 & 438 & 30 & 1.588 & 0.053 & text & URL 1\\
				\hline
			\end{tabular}
		\end{table}
		\footnotetext[1]{URL 1: http://mulan.sourceforge.net/datasets-mlc.html}
		\footnotetext[1]{URL 2: http://palm.seu.edu.cn/zhangml/}
		
		We compare the proposed approach RBRL with the following state-of-the-art baseline approaches for MLC. For each baseline approach, the setting and search range of the hyper-parameters are employed by recommendations of the original literature.
		\begin{enumerate}[(1)]
			\item RBRL\footnotemark[2]. It is our proposed approach, where the hyper-parameters $\lambda_1, \lambda_2, \lambda_3$ are tuned in $\{10^{-4}, 10^{-3}, ..., 10^2 \}$.
			\item Rank-SVM\footnotemark[3] \cite{elisseeff2001kernel}. It is an adaption of the maximum margin strategy for MLC.
			\item Rank-SVMz\footnotemark[4] \cite{xu2012efficient}. It is a variant method of Rank-SVM for MLC.
			\item BR \cite{boutell2004learning}. It converts MLC task into many independent binary classification problems.
			\item ML-kNN \cite{zhang2007ml}. It is an adaption of the lazy learning kNN classifier for MLC.
			\item CLR \cite{furnkranz2008multilabel}. It converts MLC task into the pairwise label ranking problem.
			\item RAKEL \cite{tsoumakas2011random}. It converts MLC task into an ensemble of multi-class classification problems.
			\item CPNL \cite{wu2018cost}. It is a recent approach which presents a cost-sensitive loss to address the class-imbalance issue and boosts the performance by exploitations of negative and positive label correlations.
			\item MLFE\footnotemark[5] \cite{zhang2018feature}. It is also a recent approach that leverages the structural information of the feature space to enrich the labeling information. 
		\end{enumerate}
		\footnotetext[2]{source code: https://github.com/GuoqiangWoodrowWu/RBRL}
		\footnotetext[3]{source code: http://palm.seu.edu.cn/zhangml/}
		\footnotetext[4]{source code: http://computer.njnu.edu.cn/Lab/LABIC/LABIC\_Software.html}
		\footnotetext[5]{source code: http://palm.seu.edu.cn/zhangml/}
		
		Motivated by \cite{hsu2003practical, wu2018cost}, which concludes that the linear model is good enough for high dimensional feature problems, we test the performance of the linear model on the last six datasets and evaluate the kernel model on the first four datasets (i.e., \emph{emotions}, \emph{image}, \emph{scene} and \emph{yeast}).
		
		For the last six datasets, LIBLINEAR \cite{fan2008liblinear} is employed as the base learner for BR, CLR and RAKEL, where the hyper-parameter $C$ is tuned in $\{10^{-4}, 10^{-3}, ..., 10^4 \}$. The commonly-used MULAN \cite{tsoumakas2011mulan} implementations are adopted for BR, ML-kNN, CLR, and RAKEL. Besides, thanks to the publicly available, other approaches employ the original implementations from the corresponding source code website. Moreover, for the sake of fairness, Rank-SVM and Rank-SVMz adopt the linear kernel, and we employ the linear RBRL.
		
		For the first four datasets, LIBSVM \cite{CC01a} is employed as the base learner for BR, CLR and RAKEL, where the hyper-parameter $C$ is tuned in $\{10^{-4}, 10^{-3}, ..., 10^4 \}$. Although other kernel functions (such as polynomial kernel function) can be used, we adopt the RBF kernel $k(\mathbf{x}_i, \mathbf{x}_j) = exp(-\gamma \| \mathbf{x}_i - \mathbf{x}_j \|^2)$ due to its wide applications in practice, and the kernel parameter $\gamma$ is set to be $1/m$ ($m$ is the feature number). Besides, Rank-SVM, Rank-SVMz and our proposed approach RBRL also use the RBF kernel for a fair comparison.

		Furthermore, for each compared approach, its hyper-parameters are selected via fivefold cross-validation on the training set. For convenience, Table \ref{tab:multi_label_para_setup} summarizes the hyper-parameters setup for each compared approach in detail.
		\begin{table}[!htb]
			\scriptsize
			\renewcommand{\arraystretch}{1.0}
			\centering
			\caption{Summary of the hyper-parameters setup for each compared approach}
			\label{tab:multi_label_para_setup}
			\begin{tabular}{lcc}
				\hline
				Approach & Hyper-parameters setup & Citation\\
				\hline
				RBRL & $\lambda_1, \lambda_2, \lambda_3 = \{10^{-4}, 10^{-3}, ..., 10^2 \}$ & N/A \\
				Rank-SVM & $C = \{10^{-4}, 10^{-3}, ..., 10^4 \}$ & \cite{elisseeff2001kernel} \\
				Rank-SVMz & $C = \{10^{-4}, 10^{-3}, ..., 10^4 \}$ & \cite{xu2012efficient} \\
				BR & $C = \{10^{-4}, 10^{-3}, ..., 10^4 \}$ & \cite{boutell2004learning} \\			
				ML-kNN & $k = \{4, 6, ..., 16\}$ & \cite{zhang2007ml} \\
				CLR & $C = \{10^{-4}, 10^{-3}, ..., 10^4 \}$ & \cite{furnkranz2008multilabel} \\
				RAKEL & (1) $ m = 10, k = l/2$ (2) $m = 2l, k = 3$ & \cite{tsoumakas2011random} \\
				& $C = \{10^{-4}, 10^{-3}, ..., 10^4 \}$ & \\
				CPNL & $\beta = 0.5$, $k = 3$ & \cite{wu2018cost} \\
				& $\lambda_1, \lambda_2, \lambda_3 = \{10^{-4}, 10^{-3}, ..., 10^2 \}$ & \\
				MLFE & $\beta_1 = \{1,2,...,10\}$, $\beta_2 = \{1, 10, 15\}$ & \cite{zhang2018feature} \\
				& $\beta_3 = \{1, 10\}$ & \\	
				\hline
			\end{tabular}
		\end{table}
	
	\subsection{Evaluation metrics}
	
		Given a testing set $\mathcal{D}_t = \{\mathbf{x}_i, \mathbf{y}_i\}_{i=1}^{n_t}$ and the family of $l$ learned functions $F = \{f_1,f_2,...,f_l\}$, where $\mathbf{y}_i \in \{-1, +1\}^l$ is the ground-truth labels of $\mathbf{x}_i$. Besides, $H = \{h_1,h_2,...,h_l\}$ denotes the multi-label classifier. In this paper, we employ the following six commonly evaluation metrics \cite{zhang2014review,wu2017unified}, which includes three classification-based metrics (i.e., Hamming Loss, Subset Accuracy and F1-Example) and three ranking-based metrics (i.e., Ranking Loss, Coverage and Average Precision).
		\begin{enumerate}[(1)]
			\item Hamming Loss (Hal): It measures the fraction of misclassified example-label pairs.
			\begin{equation}
			{\rm Hal} = \frac{1}{n_t l} \sum_{i=1}^{n_t} \sum_{j=1}^{l} [\![ h_j(\mathbf{x}_i) \neq y_{ij} ]\!]
			\end{equation}
			\item Subset Accuracy (Sa): It measures the fraction that the predicted label subset and the ground-truth label subset are the same.
			\begin{equation}
			{\rm Sa} = \frac{1}{n_t} \sum_{i=1}^{n_t} [\![ H(\mathbf{x}_i) = \mathbf{y}_i ]\!]
			\end{equation}
			\item F1-Example (F1e): It is the average F1 measure that is the harmonic mean of recall and precision over each instance.
			\begin{equation}
			{\rm F1e} = \frac{1}{n_t} \sum_{i=1}^{n_t} \frac{2 | P_i^+ \cap Y_i^+|}{| P_i^+ | + | Y_i^+ |}
			\end{equation}
			where $Y_i^{+}$ (or $Y_i^{-}$) denotes the index set of the ground-truth relevant (or irrelevant) labels associated with $\mathbf{x}_i$, and $P_i^+$ (or $P_i^-$) denotes the index set of the predicted relevant (or irrelevant) labels associated with $\mathbf{x}_i$.
			\item Ranking Loss (Ral): It measures the average fraction of the label pairs that an irrelevant label ranks higher than a relevant label over each instance.
			\begin{equation}
			{\rm Ral} = \frac{1}{n_t} \sum_{i=1}^{n_t} \frac{|SetR_i|}{| Y_i^+ | | Y_i^- |}
			\end{equation}
			where $SetR_i = \{ (p, q) | f_p(\mathbf{x}_i) \leq f_q(\mathbf{x}_i), (p, q) \in Y_i^+ \times Y_i^- \}$.
			\item Coverage (Cov): It measures how many steps are needed to cover all relevant labels averagely based on the predicted ranking label list. Besides, this metric is normalized in $[0, 1]$ by the number of possible labels here.
			\begin{equation}
			{\rm Cov} = \frac{1}{l} (\frac{1}{n_t} \sum_{i=1}^{n_t} \mathop{\max}_{j \in Y_i^+} rank_F(\mathbf{x}_i, j) - 1)
			\end{equation}
			where $rank_F(\mathbf{x}_i, j)$ stands for the rank of label $j$ in the ranking list based on $F(\mathbf{x}_i)$ which is sorted in descending order.
			\item Average Precision (Ap): It measures the average fraction of relevant labels ranked higher than a specific relevant label.
			\begin{equation}
			{\rm Ap} = \frac{1}{n_t} \sum_{i=1}^{n_t} \frac{1}{| Y_i^+ |} \sum_{j \in Y_i^+} \frac{| SetP_{ij} |}{rank_F(\mathbf{x}_i, j)}	
			\end{equation}
			where $SetP_{ij} = \{ k \in Y_i^+ | rank_F(\mathbf{x}_i, k) \le rank_F(\mathbf{x}_i, j) \}$.
		\end{enumerate}
		
		For Sa, F1e and Ap, the larger the value, the better performance of the multi-label classifier; while for the others, the smaller the value, the better performance of the multi-label classifier.
	
	\subsection{Experiments Results}
		\subsubsection{Performance Comparison}
		\label{sec:comparison}
\begin{figure}[!htbp]
				\centering
				\includegraphics[width=5.0in]{{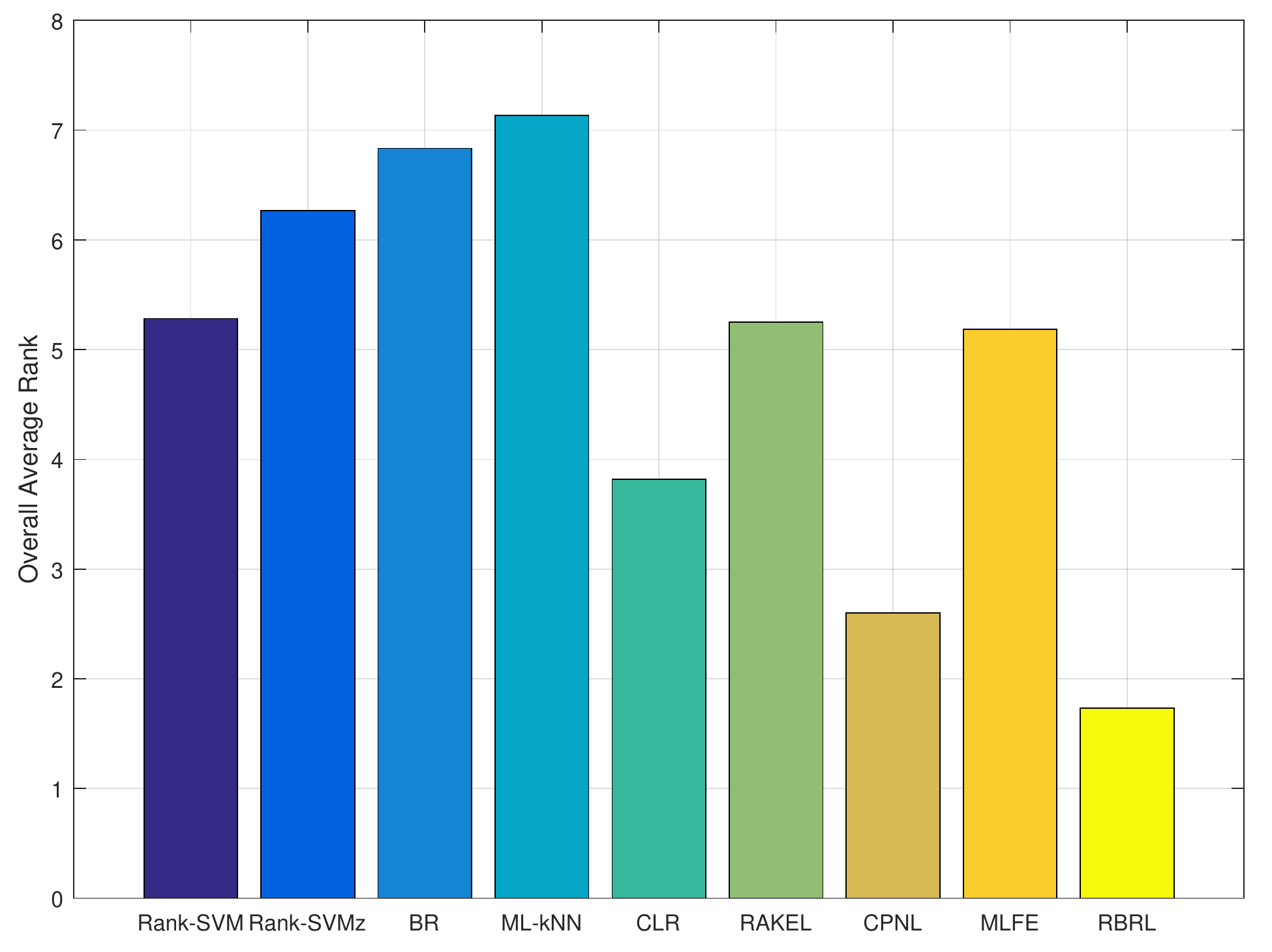}}
				\caption{Overall average ranks of the compared approaches on all the metrics.}
				\label{fig:averagerank_multi_label} 
			\end{figure}

			The multi-label prediction results of the proposed RBRL and the comparison with several state-of-the-art approaches are summarized in Table \ref{tab:multi_label_result}. Besides, the average ranks of these compared approaches in terms of each metric on all the datasets are summarized in Table \ref{tab:multi_label_result_average_rank}. Intuitively, Figure \ref{fig:averagerank_multi_label} shows the overall average ranks of these compared approaches over all the metrics.
			\begin{table}[!htbp]
				\tiny
				\renewcommand{\arraystretch}{1.44}
				\caption{Experimental results of each benchmark approach (mean $\pm$ std) on 10 multi-label datasets. $\downarrow(\uparrow)$ indicates the smaller(larger) the better. Best results are highlighted in bold.}
				\hskip-115pt
				\label{tab:multi_label_result}
				\begin{tabular}{c|c|ccccccccc}
					\hline
					& Metric & Rank-SVM & Rank-SVMz & BR & ML-kNN & CLR & RAKEL & CPNL & MLFE & RBRL \\
					\hline
					\multirow{6}{*}{emotions} & Hal ($\downarrow$) & $0.189 \pm 0.008$ & $0.201 \pm 0.008$ & $0.183 \pm 0.009$ & $0.200 \pm 0.013$ & $0.182 \pm 0.008$ & $\bf 0.177 \pm 0.008$ & $0.183 \pm 0.009$ & $0.186 \pm 0.009$ & $0.181 \pm 0.012$ \\
					& Sa ($\uparrow$)  & $0.291 \pm 0.025$ & $0.292 \pm 0.024$ & $0.313 \pm 0.015$ & $0.285 \pm 0.028$ & $0.318 \pm 0.013$ & $\bf 0.356 \pm 0.028$ & $0.324 \pm 0.035$ & $0.291 \pm 0.035$ & $0.334 \pm 0.032$ \\
					& F1e ($\uparrow$) & $0.645 \pm 0.020$ & $0.675 \pm 0.012$ & $0.620 \pm 0.020$ & $0.605 \pm 0.039$ & $0.624 \pm 0.019$ & $0.679 \pm 0.019$ & $\bf 0.684 \pm 0.019$ & $0.621 \pm 0.021$ & $0.666 \pm 0.022$ \\
					& Ral ($\downarrow$) & $0.155 \pm 0.009$ & $0.149 \pm 0.008$ & $0.246 \pm 0.015$ & $0.169 \pm 0.016$ & $0.149 \pm 0.014$ & $0.192 \pm 0.017$ & $0.139 \pm 0.010$ & $0.142 \pm 0.011$ & $\bf 0.138 \pm 0.011$ \\
					& Cov ($\downarrow$) & $0.294 \pm 0.011$ & $0.291 \pm 0.008$ & $0.386 \pm 0.017$ & $0.306 \pm 0.016$ & $0.283 \pm 0.012$ & $0.338 \pm 0.014$ & $\bf 0.277 \pm 0.010$ & $0.282 \pm 0.013$ & $\bf 0.277 \pm 0.010$ \\
					& Ap ($\uparrow$) & $0.808 \pm 0.010$ & $0.819 \pm 0.012$ & $0.760 \pm 0.015$ & $0.796 \pm 0.016$ & $0.813 \pm 0.014$ & $0.801 \pm 0.015$ & $\bf 0.828 \pm 0.010$ & $0.822 \pm 0.012$ & $\bf 0.828 \pm 0.014$ \\
					\hline
					\multirow{6}{*}{image} & Hal ($\downarrow$) & $0.161 \pm 0.005$ & $0.177 \pm 0.010$ & $0.156 \pm 0.006$ & $0.175 \pm 0.007$ & $0.157 \pm 0.006$ & $0.154 \pm 0.006$ & $0.150 \pm 0.006$ & $0.156 \pm 0.007$ & $\bf 0.149 \pm 0.005$ \\
					& Sa ($\uparrow$)  & $0.451 \pm 0.018$ & $0.411 \pm 0.025$ & $0.482 \pm 0.018$ & $0.393 \pm 0.024$ & $0.477 \pm 0.016$ & $0.527 \pm 0.013$ & $0.533 \pm 0.017$ & $0.463 \pm 0.015$ & $\bf 0.552 \pm 0.011$ \\
					& F1e ($\uparrow$) & $0.631 \pm 0.016$ & $0.670 \pm 0.022$ & $0.623 \pm 0.014$ & $0.503 \pm 0.026$ & $0.627 \pm 0.014$ & $0.680 \pm 0.012$ & $\bf 0.698 \pm 0.012$ & $0.593 \pm 0.014$ & $0.688 \pm 0.012$ \\
					& Ral ($\downarrow$) & $0.143 \pm 0.008$ & $0.142 \pm 0.014$ & $0.220 \pm 0.012$ & $0.180 \pm 0.010$ & $0.144 \pm 0.006$ & $0.173 \pm 0.009$ & $\bf 0.132 \pm 0.006$ & $0.142 \pm 0.007$ & $0.133 \pm 0.007$ \\
					& Cov ($\downarrow$) & $0.171 \pm 0.008$ & $0.170 \pm 0.012$ & $0.227 \pm 0.008$ & $0.198 \pm 0.009$ & $0.168 \pm 0.007$ & $0.191 \pm 0.007$ & $\bf 0.157 \pm 0.006$ & $0.165 \pm 0.006$ & $0.160 \pm 0.006$ \\
					& Ap ($\uparrow$) & $0.823 \pm 0.010$ & $0.826 \pm 0.016$ & $0.772 \pm 0.011$ & $0.786 \pm 0.009$ & $0.826 \pm 0.006$ & $0.813 \pm 0.008$ & $\bf 0.839 \pm 0.007$ & $0.826 \pm 0.008$ & $0.836 \pm 0.008$ \\
					\hline
					\multirow{6}{*}{scene} & Hal ($\downarrow$) & $0.092 \pm 0.005$ & $0.113 \pm 0.004$ & $0.077 \pm 0.002$ & $0.091 \pm 0.003$ & $0.078 \pm 0.002$ & $0.075 \pm 0.004$ & $0.077 \pm 0.003$ & $0.083 \pm 0.003$ & $\bf 0.073 \pm 0.004$ \\
					& Sa ($\uparrow$)  & $0.563 \pm 0.018$ & $0.500 \pm 0.015$ & $0.655 \pm 0.009$ & $0.615 \pm 0.019$ & $0.650 \pm 0.010$ & $0.696 \pm 0.014$ & $0.699 \pm 0.014$ & $0.617 \pm 0.009$ & $\bf 0.735 \pm 0.013$ \\
					& F1e ($\uparrow$) & $0.664 \pm 0.015$ & $0.756 \pm 0.005$ & $0.717 \pm 0.010$ & $0.678 \pm 0.022$ & $0.718 \pm 0.011$ & $0.756 \pm 0.012$ & $0.802 \pm 0.009$ & $0.685 \pm 0.010$ & $\bf 0.803 \pm 0.010$ \\
					& Ral ($\downarrow$) & $0.065 \pm 0.005$ & $0.072 \pm 0.005$ & $0.128 \pm 0.006$ & $0.083 \pm 0.006$ & $0.061 \pm 0.003$ & $0.087 \pm 0.005$ & $0.059 \pm 0.003$ & $0.063 \pm 0.003$ & $\bf 0.058 \pm 0.004$ \\
					& Cov ($\downarrow$) & $0.068 \pm 0.004$ & $0.075 \pm 0.004$ & $0.119 \pm 0.004$ & $0.084 \pm 0.005$ & $0.064 \pm 0.003$ & $0.089 \pm 0.005$ & $0.064 \pm 0.002$ & $0.067 \pm 0.003$ & $\bf 0.062 \pm 0.003$ \\
					& Ap ($\uparrow$) & $0.882 \pm 0.008$ & $0.874 \pm 0.007$ & $0.834 \pm 0.006$ & $0.858 \pm 0.007$ & $0.887 \pm 0.004$ & $0.875 \pm 0.006$ & $0.893 \pm 0.006$ & $0.885 \pm 0.005$ & $\bf 0.895 \pm 0.006$ \\
					\hline
					\multirow{6}{*}{yeast} & Hal ($\downarrow$) & $0.203 \pm 0.004$ & $0.207 \pm 0.007$ & $0.188 \pm 0.003$ & $0.195 \pm 0.003$ & $0.188 \pm 0.003$ & $0.195 \pm 0.003$ & $0.192 \pm 0.004$ & $0.194 \pm 0.004$ & $\bf 0.187 \pm 0.004$ \\
					& Sa ($\uparrow$)  & $0.156 \pm 0.012$ & $0.179 \pm 0.009$ & $0.190 \pm 0.009$ & $0.177 \pm 0.013$ & $0.194 \pm 0.009$ & $\bf 0.248 \pm 0.006$ & $0.179 \pm 0.006$ & $0.172 \pm 0.014$ & $0.192 \pm 0.009$ \\
					& F1e ($\uparrow$) & $0.632 \pm 0.007$ & $0.643 \pm 0.009$ & $0.623 \pm 0.006$ & $0.615 \pm 0.012$ & $0.625 \pm 0.006$ & $\bf 0.647 \pm 0.005$ & $0.630 \pm 0.007$ & $0.607 \pm 0.011$ & $0.628 \pm 0.007$ \\
					& Ral ($\downarrow$) & $0.170 \pm 0.005$ & $0.172 \pm 0.005$ & $0.308 \pm 0.008$ & $0.170 \pm 0.005$ & $0.158 \pm 0.005$ & $0.244 \pm 0.008$ & $0.158 \pm 0.006$ & $0.166 \pm 0.005$ & $\bf 0.157 \pm 0.005$ \\
					& Cov ($\downarrow$) & $0.446 \pm 0.006$ & $0.458 \pm 0.006$ & $0.627 \pm 0.007$ & $0.451 \pm 0.009$ & $\bf 0.436 \pm 0.006$ & $0.543 \pm 0.005$ & $0.445 \pm 0.006$ & $0.452 \pm 0.006$ & $\bf 0.436 \pm 0.007$ \\
					& Ap ($\uparrow$) & $0.755 \pm 0.005$ & $0.765 \pm 0.006$ & $0.680 \pm 0.007$ & $0.762 \pm 0.005$ & $0.773 \pm 0.008$ & $0.727 \pm 0.004$ & $0.775 \pm 0.009$ & $0.769 \pm 0.008$ & $\bf 0.777 \pm 0.005$ \\
					\hline
					\multirow{6}{*}{enron} & Hal ($\downarrow$) & $0.051 \pm 0.003$ & $0.061 \pm 0.002$ & $0.052 \pm 0.001$ & $0.054 \pm 0.001$ & $0.050 \pm 0.001$ & $0.052 \pm 0.001$ & $0.049 \pm 0.001$ & $\bf 0.046 \pm 0.001$ & $\bf 0.046 \pm 0.001$ \\
					& Sa ($\uparrow$)  & $0.119 \pm 0.033$ & $0.060 \pm 0.007$ & $0.128 \pm 0.013$ & $0.057 \pm 0.015$ & $0.131 \pm 0.014$ & $\bf 0.154 \pm 0.010$ & $0.128 \pm 0.006$ & $0.124 \pm 0.014$ & $0.137 \pm 0.014$ \\
					& F1e ($\uparrow$) & $0.563 \pm 0.026$ & $0.463 \pm 0.022$ & $0.529 \pm 0.010$ & $0.401 \pm 0.029$ & $0.552 \pm 0.011$ & $0.551 \pm 0.009$ & $\bf 0.585 \pm 0.007$ & $0.538 \pm 0.012$ & $0.569 \pm 0.009$ \\
					& Ral ($\downarrow$) & $0.081 \pm 0.008$ & $0.098 \pm 0.005$ & $0.298 \pm 0.007$ & $0.096 \pm 0.003$ & $\bf 0.071 \pm 0.003$ & $0.208 \pm 0.006$ & $0.078 \pm 0.003$ & $0.076 \pm 0.004$ & $0.072 \pm 0.003$ \\
					& Cov ($\downarrow$) & $0.235 \pm 0.021$ & $0.267 \pm 0.014$ & $0.580 \pm 0.012$ & $0.260 \pm 0.006$ & $\bf 0.210 \pm 0.007$ & $0.472 \pm 0.012$ & $0.232 \pm 0.006$ & $0.228 \pm 0.010$ & $0.214 \pm 0.008$ \\
					& Ap ($\uparrow$) & $0.672 \pm 0.025$ & $0.630 \pm 0.011$ & $0.482 \pm 0.010$ & $0.614 \pm 0.010$ & $0.705 \pm 0.010$ & $0.592 \pm 0.008$ & $0.702 \pm 0.010$ & $0.705 \pm 0.008$ & $\bf 0.709 \pm 0.007$ \\
					\hline
					\multirow{6}{*}{arts} & Hal ($\downarrow$) & $0.061 \pm 0.001$ & $0.106 \pm 0.008$ & $\bf 0.054 \pm 0.008$ & $0.061 \pm 0.001$ & $0.055 \pm 0.001$ & $0.057 \pm 0.001$ & $0.059 \pm 0.001$ & $0.058 \pm 0.001$ & $0.059 \pm 0.001$ \\
					& Sa ($\uparrow$)  & $0.274 \pm 0.006$ & $0.098 \pm 0.030$ & $0.241 \pm 0.009$ & $0.051 \pm 0.008$ & $0.237 \pm 0.008$ & $0.319 \pm 0.008$ & $0.339 \pm 0.010$ & $0.242 \pm 0.008$ & $\bf 0.348 \pm 0.007$ \\
					& F1e ($\uparrow$) & $0.411 \pm 0.006$ & $0.438 \pm 0.009$ & $0.333 \pm 0.009$ & $0.073 \pm 0.009$ & $0.335 \pm 0.006$ & $0.427 \pm 0.006$ & $0.462 \pm 0.009$ & $0.361 \pm 0.008$ & $\bf 0.465 \pm 0.008$ \\
					& Ral ($\downarrow$) & $0.109 \pm 0.002$ & $0.116 \pm 0.004$ & $0.344 \pm 0.006$ & $0.154 \pm 0.004$ & $0.112 \pm 0.002$ & $0.261 \pm 0.007$ & $0.108 \pm 0.003$ & $0.141 \pm 0.004$ & $\bf 0.106 \pm 0.002$ \\
					& Cov ($\downarrow$) & $0.166 \pm 0.004$ & $0.181 \pm 0.006$ & $0.432 \pm 0.007$ & $0.211 \pm 0.006$ & $0.170 \pm 0.005$ & $0.352 \pm 0.007$ & $0.165 \pm 0.004$ & $0.211 \pm 0.006$ & $\bf 0.164 \pm 0.004$ \\
					& Ap ($\uparrow$) & $0.617 \pm 0.003$ & $0.616 \pm 0.007$ & $0.445 \pm 0.006$ & $0.505 \pm 0.008$ & $0.618 \pm 0.004$ & $0.557 \pm 0.007$ & $0.629 \pm 0.008$ & $0.597 \pm 0.007$ & $\bf 0.635 \pm 0.005$ \\
					\hline
					\multirow{6}{*}{education} & Hal ($\downarrow$) & $0.046 \pm 0.001$ & $0.079 \pm 0.016$ & $\bf 0.038 \pm 0.000$ & $0.040 \pm 0.001$ & $0.039 \pm 0.001$ & $0.040 \pm 0.000$ & $0.043 \pm 0.001$ & $0.041 \pm 0.001$ & $0.042 \pm 0.001$ \\
					& Sa ($\uparrow$)  & $0.249 \pm 0.007$ & $0.096 \pm 0.052$ & $0.250 \pm 0.005$ & $0.153 \pm 0.010$ & $0.227 \pm 0.037$ & $0.315 \pm 0.008$ & $0.340 \pm 0.005$ & $0.249 \pm 0.010$ & $\bf 0.349 \pm 0.011$ \\
					& F1e ($\uparrow$) & $0.412 \pm 0.007$ & $0.430 \pm 0.020$ & $0.335 \pm 0.005$ & $0.192 \pm 0.013$ & $0.303 \pm 0.059$ & $0.404 \pm 0.005$ & $0.451 \pm 0.008$ & $0.360 \pm 0.009$ & $\bf 0.461 \pm 0.010$ \\
					& Ral ($\downarrow$) & $0.072 \pm 0.002$ & $0.081 \pm 0.001$ & $0.458 \pm 0.005$ & $0.083 \pm 0.002$ & $0.081 \pm 0.019$ & $0.358 \pm 0.006$ & $\bf 0.070 \pm 0.002$ & $0.105 \pm 0.003$ & $\bf 0.070 \pm 0.001$ \\
					& Cov ($\downarrow$) & $0.100 \pm 0.002$ & $0.148 \pm 0.003$ & $0.518 \pm 0.005$ & $0.110 \pm 0.001$ & $0.110 \pm 0.019$ & $0.424 \pm 0.003$ & $0.101 \pm 0.003$ & $0.148 \pm 0.006$ & $\bf 0.099 \pm 0.002$ \\
					& Ap ($\uparrow$) & $0.613 \pm 0.006$ & $0.578 \pm 0.015$ & $0.376 \pm 0.004$ & $0.587 \pm 0.006$ & $0.605 \pm 0.055$ & $0.503 \pm 0.007$ & $0.635 \pm 0.006$ & $0.612 \pm 0.006$ & $\bf 0.643 \pm 0.005$ \\
					\hline
					\multirow{6}{*}{recreation} & Hal ($\downarrow$) & $0.062 \pm 0.001$ & $0.093 \pm 0.020$ & $\bf 0.054 \pm 0.001$ & $0.062 \pm 0.000$ & $0.055 \pm 0.001$ & $0.057 \pm 0.001$ & $0.063 \pm 0.001$ & $0.058 \pm 0.001$ & $0.061 \pm 0.001$ \\
					& Sa ($\uparrow$)  & $0.280 \pm 0.009$ & $0.195 \pm 0.019$ & $0.274 \pm 0.009$ & $0.062 \pm 0.005$ & $0.272 \pm 0.009$ & $0.365 \pm 0.010$ & $0.381 \pm 0.010$ & $0.270 \pm 0.005$ & $\bf 0.397 \pm 0.008$ \\
					& F1e ($\uparrow$) & $0.394 \pm 0.008$ & $0.410 \pm 0.076$ & $0.342 \pm 0.011$ & $0.070 \pm 0.006$ & $0.345 \pm 0.009$ & $0.438 \pm 0.009$ & $0.466 \pm 0.012$ & $0.352 \pm 0.008$ & $\bf 0.483 \pm 0.007$ \\
					& Ral ($\downarrow$) & $0.122 \pm 0.004$ & $0.135 \pm 0.011$ & $0.393 \pm 0.008$ & $0.194 \pm 0.004$ & $0.125 \pm 0.002$ & $0.293 \pm 0.007$ & $0.123 \pm 0.005$ & $0.151 \pm 0.003$ & $\bf 0.120 \pm 0.003$ \\
					& Cov ($\downarrow$) & $\bf 0.162 \pm 0.005$ & $0.181 \pm 0.014$ & $0.452 \pm 0.007$ & $0.231 \pm 0.005$ & $0.167 \pm 0.003$ & $0.357 \pm 0.008$ & $0.167 \pm 0.004$ & $0.198 \pm 0.004$ & $0.164 \pm 0.004$ \\
					& Ap ($\uparrow$) & $0.619 \pm 0.007$ & $0.610 \pm 0.018$ & $0.451 \pm 0.009$ & $0.450 \pm 0.010$ & $0.627 \pm 0.008$ & $0.565 \pm 0.006$ & $0.629 \pm 0.009$ & $0.611 \pm 0.004$ & $\bf 0.642 \pm 0.005$ \\
					\hline
					\multirow{6}{*}{science} & Hal ($\downarrow$) & $0.036 \pm 0.001$ & $0.113 \pm 0.004$ & $\bf 0.032 \pm 0.000$ & $0.034 \pm 0.000$ & $\bf 0.032 \pm 0.000$ & $0.033 \pm 0.000$ & $0.037 \pm 0.001$ & $\bf 0.032 \pm 0.000$ & $0.036 \pm 0.001$ \\
					& Sa ($\uparrow$)  & $0.244 \pm 0.007$ & $0.011 \pm 0.005$ & $0.244 \pm 0.008$ & $0.112 \pm 0.010$ & $0.236 \pm 0.007$ & $0.323 \pm 0.005$ & $0.336 \pm 0.010$ & $0.235 \pm 0.007$ & $\bf 0.351 \pm 0.010$ \\
					& F1e ($\uparrow$) & $0.372 \pm 0.011$ & $0.327 \pm 0.004$ & $0.314 \pm 0.086$ & $0.138 \pm 0.012$ & $0.313 \pm 0.007$ & $0.401 \pm 0.004$ & $0.425 \pm 0.011$ & $0.311 \pm 0.010$ & $\bf 0.443 \pm 0.011$ \\
					& Ral ($\downarrow$) & $0.095 \pm 0.004$ & $0.102 \pm 0.002$ & $0.402 \pm 0.008$ & $0.124 \pm 0.004$ & $0.092 \pm 0.003$ & $0.291 \pm 0.004$ & $0.091 \pm 0.002$ & $0.129 \pm 0.002$ & $\bf 0.090 \pm 0.003$ \\
					& Cov ($\downarrow$) & $0.132 \pm 0.005$ & $0.137 \pm 0.003$ & $0.456 \pm 0.008$ & $0.159 \pm 0.004$ & $0.128 \pm 0.003$ & $0.349 \pm 0.006$ & $0.129 \pm 0.003$ & $0.176 \pm 0.002$ & $\bf 0.127 \pm 0.004$ \\
					& Ap ($\uparrow$) & $0.583 \pm 0.019$ & $0.542 \pm 0.005$ & $0.359 \pm 0.009$ & $0.506 \pm 0.009$ & $0.601 \pm 0.007$ & $0.509 \pm 0.005$ & $0.593 \pm 0.009$ & $0.578 \pm 0.003$ & $\bf 0.605 \pm 0.008$ \\
					\hline
					\multirow{6}{*}{business} & Hal ($\downarrow$) & $0.030 \pm 0.004$ & $0.045 \pm 0.001$ & $0.026 \pm 0.001$ & $0.026 \pm 0.001$ & $\bf 0.025 \pm 0.000$ & $\bf 0.025 \pm 0.000$ & $\bf 0.025 \pm 0.001$ & $\bf 0.025 \pm 0.001$ & $\bf 0.025 \pm 0.001$ \\
					& Sa ($\uparrow$)  & $0.454 \pm 0.085$ & $0.160 \pm 0.003$ & $0.565 \pm 0.010$ & $0.552 \pm 0.011$ & $0.557 \pm 0.010$ & $\bf 0.578 \pm 0.011$ & $0.557 \pm 0.008$ & $0.538 \pm 0.006$ & $0.563 \pm 0.007$ \\
					& F1e ($\uparrow$) & $0.734 \pm 0.036$ & $0.658 \pm 0.002$ & $0.763 \pm 0.005$ & $0.760 \pm 0.007$ & $0.768 \pm 0.005$ & $\bf 0.783 \pm 0.005$ & $0.770 \pm 0.006$ & $0.765 \pm 0.005$ & $0.769 \pm 0.005$ \\
					& Ral ($\downarrow$) & $0.034 \pm 0.005$ & $0.036 \pm 0.001$ & $0.205 \pm 0.005$ & $0.036 \pm 0.002$ & $0.032 \pm 0.004$ & $0.151 \pm 0.006$ & $\bf 0.030 \pm 0.002$ & $0.041 \pm 0.002$ & $\bf 0.030 \pm 0.002$ \\
					& Cov ($\downarrow$) & $0.068 \pm 0.005$ & $0.074 \pm 0.001$ & $0.338 \pm 0.009$ & $0.070 \pm 0.002$ & $0.066 \pm 0.005$ & $0.269 \pm 0.007$ & $\bf 0.065 \pm 0.002$ & $0.082 \pm 0.004$ & $\bf 0.065 \pm 0.003$ \\
					& Ap ($\uparrow$) & $0.860 \pm 0.036$ & $0.884 \pm 0.005$ & $0.742 \pm 0.005$ & $0.887 \pm 0.004$ & $\bf 0.895 \pm 0.006$ & $0.790 \pm 0.005$ & $0.891 \pm 0.005$ & $0.885 \pm 0.004$ & $0.890 \pm 0.004$ \\
					\hline
				\end{tabular}
			\end{table}
			
			\begin{table}[!htbp]
				\scriptsize
				\renewcommand{\arraystretch}{1.0}
				\caption{Average ranks of the compared approaches on all datasets in terms of each evaluation metric. Best results are highlighted in bold. (``Overall" denotes the average rank for all the metrics.)}
				\hskip-40pt
				\label{tab:multi_label_result_average_rank}
				\begin{tabular}{c|ccccccccc}
					\hline
					Evaluation
					& \multicolumn{9}{c}{ Average Rank } \\
					\cline{2-10}
					Metrics & Rank-SVM & Rank-SVMz & BR & ML-kNN & CLR & RAKEL & CPNL & MLFE & RBRL \\
					\hline
					Hamming Loss & $7.00$ & $9.00$ & $2.90$ & $6.40$ & $\bf 2.80$ & $3.20$ & $4.50$ & $3.70$ & $2.90$ \\
					Suset Accuracy & $6.30$ & $7.90$ & $4.20$ & $8.10$ & $4.90$ & $2.20$ & $2.80$ & $6.40$ & $\bf 1.60$ \\
					F1-Example & $5.10$ & $4.40$ & $7.00$ & $8.60$ & $6.00$ & $3.00$ & $\bf 1.90$ & $6.80$ & $2.10$ \\
					Ranking Loss & $4.20$ & $5.10$ & $9.00$ & $6.40$ & $3.40$ & $7.90$ & $2.00$ & $5.00$ & $\bf 1.20$ \\
					Coverage  & $4.00$ & $5.70$ & $9.00$ & $6.10$ & $2.80$ & $7.90$ & $2.30$ & $5.10$ & $\bf 1.30$ \\
					Average Precision & $5.10$ & $5.50$ & $8.90$ & $7.20$ & $3.00$ & $7.30$ & $2.10$ & $4.10$ & $\bf 1.30$ \\
					\hline
					Overall    & $5.28$ & $6.27$ & $6.83$ & $7.13$ & $3.82$ & $5.25$ & $2.60$ & $5.18$ & $\bf 1.73$ \\
					\hline
				\end{tabular}
			\end{table}	
			
			Furthermore, to systematically conduct a performance analysis among different compared approaches, \emph{Friedman test} \cite{friedman1937use,demvsar2006statistical} is utilized to carry out a nonparametric statistical analysis. The results of the \emph{Friedman test} on each evaluation metric are summarized in Table \ref{tab:multi_label_friedman_test}. As shown in Table \ref{tab:multi_label_friedman_test}, at significance level $\alpha = 0.05$, the null hypothesis that each compared approach has ``equal" performance is clearly rejected on each evaluation metric. Therefore, we can continue on a certain \emph{post-hoc test} \cite{demvsar2006statistical} to further analyze the relative performance among the compared approaches.
			
\begin{figure}[!htb]
				\centering
				\subfigure[HammingLoss]{
					\label{fig:cd_multi_label:hammingloss} 
					\includegraphics[width=2.3in]{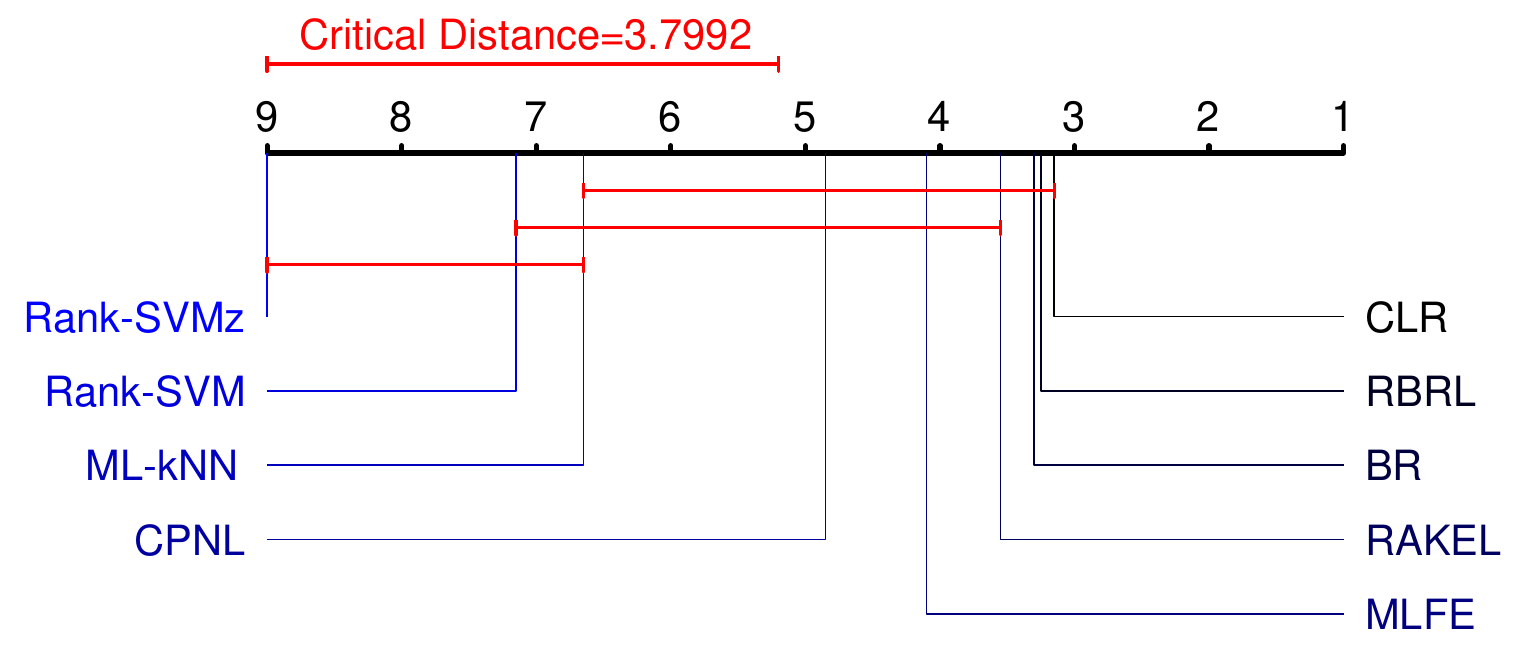}}
				\subfigure[SubsetAccuracy]{
					\label{fig:cd_multi_label:subsetaccuracy} 
					\includegraphics[width=2.3in]{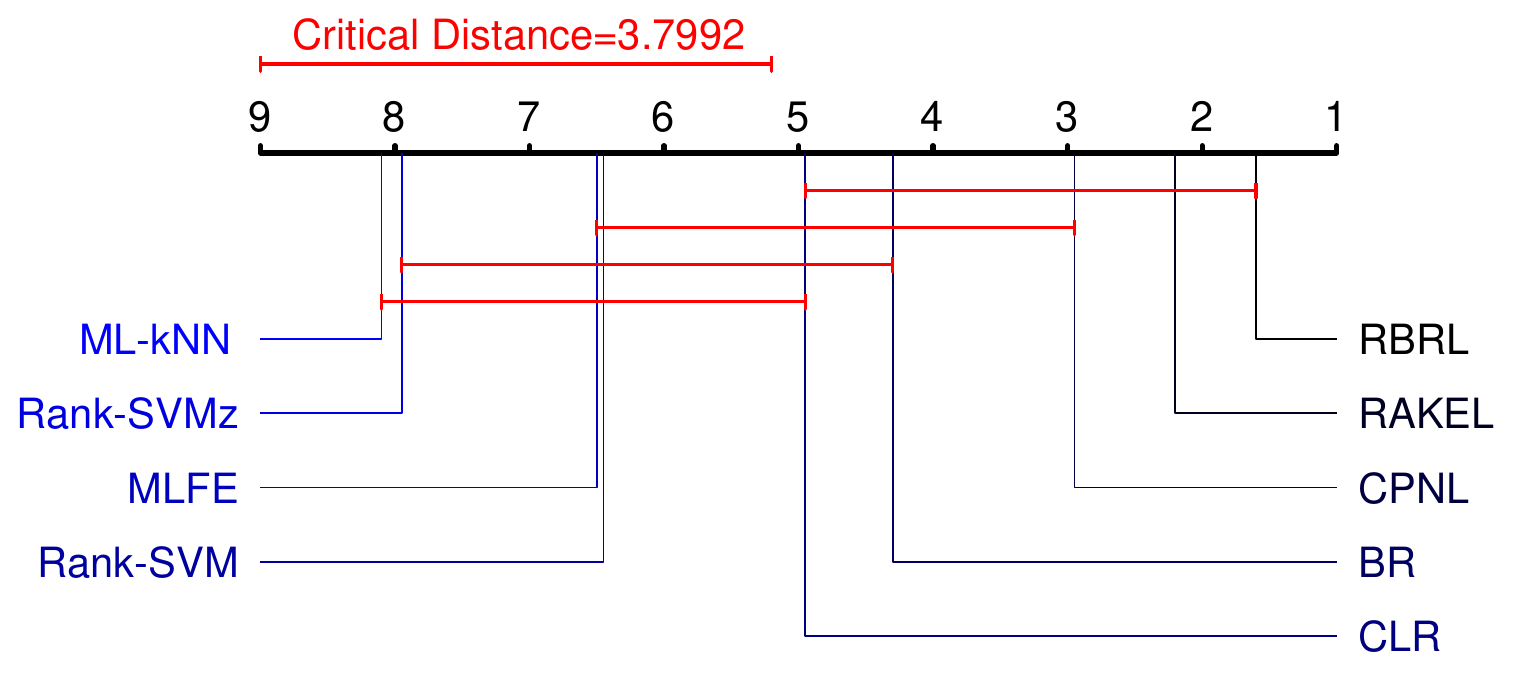}}
				\subfigure[F1-Example]{
					\label{fig:cd_multi_label:f1example} 
					\includegraphics[width=2.3in]{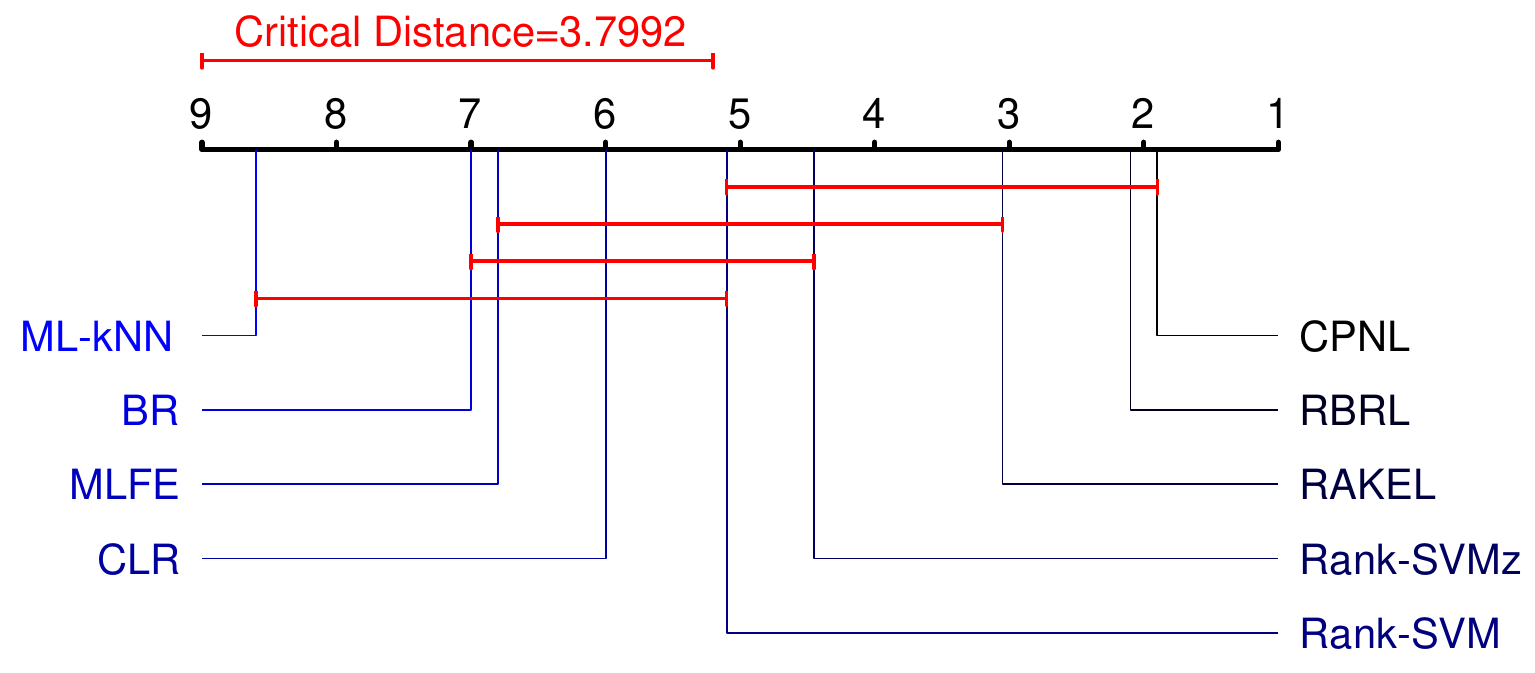}}
				\subfigure[RankingLoss]{
					\label{fig:cd_multi_label:rankingloss} 
					\includegraphics[width=2.3in]{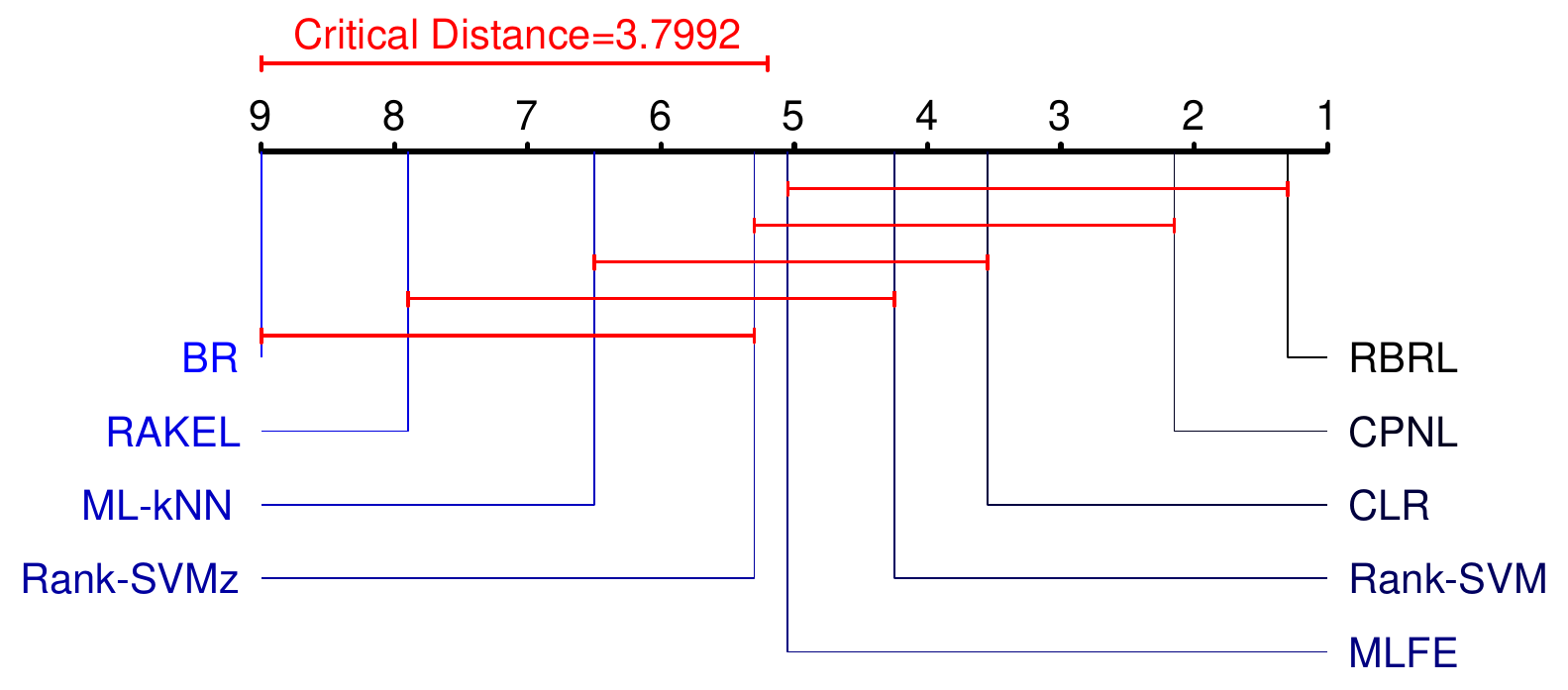}}
				\subfigure[Coverage]{
					\label{fig:cd_multi_label:coverage} 
					\includegraphics[width=2.3in]{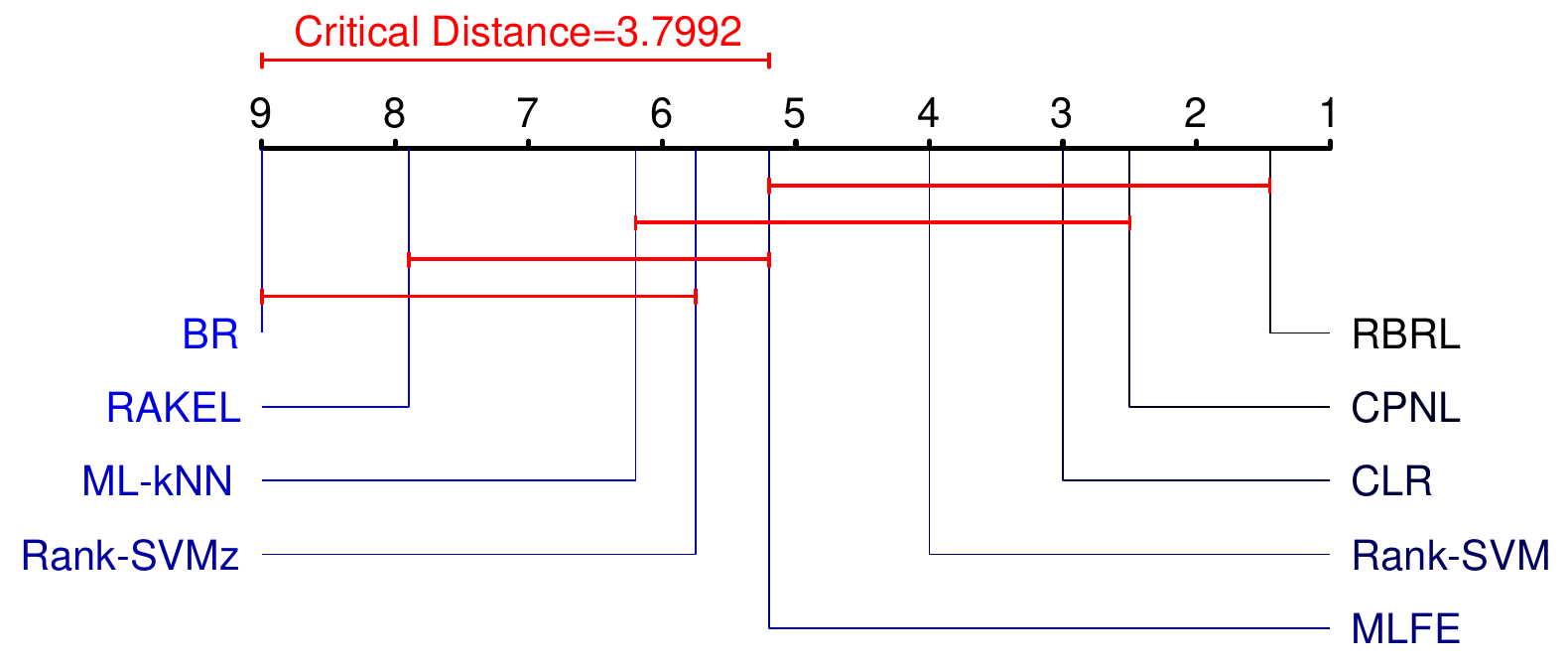}}
				\subfigure[AveragePrecision]{
					\label{fig:cd_multi_label:averageprecision} 
					\includegraphics[width=2.3in]{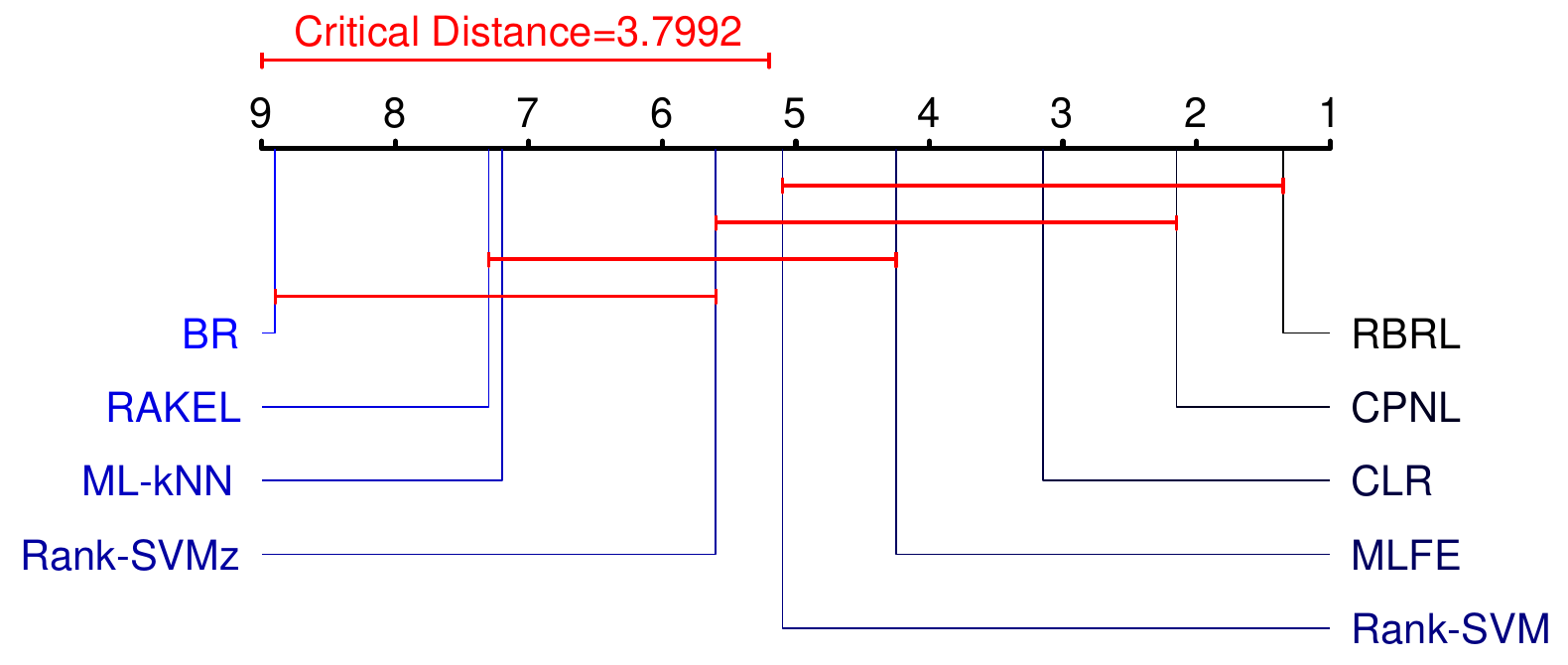}}
				\caption{Comparison of pairwise approaches with the \emph{Nemenyi test} on each evaluation metric.}
				\label{fig:nemenyi_multi_label} 
			\end{figure}

			\emph{Nemenyi test} \cite{demvsar2006statistical}, which makes pairwise performance tests, is utilized to test whether one approach obtains a competitive performance against the other compared approaches. If the average ranks of pairwise approaches differ by at least one critical distance (CD) value, they are viewed to have significantly different performance. For Nemenyi test, at significance level $\alpha = 0.05$, we have $q_\alpha = 3.102$, and thus ${\rm CD} = q_\alpha \sqrt{k(k+1) / 6N} = 3.7992 \ (k = 9, N = 10)$. To visually illustrate the relative performance of the compared approaches, Figure \ref{fig:nemenyi_multi_label} shows the CD diagrams on each evaluation metric. In each subfigure, each pairwise approaches are interconnected with a red line when their average ranks are within one CD. Otherwise, any approaches not interconnected are believed to have significantly different performance among them. 
			\begin{table}[!htb]
				\scriptsize
				\renewcommand{\arraystretch}{1.0}
				\centering
				\caption{Summary of Friedman Statistics $F_F(k = 9, N = 10)$ and the Critical Value in terms of each metric $(k: \# $ Compared Approaches; $N: \#$ {Data sets})}
				\label{tab:multi_label_friedman_test}
				\begin{tabular}{lcc}
					\hline
					Metric & $F_F$ & critical value$(\alpha = 0.05)$ \\
					\hline
					HammingLoss & $ 13.1812 $ &
					\multirow{5}{*}{$ 2.0698 $} \\
					SubsetAccuracy & $ 31.7885 $ &  \\
					F1-Example & $ 23.4032 $ &  \\
					RankingLoss & $ 52.5736 $ &  \\
					Coverage & $ 46.6988 $ &  \\
					AveragePrecision & $ 51.9481 $ &  \\
					\hline
				\end{tabular}
			\end{table}
			
			Our proposed method RBRL can be viewed to minimize the Hamming Loss and Ranking Loss simultaneously. According to these experimental results, we can achieve the following observations.  
			\begin{enumerate}[(1)]
				\item In comparison with Rank-SVM, RBRL achieves better performance in terms of classification-based metrics (i.e., Hamming Loss, Subset Accuracy and F1-Example) mainly because RBRL explicitly minimizes the (approximate) Hamming Loss and is learned in only one step which doesn’t have the \emph{error accumulation} issue. Besides, RBRL experimentally outperforms Rank-SVM in terms of ranking-based metrics (i.e., Ranking Loss, Coverage and Average Precision), which is interesting. We argue this is because the first term of minimization of the (approximate) Hamming Loss and the low-rank constraint term can be viewed as regularizers to prune the hypothesis set to improve the generalization performance in Ranking Loss.
				\item In comparison with Rank-SVMz, RBRL achieves better performance in terms of all the metrics, especially in Hamming Loss and Subset Accuracy. It's mainly because RBRL explicitly minimizes the (approximate) Hamming Loss and has smaller complexity of the model (i.e., the hypothesis set), which generally makes it easy to control the generalization error.
				\item For ranking-based metrics (i.e., Ranking Loss, Coverage and Average Precision), RBRL obtains better performance over other approaches, especially BR. It’s because, in comparison with BR, RBRL explicitly minimizes the (approximate) ranking loss and further exploits label correlations under the low-rank constraint.
				\item RBRL obtains comparable performance on Subset Accuracy against RAKEL which is a high-order approach and tries to optimize Subset Accuracy. Besides, RBRL statistically outperforms CLR on Subset Accuracy.
				\item RBRL statistically outperforms the recent approach MLFE on five evaluation metrics (i.e., Subset Accuracy, F1-Example, Ranking Loss, Coverage and Average Precision).
				\item RBRL obtains highly competitive or superior performance on five evaluation metrics (i.e., Hamming Loss, Subset Accuracy, Ranking Loss, Coverage and Average Precision) over CPNL which is an extension of BR. Besides, our proposed method RBRL can also be viewed as an extension of BR, which additionally considers the minimization of the Ranking Loss and further explores the label correlations under the low-rank constraint. The better performance of RBRL confirms the effectiveness of minimization of the ranking loss and low-rank label correlations.
				\item As Table \ref{tab:multi_label_result_average_rank} and Figure \ref{fig:averagerank_multi_label} shown, RBRL obtains better performance than other approaches on the overall metrics.
			\end{enumerate}
			
			All in all, RBRL obtains highly comparable or superior performance over several state-of-the-art approaches for MLC.

		\subsubsection{Validation of the Ranking Loss Term}
			Since previous work \cite{xu2014learning, yu2014large, jing2015semi, xu2016local} has shown the effectiveness of the low-rank constraint for MLC, here we focus on validating the effectiveness of the (approximate) ranking loss minimization term. For the sake of fairness, we consider a degenerative version of the model RBRL without the ranking loss minimization term (i.e., $\lambda_2 = 0$), which is named BRL. Besides, for simplicity, we evaluate the performance of BRL and RBRL on two representative datasets (i.e., \emph{emotions} and \emph{arts}), and the kernel models are evaluated on the \emph{emotions} dataset and the linear models are evaluated on the \emph{arts} dataset. From Table \ref{tab:multi_label_result_rank_loss_term}, we can observe that RBRL is clearly superior to BRL, which confirms the effectiveness of the ranking loss minimization term.
			\begin{table}[!htbp]
				\scriptsize
				\renewcommand{\arraystretch}{1.0}
				\centering
				\caption{Experimental results of two models (i.e., BRL and RBRL) on the \emph{emotions} (for the kernel model) and \emph{arts} datasets (for the linear model). $\downarrow(\uparrow)$ indicates the smaller (larger) the better. Best results are highlighted in bold.}
				\label{tab:multi_label_result_rank_loss_term}
				\begin{tabular}{c|c|cc}
					\hline
					& Metric & BRL & RBRL \\
					\hline
					\multirow{6}{*}{emotions} & Hal ($\downarrow$) & $0.185 \pm 0.010$ & $\bf 0.181 \pm 0.012$ \\
					& Sa ($\uparrow$)    & $0.319 \pm 0.029$ & $\bf 0.334 \pm 0.032$ \\
					& F1e ($\uparrow$)   & $0.658 \pm 0.019$ & $\bf 0.666 \pm 0.022$ \\
					& Ral ($\downarrow$) & $0.146 \pm 0.010$ & $\bf 0.138 \pm 0.011$ \\
					& Cov ($\downarrow$) & $0.289 \pm 0.010$ & $\bf 0.277 \pm 0.010$ \\
					& Ap ($\uparrow$)    & $0.819 \pm 0.012$ & $\bf 0.828 \pm 0.014$ \\
					\hline
					\multirow{6}{*}{arts} & Hal ($\downarrow$) & $\bf 0.059 \pm 0.001$ & $\bf 0.059 \pm 0.001$ \\
					& Sa ($\uparrow$)    & $0.335 \pm 0.006$ & $\bf 0.348 \pm 0.007$ \\
					& F1e ($\uparrow$)   & $0.457 \pm 0.006$ & $\bf 0.465 \pm 0.008$ \\
					& Ral ($\downarrow$) & $0.112 \pm 0.002$ & $\bf 0.106 \pm 0.002$ \\
					& Cov ($\downarrow$) & $0.173 \pm 0.004$ & $\bf 0.164 \pm 0.004$ \\
					& Ap ($\uparrow$)    & $0.629 \pm 0.003$ & $\bf 0.635 \pm 0.005$ \\
					\hline
				\end{tabular}
			\end{table}
		
		\subsubsection{Convergence Analysis}
\begin{figure}[!htb]
				\centering
				\subfigure[The convergence curve of the APG for the kernel RBRL on the \emph{emotions} dataset.]{
					\label{fig:convergence:kernel} 
					\includegraphics[width=2.3in]{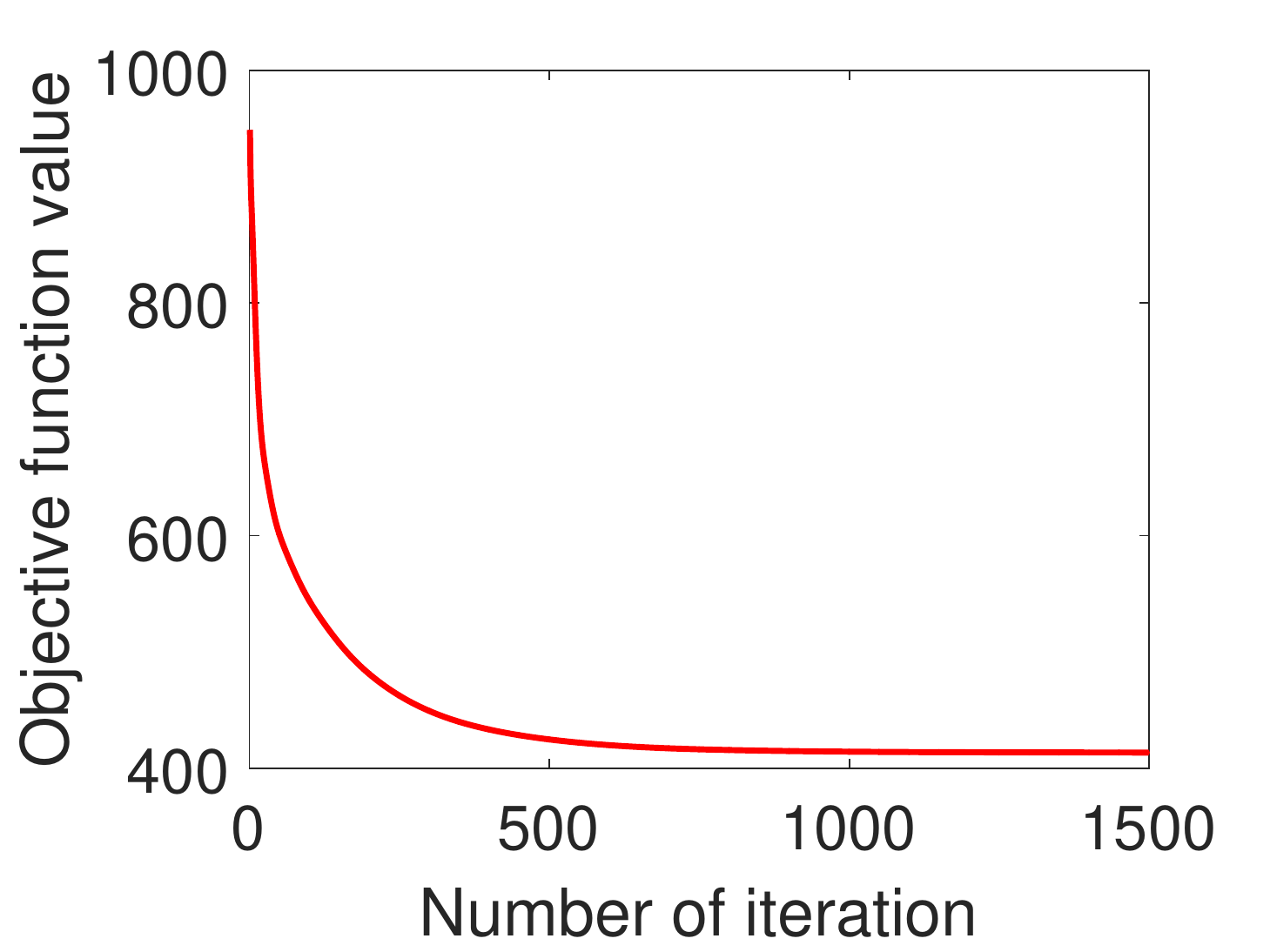}}
				\subfigure[The convergence curve of the APG for the linear RBRL on the \emph{arts} dataset.]{
					\label{fig:convergence:linear} 
					\includegraphics[width=2.3in]{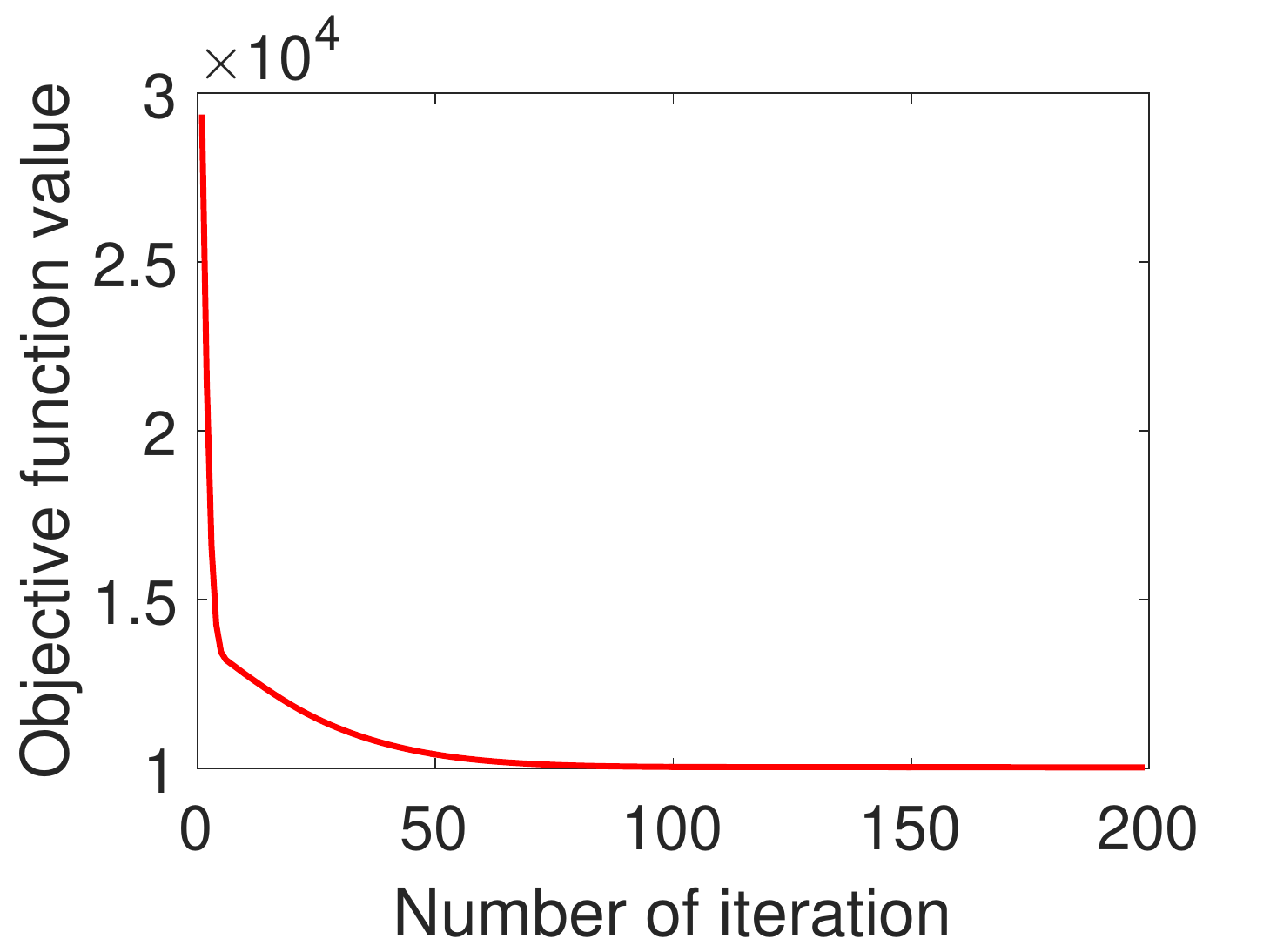}}
				\caption{The convergence curves of the accelerated proximal gradient methods (APG) for the linear and kernel RBRL.}
				\label{fig:convergence} 
			\end{figure}

			To illustrate the quick convergence efficiency of the proposed accelerated proximal gradient methods (APG) for the linear and kernel RBRL, we further plot the convergence curves of the objective function in the case of the number of iteration in Figure \ref{fig:convergence}. We only show the convergence curves on two representative datasets (i.e., \emph{emotions} for the kernel RBRL and \emph{arts} for the linear RBRL) owing to lack of space. From Figure \ref{fig:convergence}, it can be observed that the APG for the linear RBRL converges quickly within a few ($\approx 50$) iterations on the \emph{arts} dataset. In comparison, the APG for the kernel RBRL converges within more ($\approx 500$) iterations on the \emph{emotions} dataset.
		
		\subsubsection{Sensitivity to Hyper-parameters}
\begin{figure}[!htbp]
				\centering
				\subfigure[RBRL]{
					\label{fig:sensitivity_gamma:RBRL} 
					\includegraphics[width=2.3in]{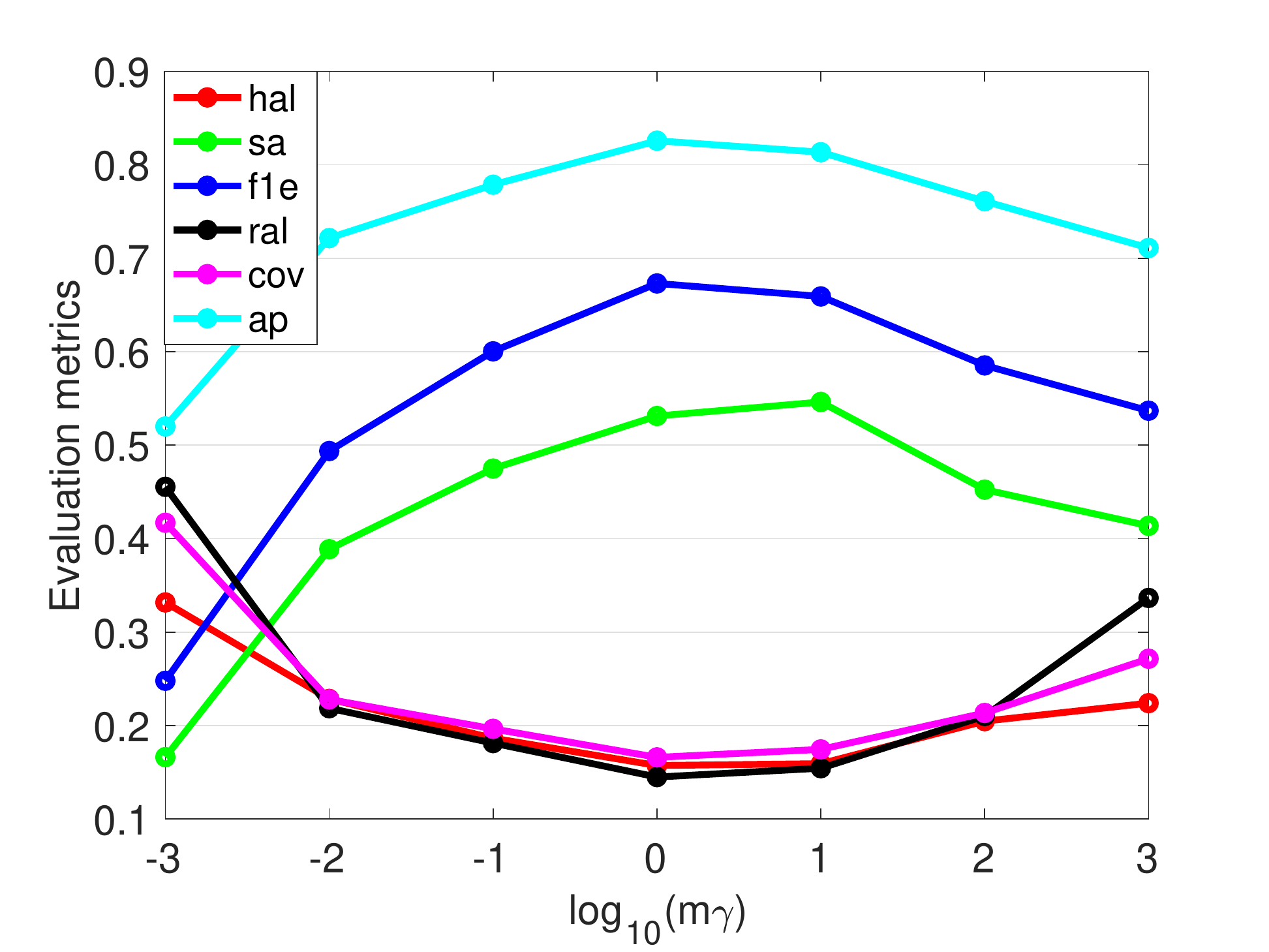}}
				\subfigure[Rank-SVM]{
					\label{fig:sensitivity_gamma:RankSVM} 
					\includegraphics[width=2.3in]{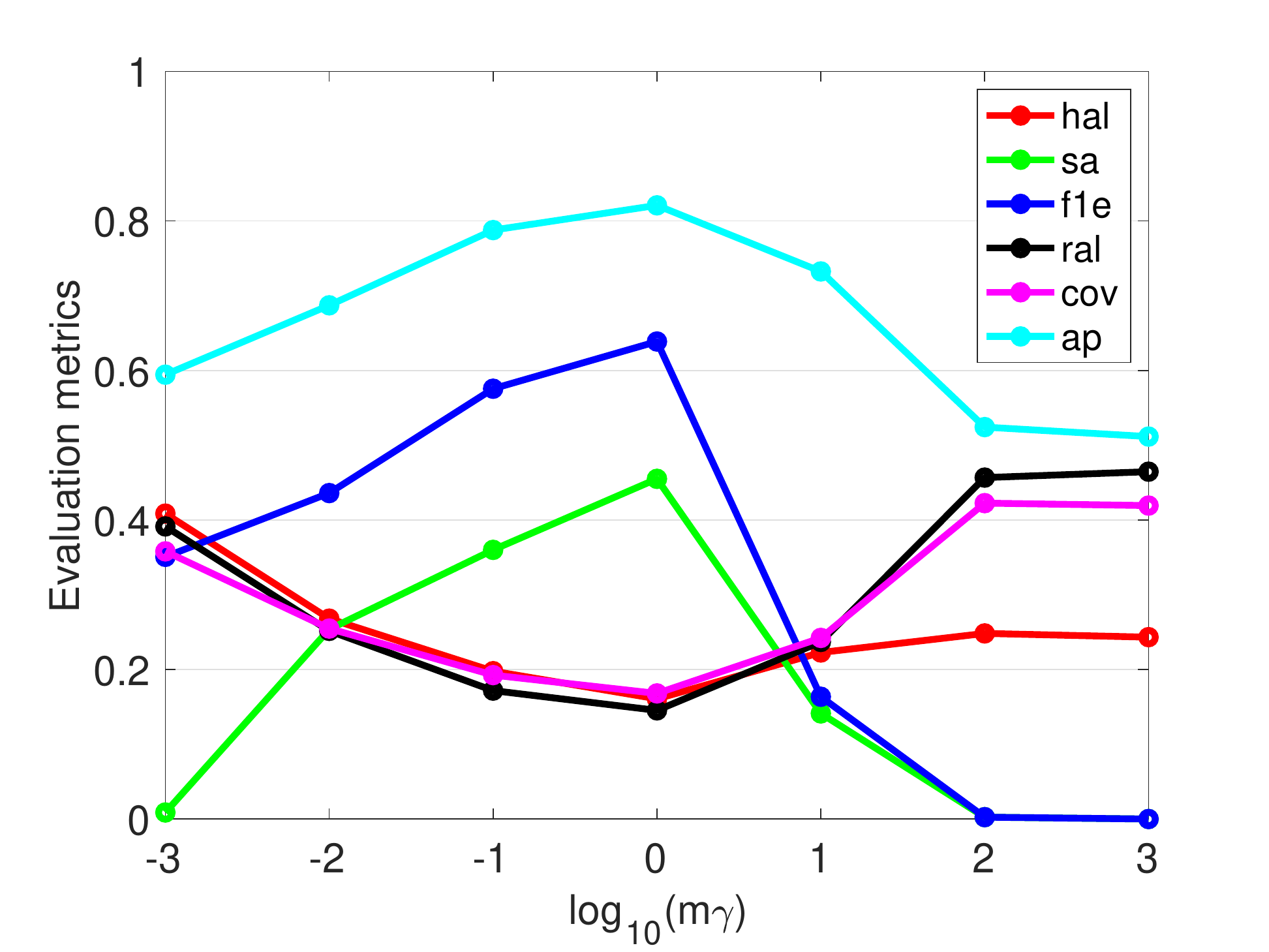}}
				\caption{Sensitivity analysis to the RBF hyper-parameter $\gamma$ of RBRL and Rank-SVM on the \emph{image} dataset. ($m$ is the feature size of the dataset.)}
				\label{fig:sensitivity_gamma}
			\end{figure}

			First, we aim to validate the default setting effectiveness of the RBF kernel hyper-parameter $\gamma$ (i.e., $1/m$). We evaluate the performance effect of $\gamma$ with other hyper-parameters fixed for two representative methods (i.e., RBRL and Rank-SVM) on the \emph{image} dataset. The search range of $\gamma$ is $\{10^{-3}/m, 10^{-2}/m, ..., 10^3/m \}$. As shown in Figure \ref{fig:sensitivity_gamma}, RBRL and Rank-SVM both achieves good performance at the default value $1/m$ although it can improve the performance by fine-tuning the hyper-parameter $\gamma$. Similar phenomena can also be found on the remaining datasets for other methods.
		
\begin{figure}[!htb]
				\centering
				\subfigure[]{
					\label{fig:sensitivity:hal_lambda1} 
					\includegraphics[width=1.5in]{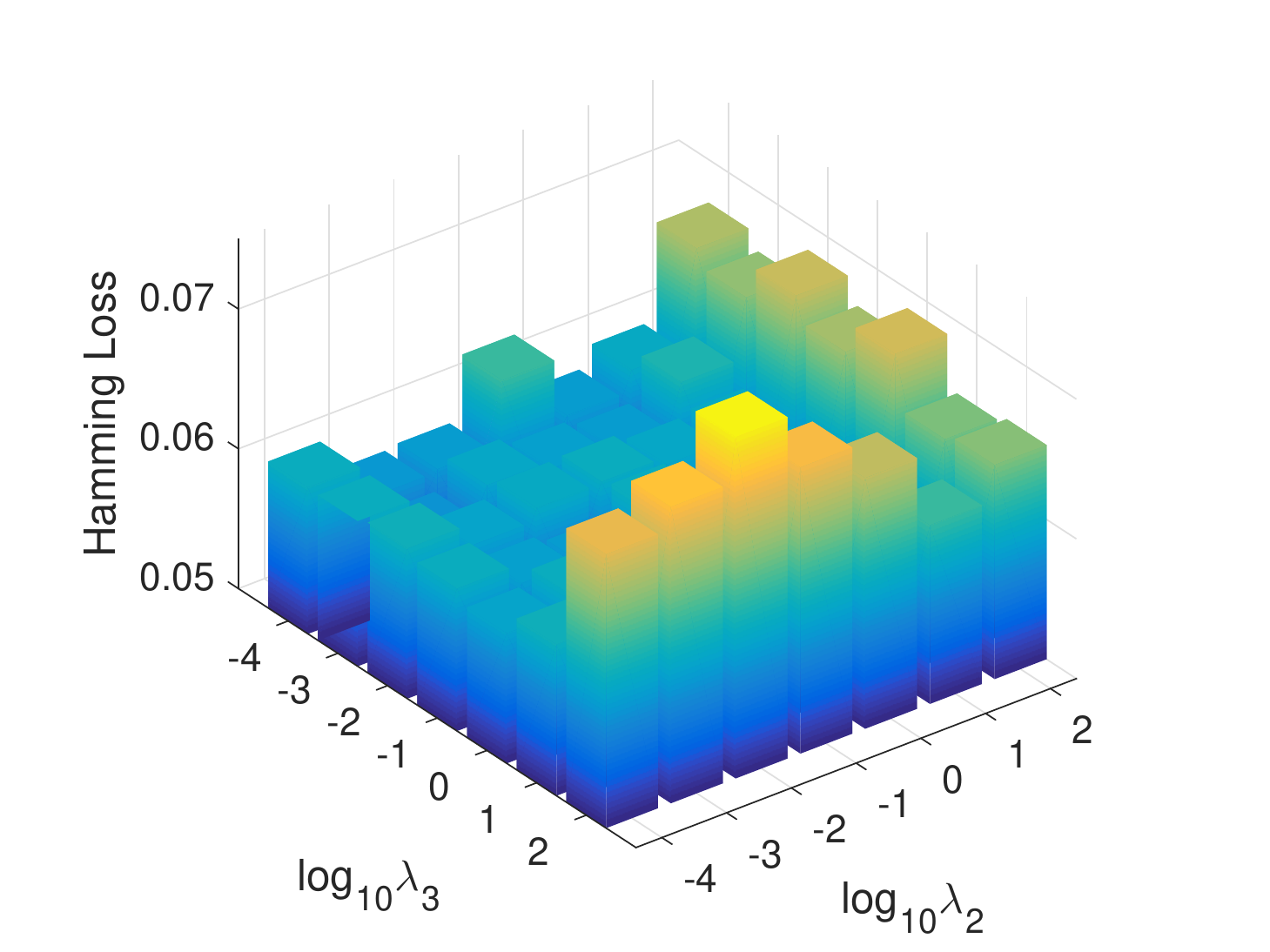}}
				\subfigure[]{
					\label{fig:sensitivity:sa_lambda1} 
					\includegraphics[width=1.5in]{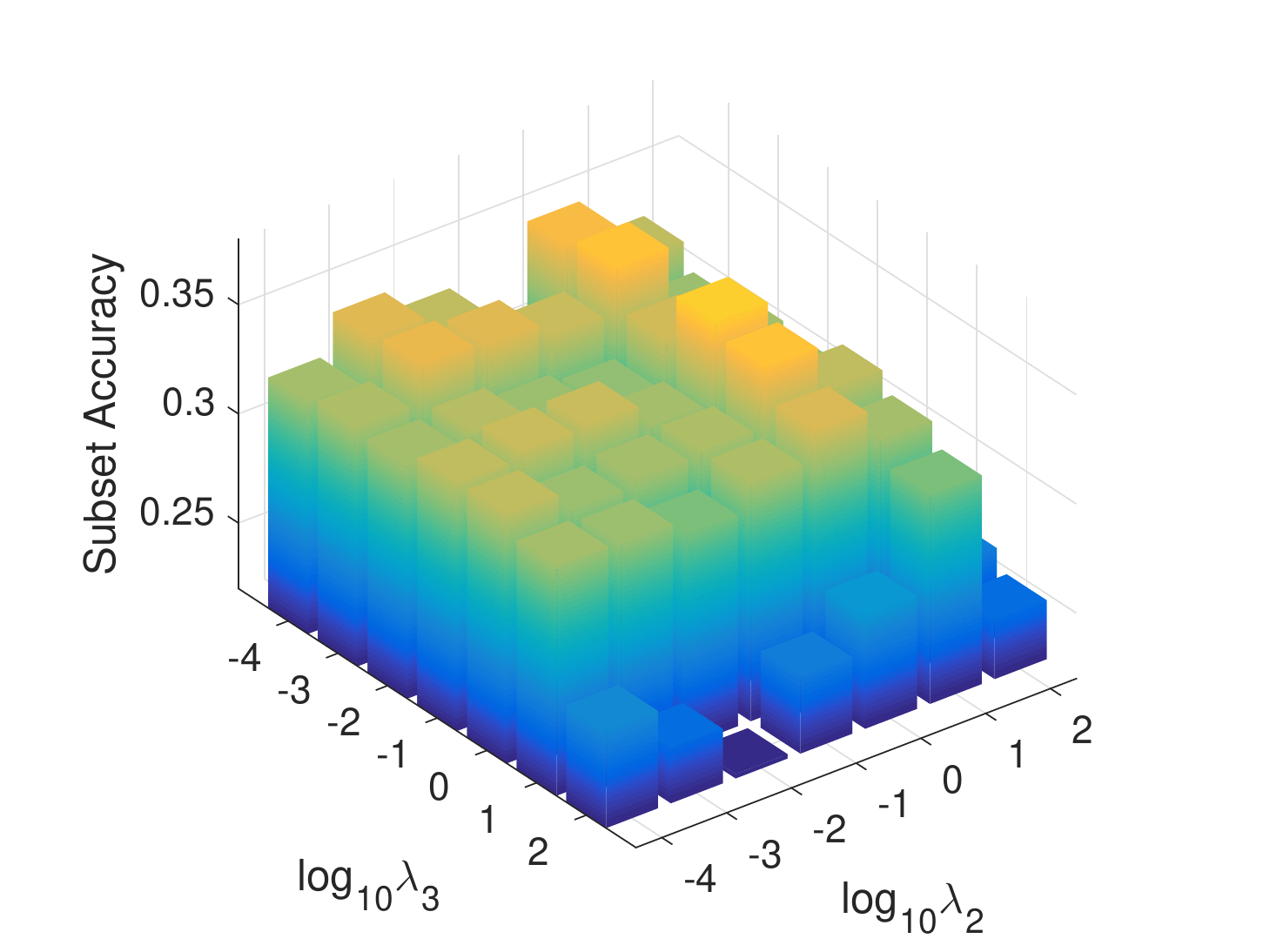}}
				\subfigure[]{
					\label{fig:sensitivity:f1e_lambda1} 
					\includegraphics[width=1.5in]{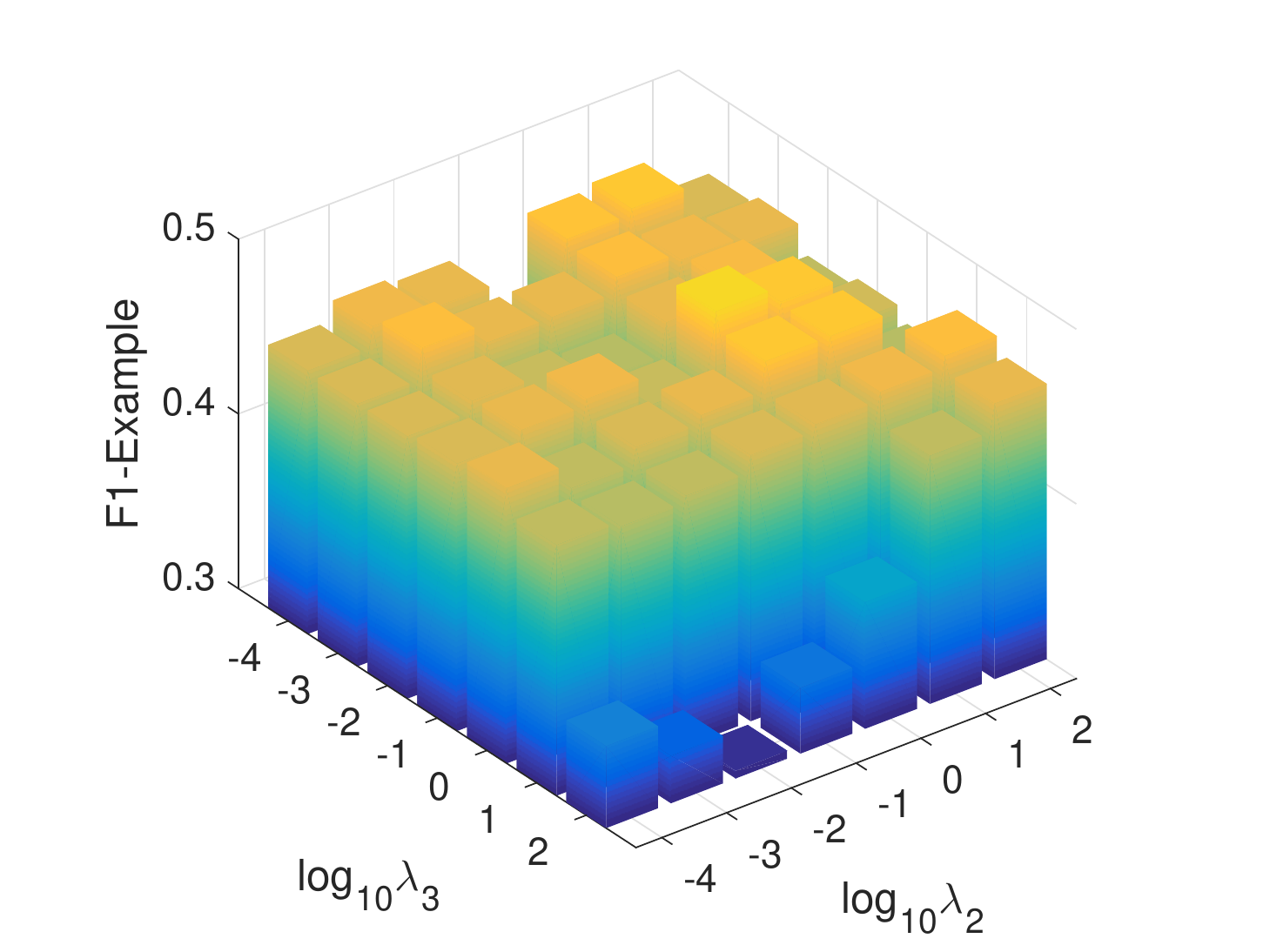}}
				\subfigure[]{
					\label{fig:sensitivity:ral_lambda1} 
					\includegraphics[width=1.5in]{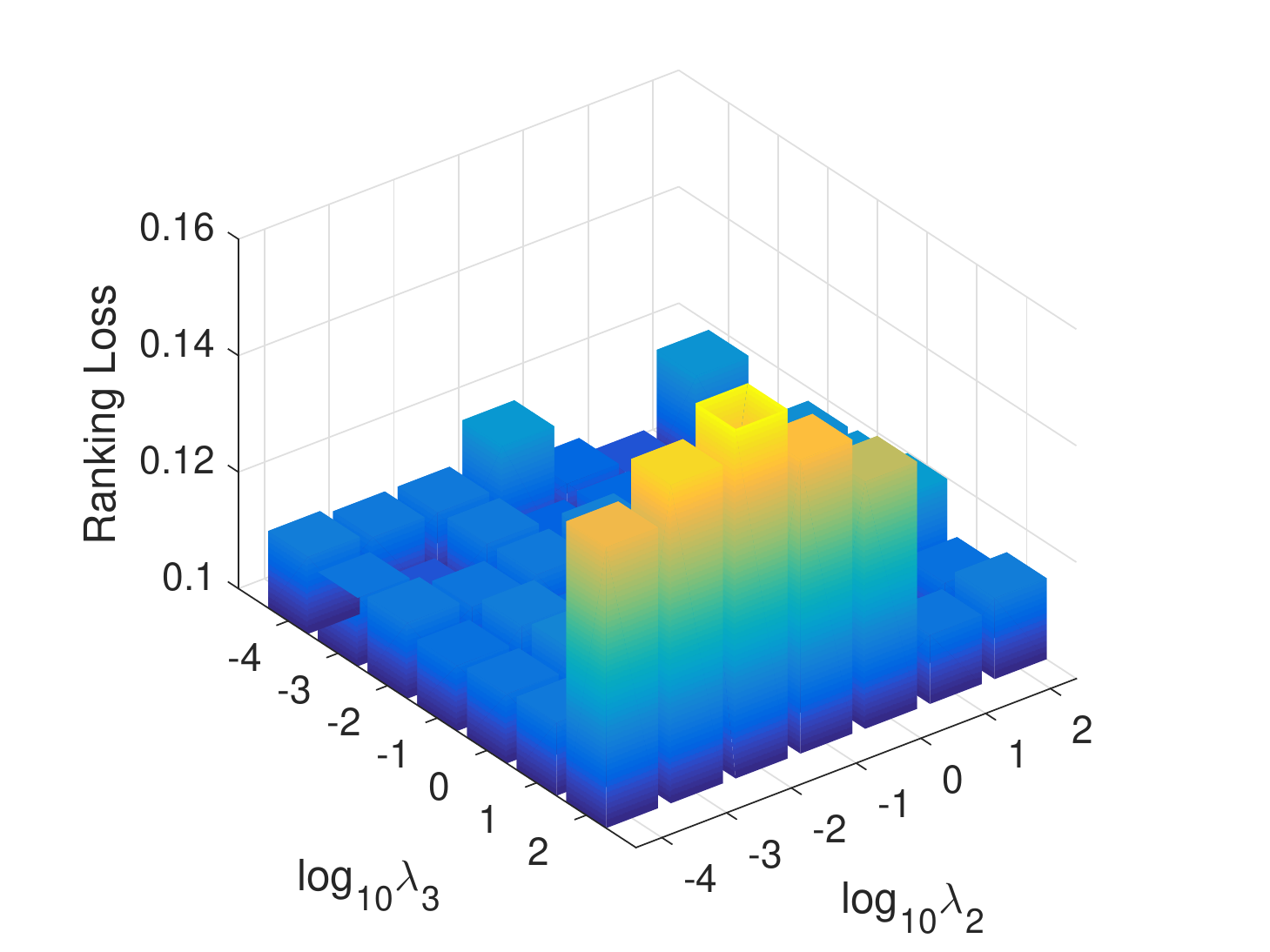}}
				\subfigure[]{
					\label{fig:sensitivity:cov_lambda1} 
					\includegraphics[width=1.5in]{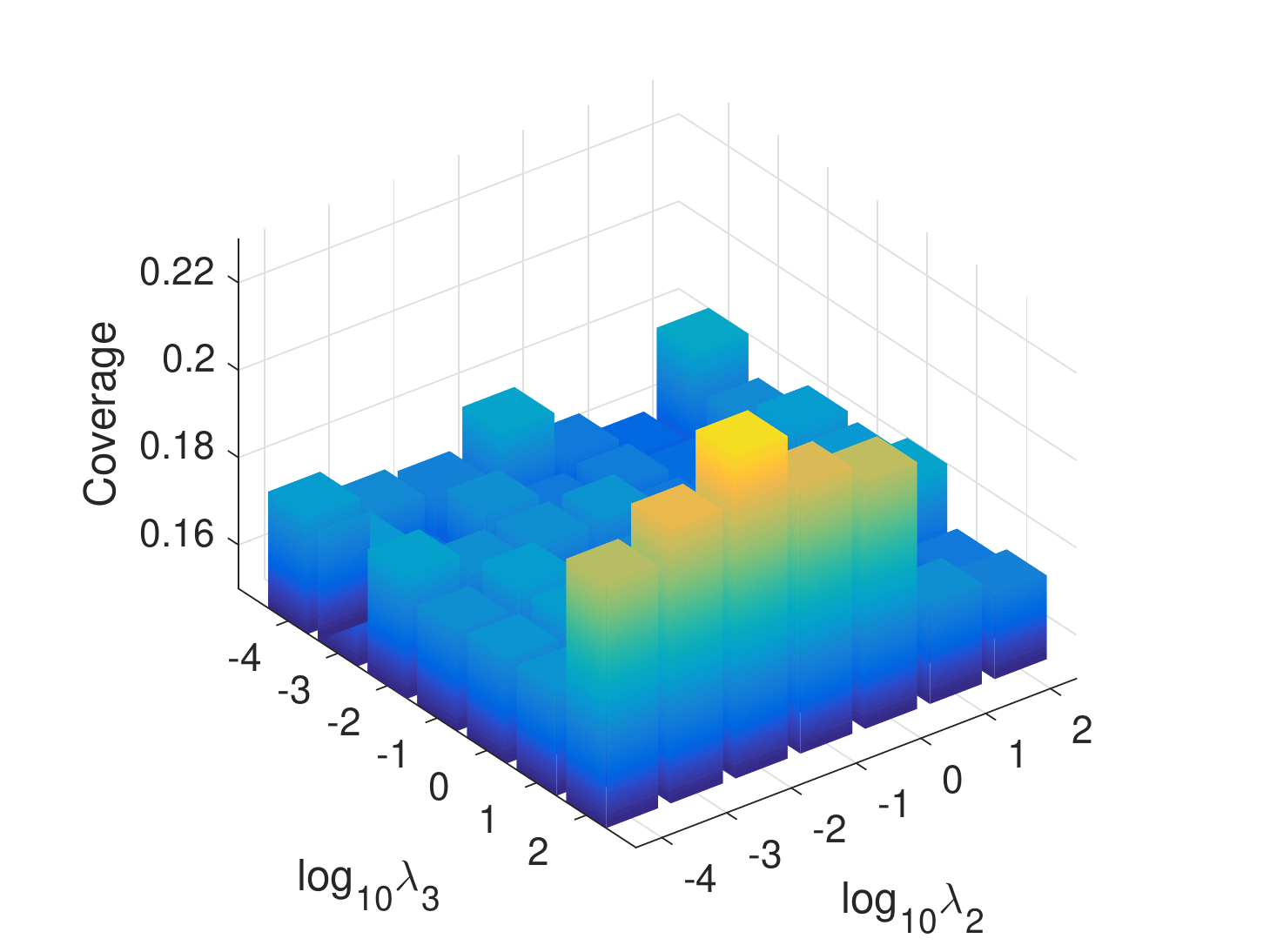}}
				\subfigure[]{
					\label{fig:sensitivity:ap_lambda1} 
					\includegraphics[width=1.5in]{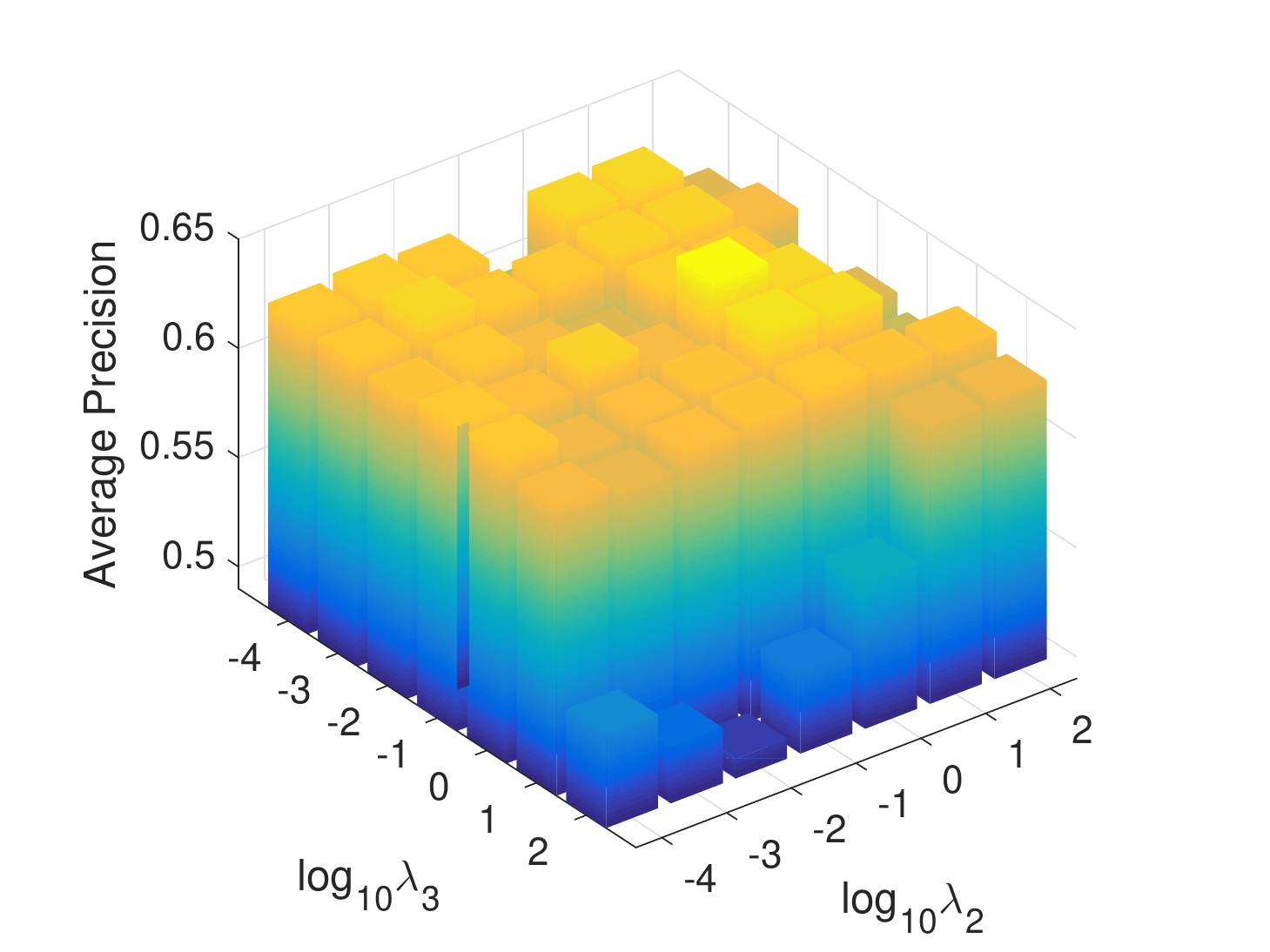}}
				\subfigure[]{
					\label{fig:sensitivity:hal_lambda2} 
					\includegraphics[width=1.5in]{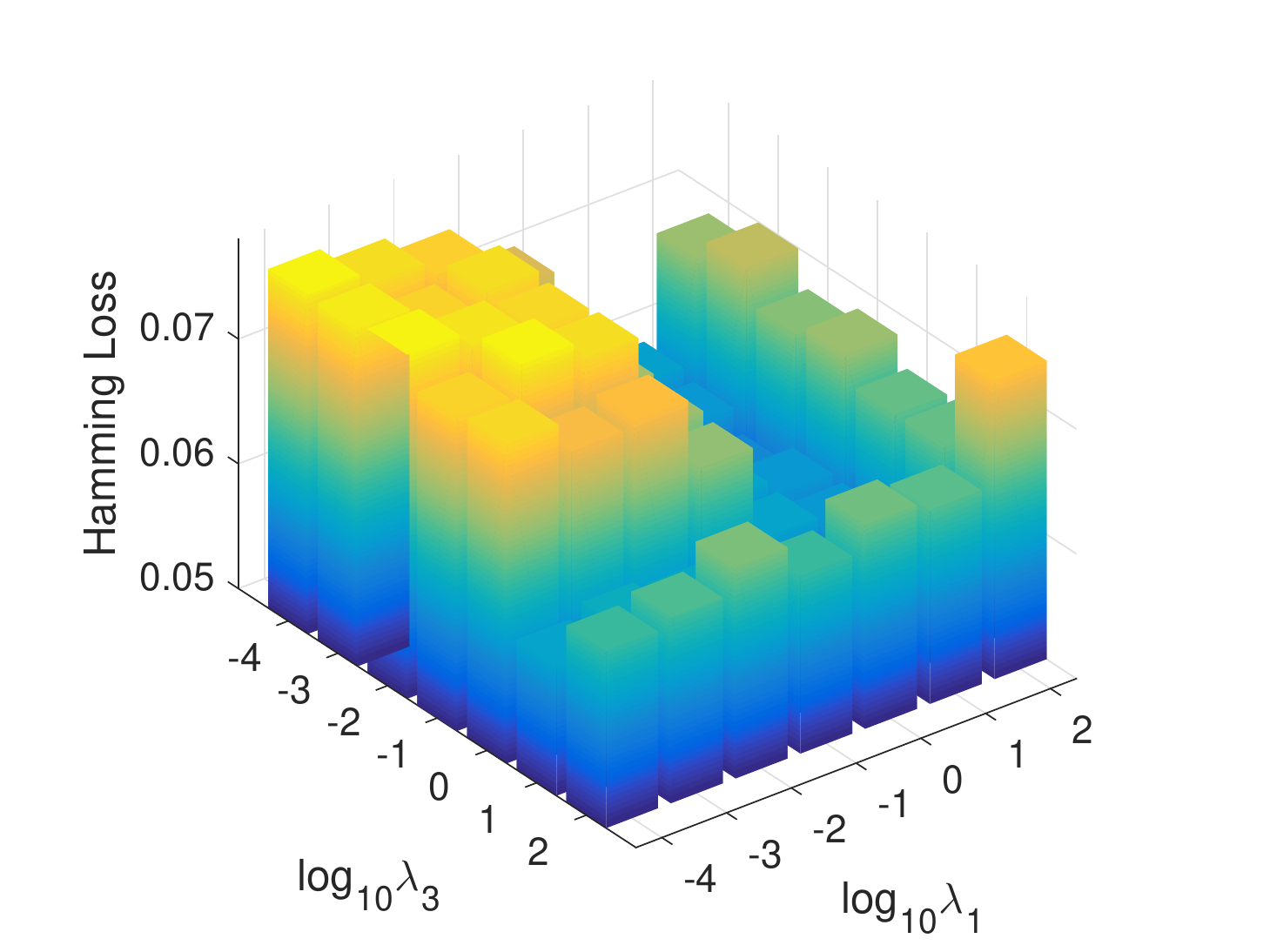}}
				\subfigure[]{
					\label{fig:sensitivity:sa_lambda2} 
					\includegraphics[width=1.5in]{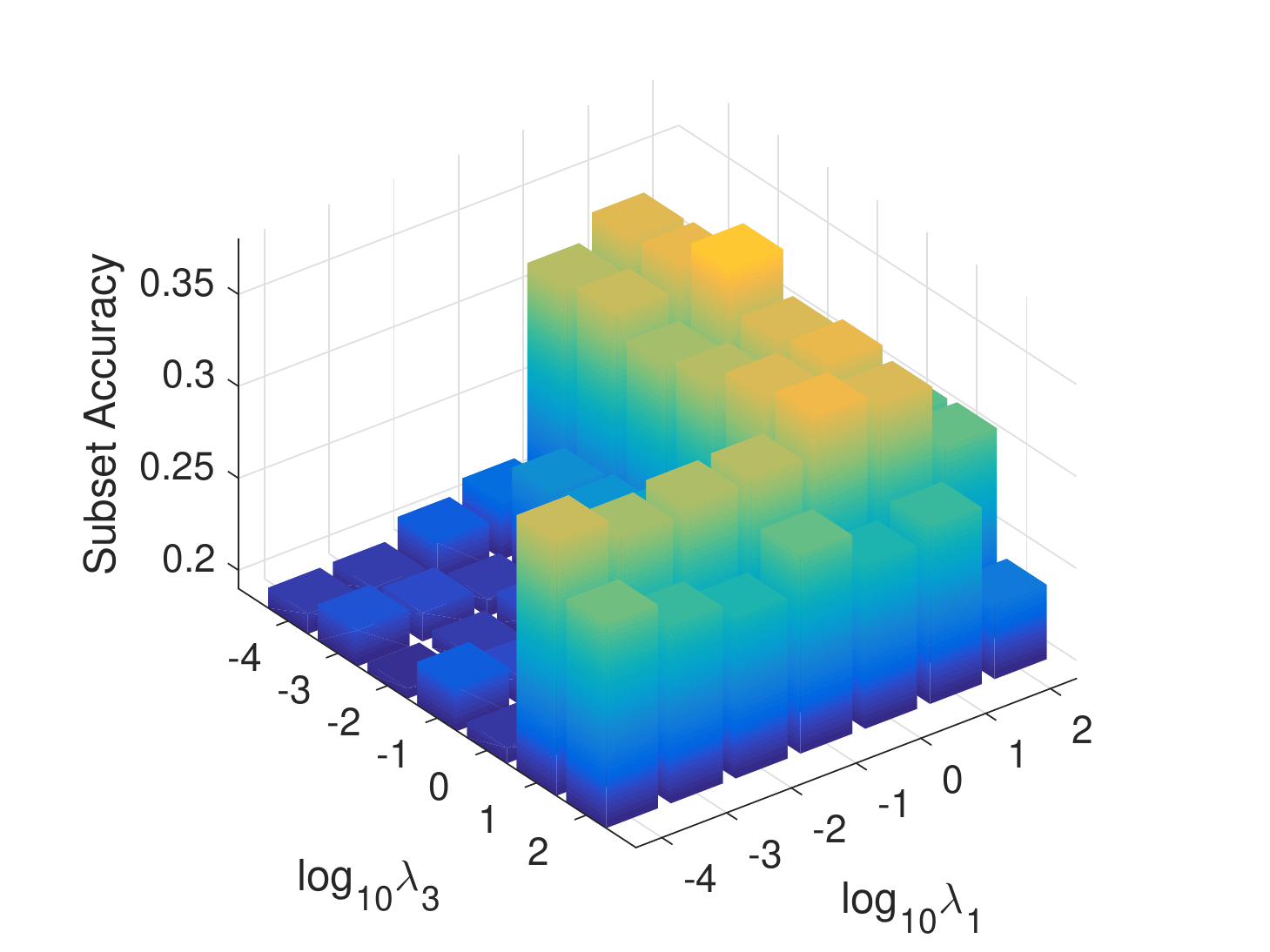}}
				\subfigure[]{
					\label{fig:sensitivity:f1e_lambda2} 
					\includegraphics[width=1.5in]{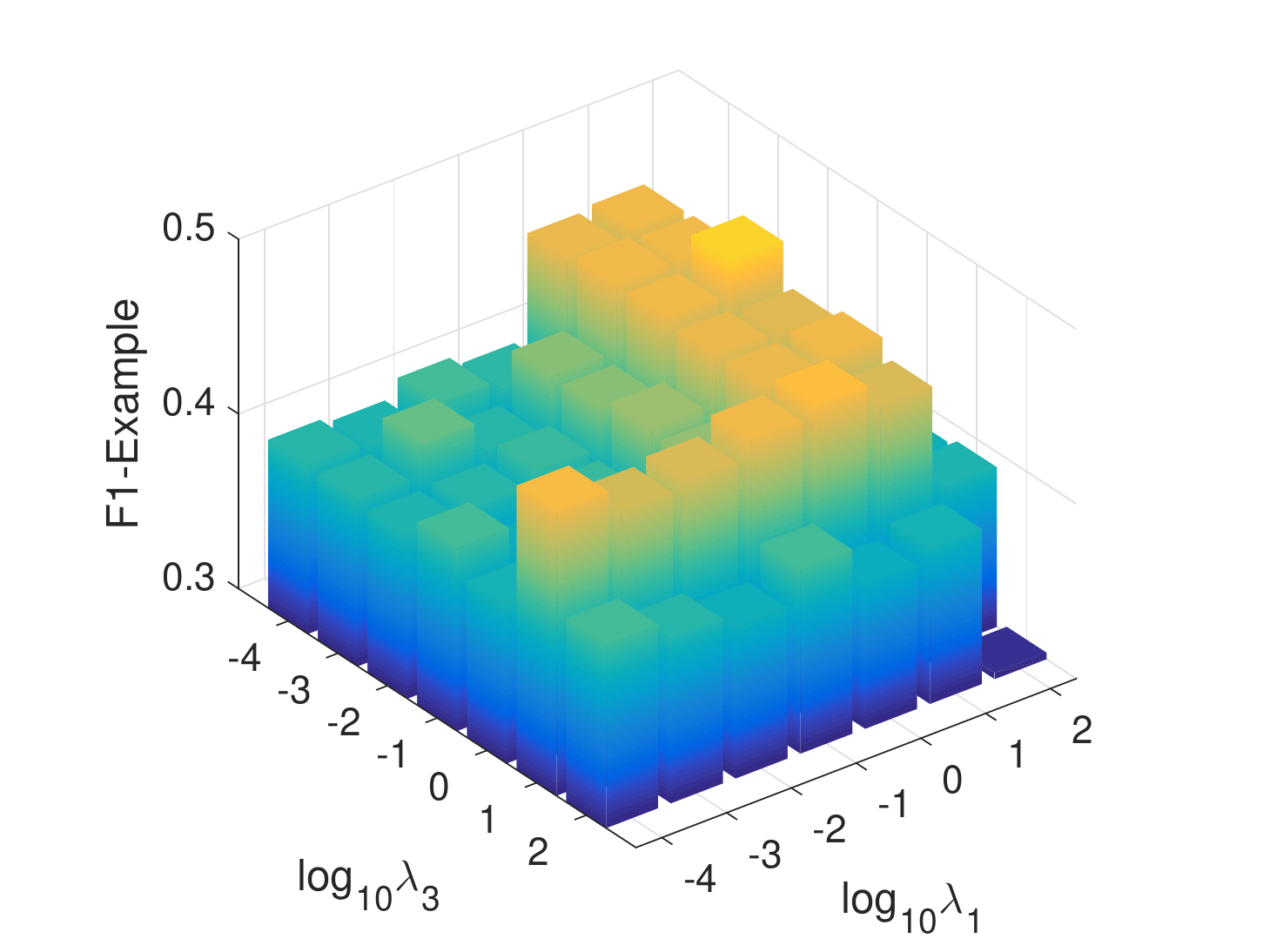}}
				\subfigure[]{
					\label{fig:sensitivity:ral_lambda2} 
					\includegraphics[width=1.5in]{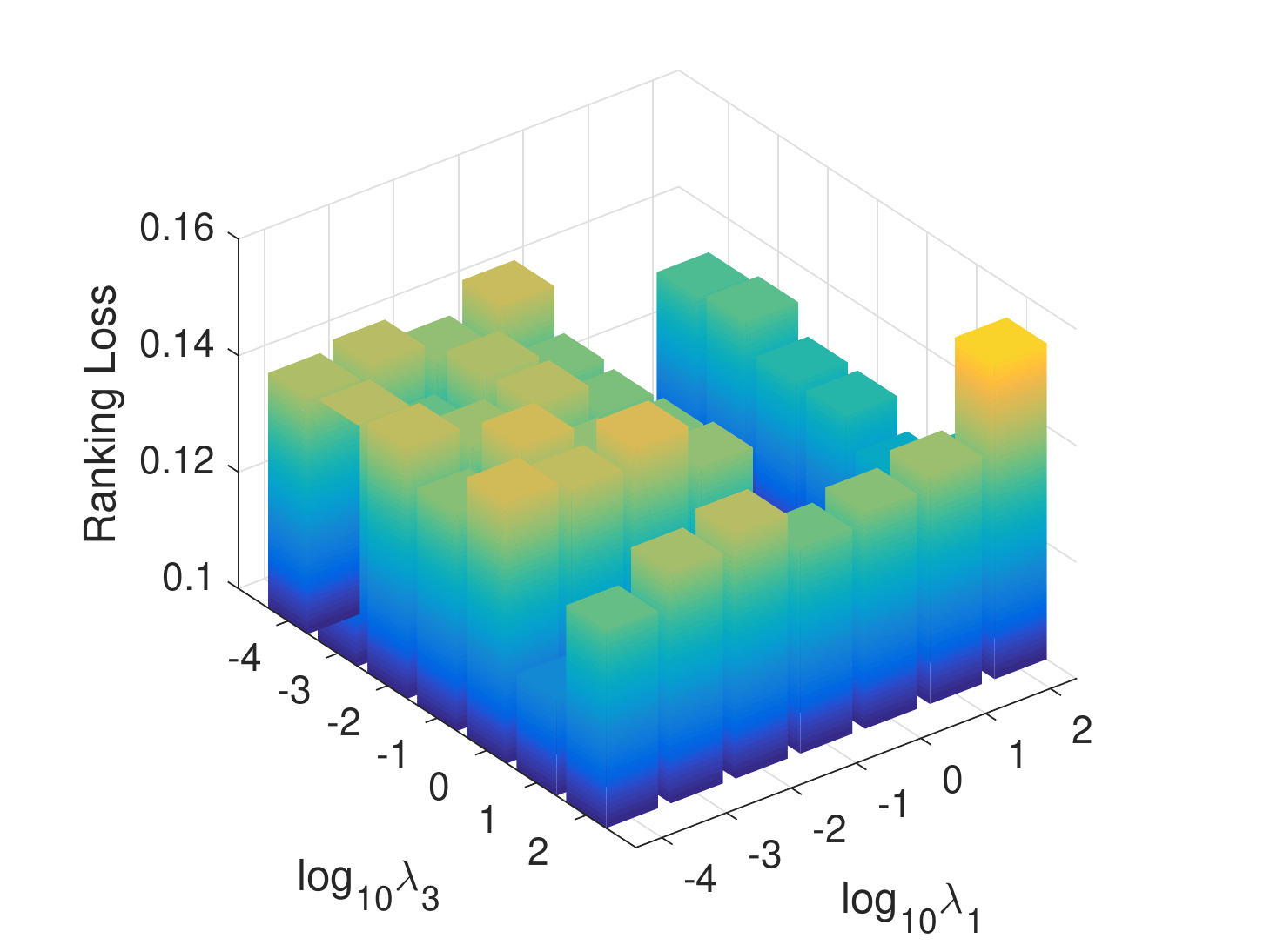}}
				\subfigure[]{
					\label{fig:sensitivity:cov_lambda2} 
					\includegraphics[width=1.5in]{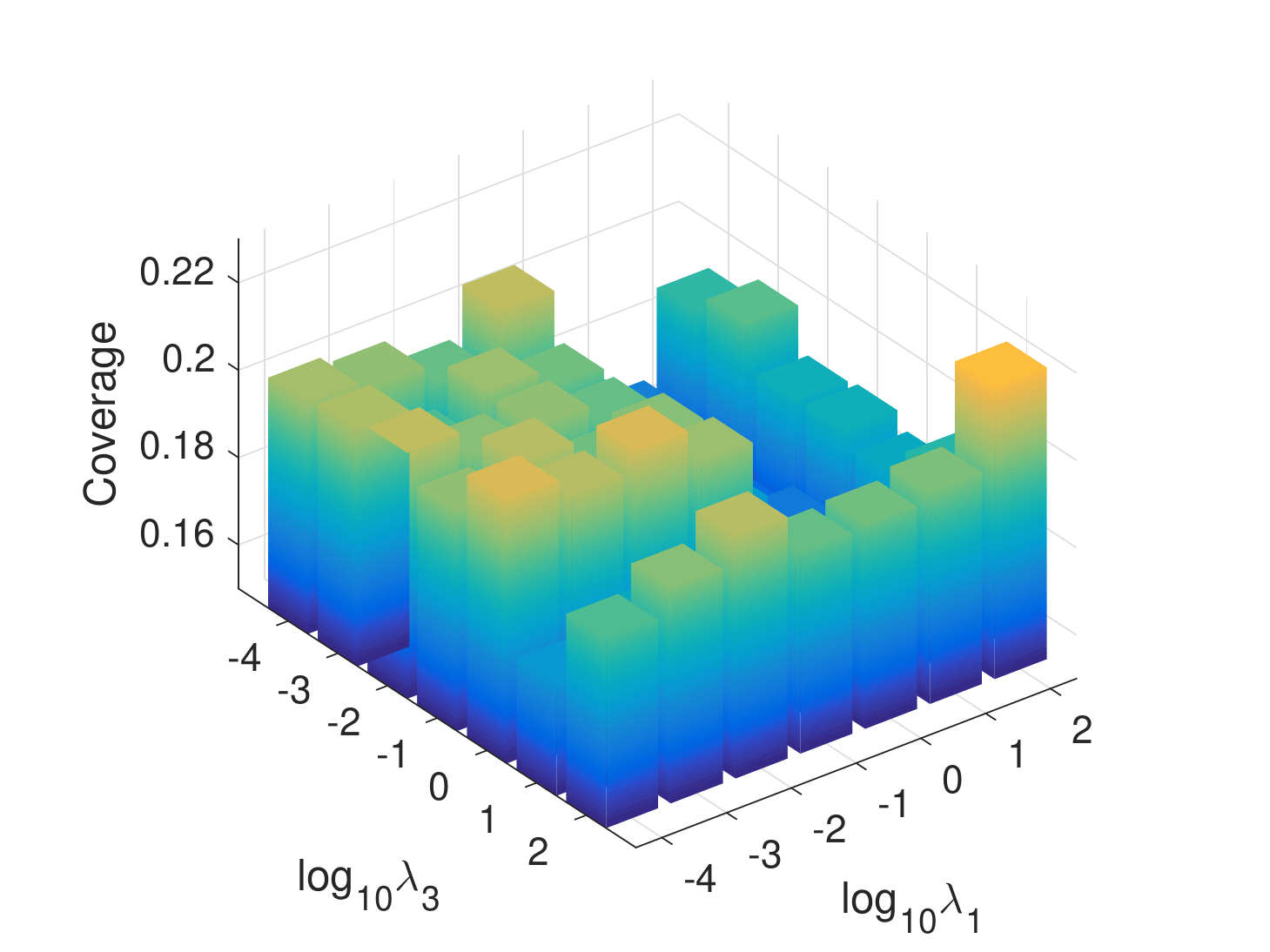}}
				\subfigure[]{
					\label{fig:sensitivity:ap_lambda2} 
					\includegraphics[width=1.5in]{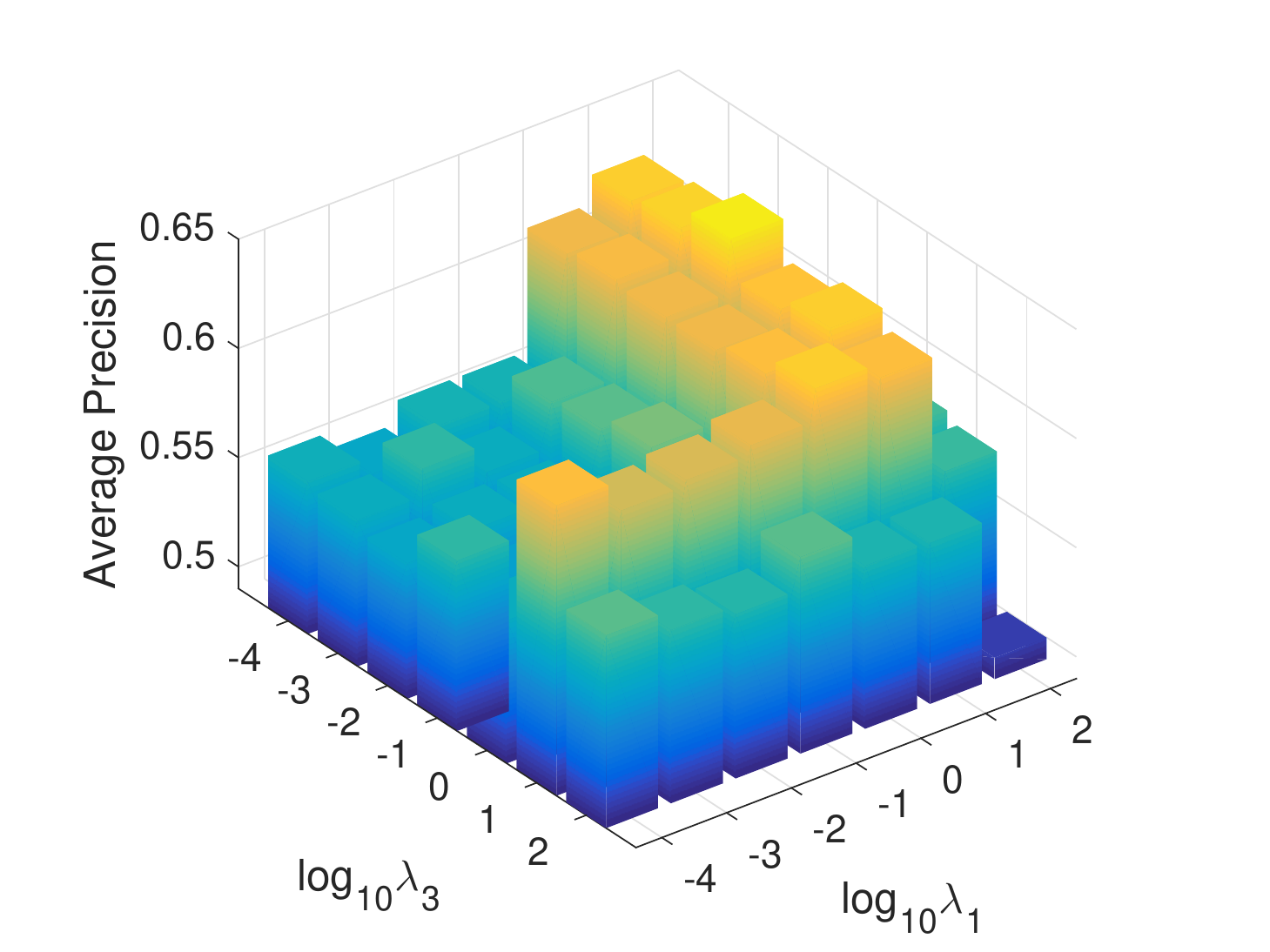}}
				\subfigure[]{
					\label{fig:sensitivity:hal_lambda3} 
					\includegraphics[width=1.5in]{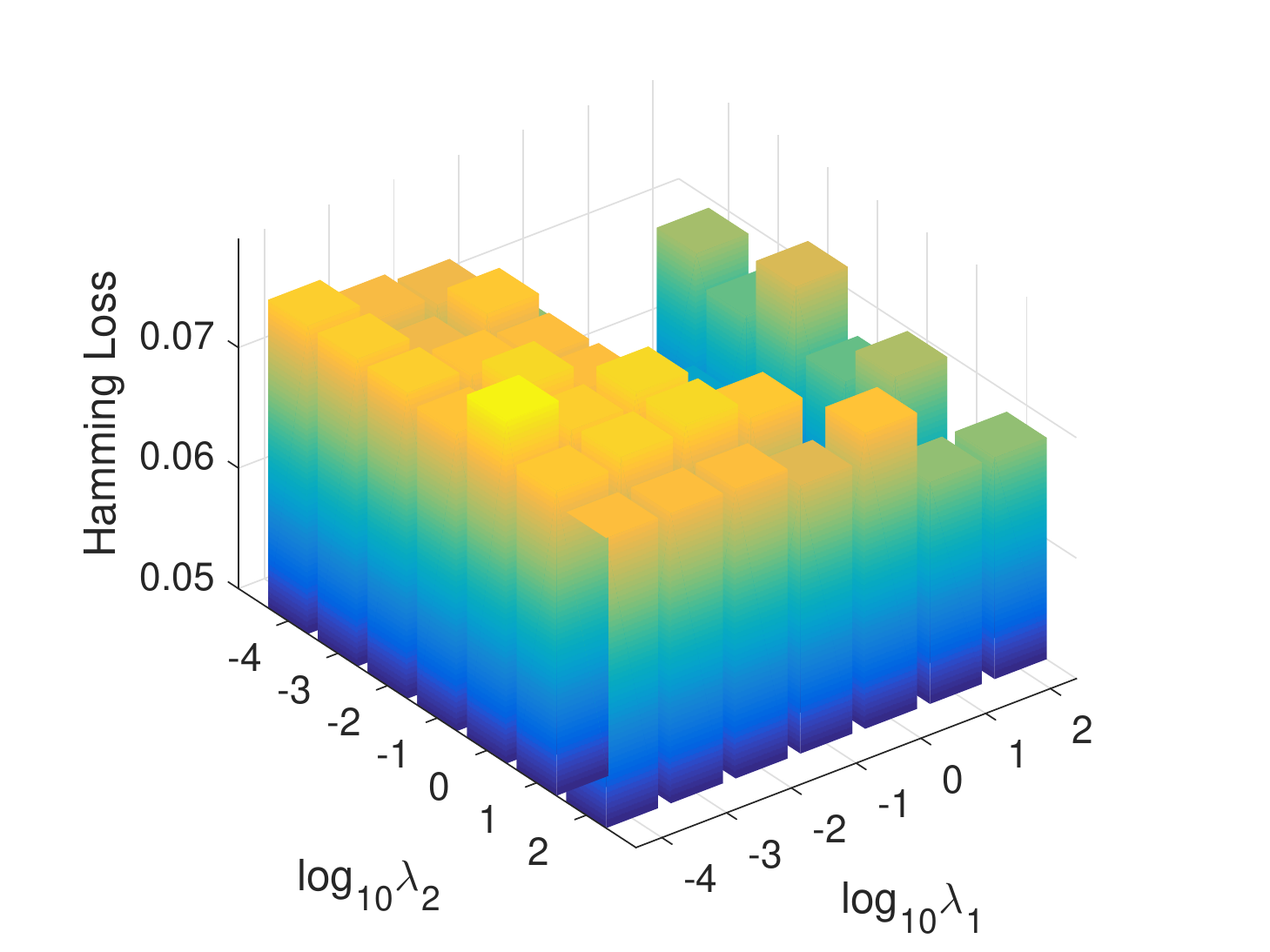}}
				\subfigure[]{
					\label{fig:sensitivity:sa_lambda3} 
					\includegraphics[width=1.5in]{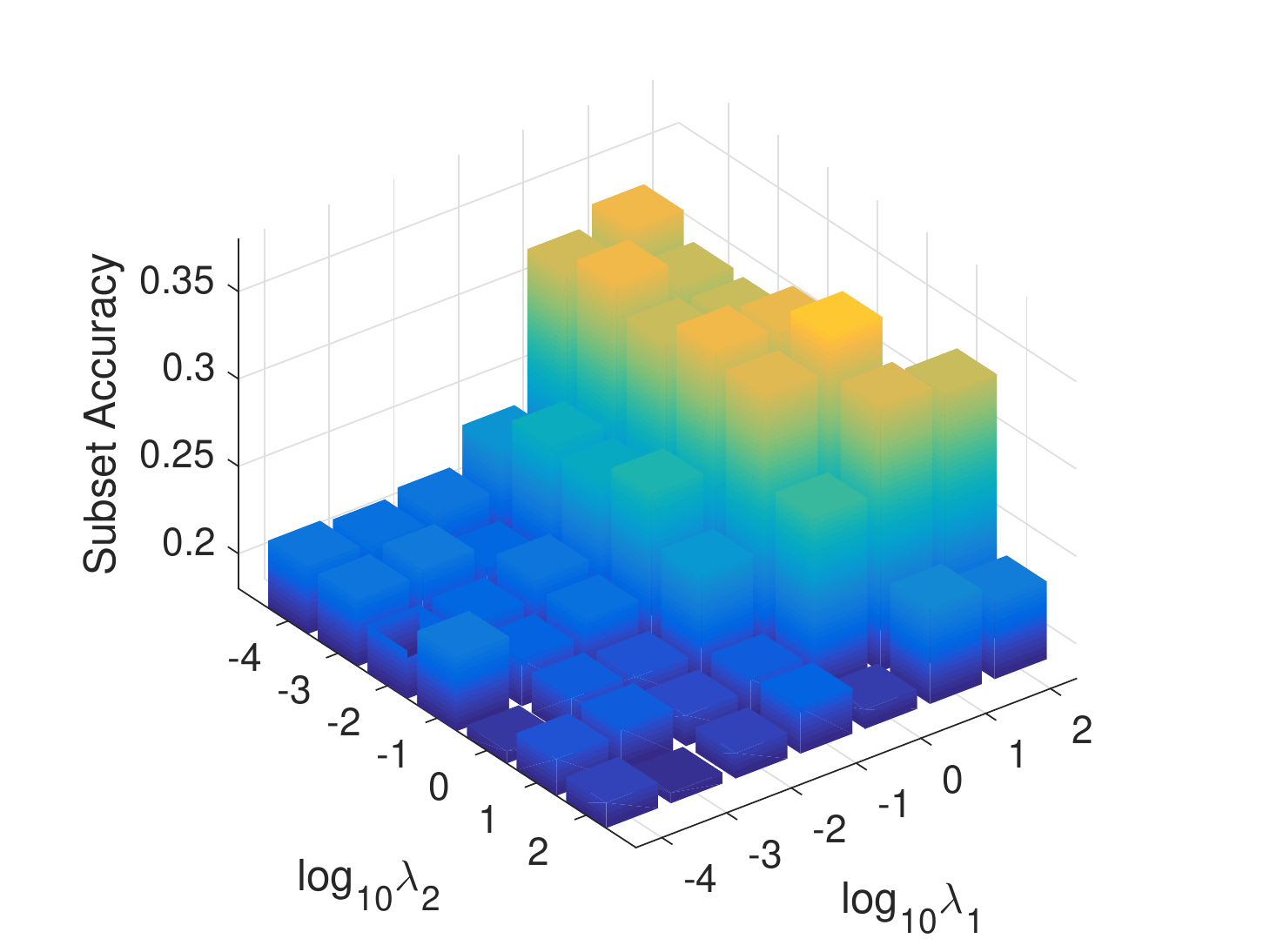}}
				\subfigure[]{
					\label{fig:sensitivity:f1e_lambda3} 
					\includegraphics[width=1.5in]{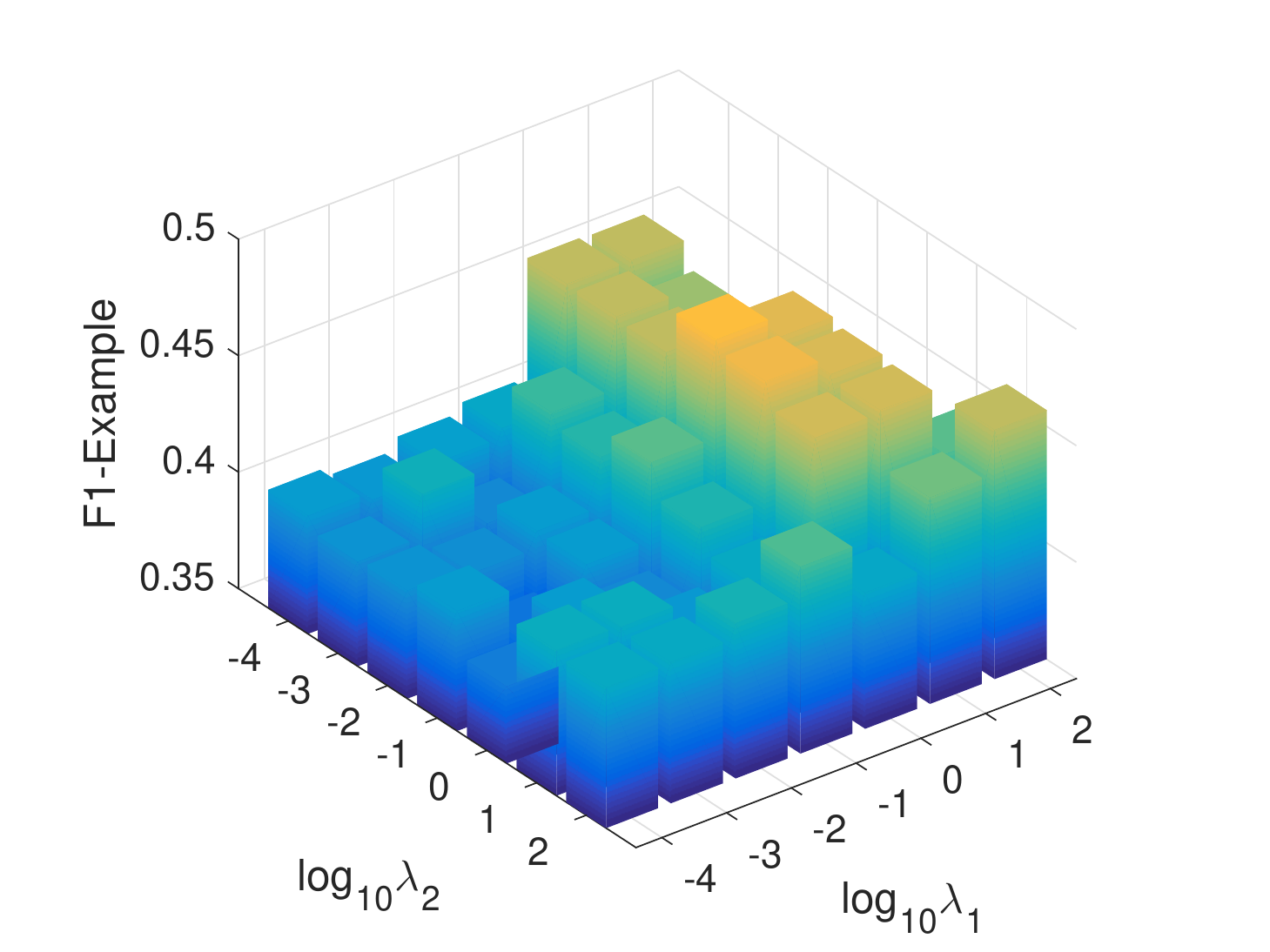}}
				\subfigure[]{
					\label{fig:sensitivity:ral_lambda3} 
					\includegraphics[width=1.5in]{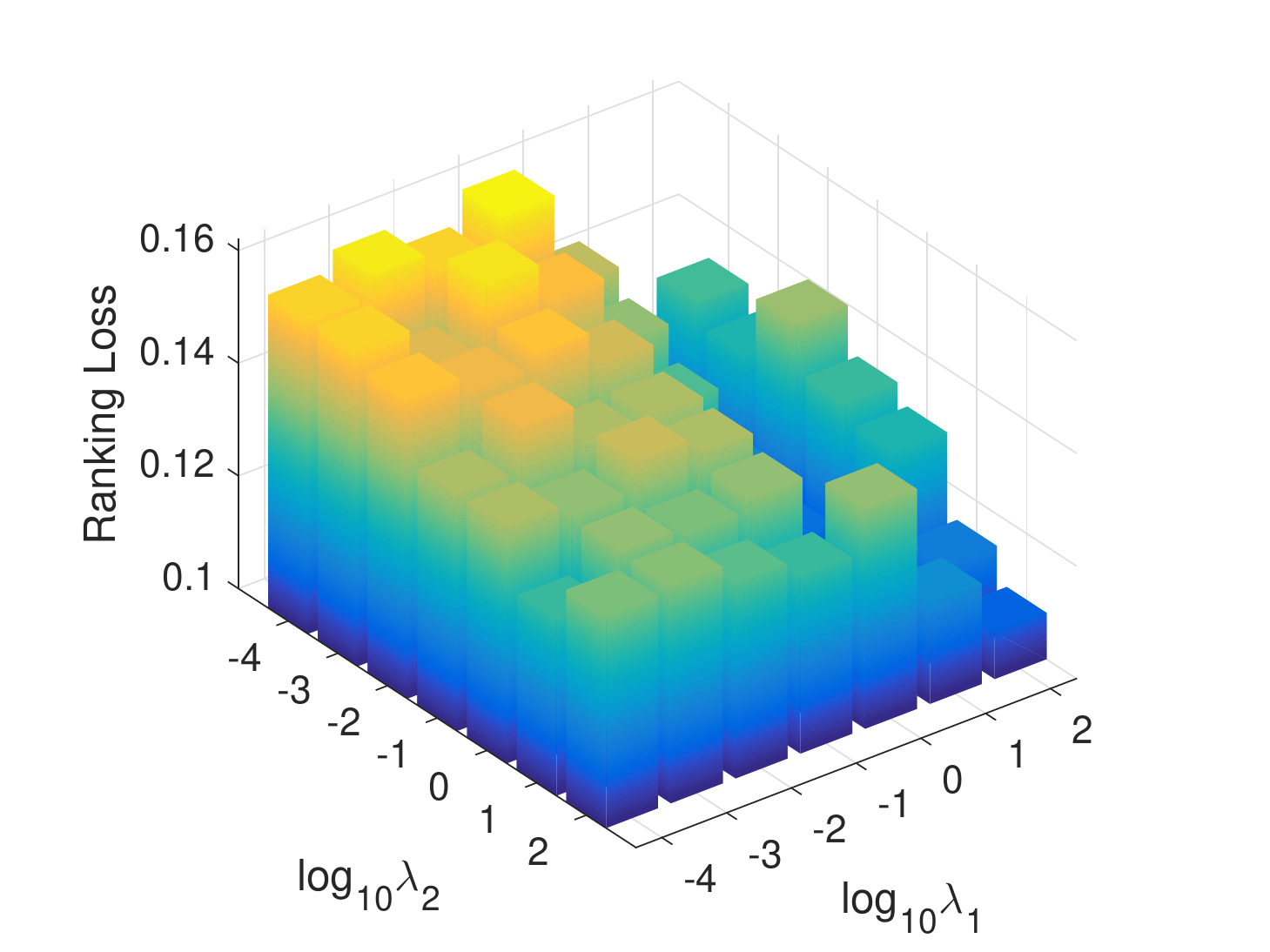}}
				\subfigure[]{
					\label{fig:sensitivity:cov_lambda3} 
					\includegraphics[width=1.5in]{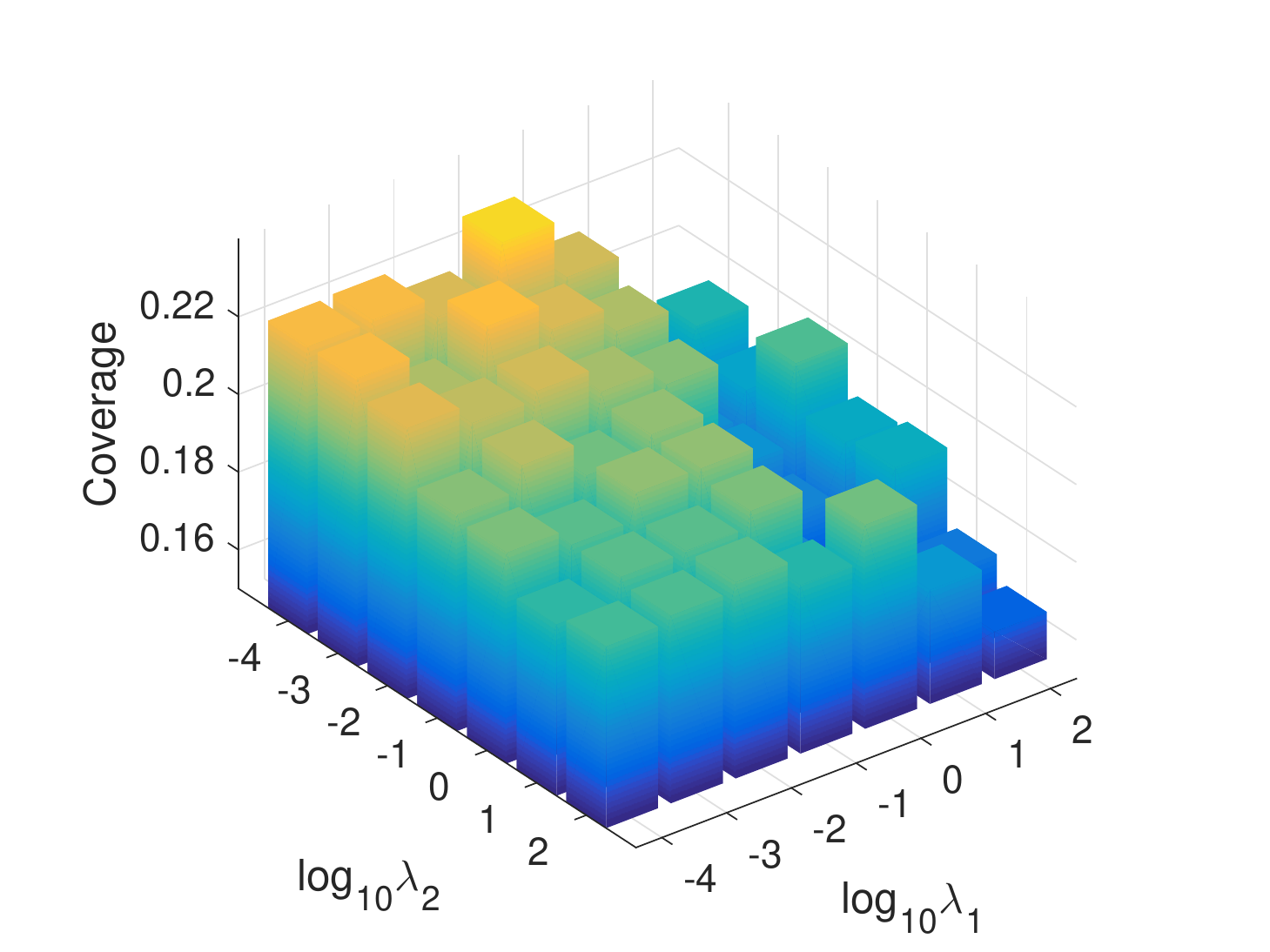}}
				\subfigure[]{
					\label{fig:sensitivity:ap_lambda3} 
					\includegraphics[width=1.5in]{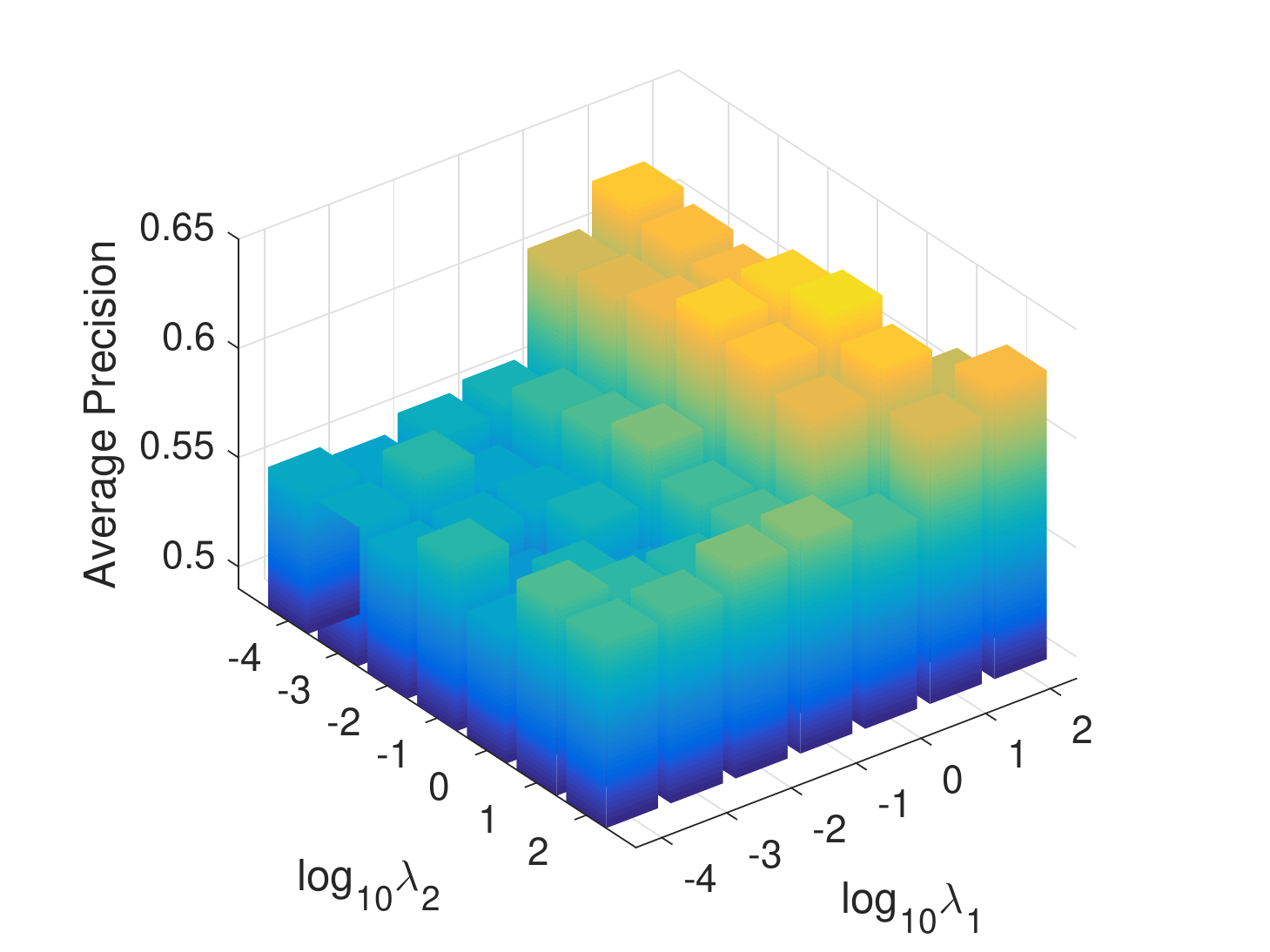}}
				\caption{Sensitivity analysis to the hyper-parameters of the linear RBRL on the \emph{arts} dataset. (a)-(f) The sensitivity of each evaluation metric with $\lambda_1$ fixed. (g)-(l) The sensitivity of each evaluation metric with $\lambda_2$ fixed. (m)-(r) The sensitivity of each evaluation metric with $\lambda_3$ fixed.}
				\label{fig:sensitivity} 
			\end{figure}

			To conduct sensitivity analysis to the hyper-parameters $\{ \lambda_1, \lambda_2, \lambda_3 \}$ of RBRL, we evaluate the linear RBRL on the \emph{arts} dataset. We first obtain the best hyper-parameters via fivefold cross-validation on the training set and then evaluate the performance effect of the other two with one hyper-parameter fixed. Figure \ref{fig:sensitivity} shows the results of the sensitivity analysis in terms of each evaluation metric. From Figure \ref{fig:sensitivity}, it can be observed that RBRL is sensitive to the hyper-parameters. Besides, the best performance is obtained at some intermediate values of $\lambda_1$, $\lambda_2$, and $\lambda_3$. Moreover, with the change of hyper-parameters, there are similar variations for all the evaluation metrics, especially ranking-based metrics (i.e., Ranking Loss, Coverage and Average Precision).
	 
	 	\subsubsection{Computational Time Cost}
	 	
	 		We compare the one partition computational time cost for all the comparing methods on all the datasets. All the experiments are conducted on a laptop machine with 4$\times$2.5GHz CPUs and 8GB RAM. Table \ref{tab:multi_label_result_time_cost} reports the average CPU time to train and test various methods. Generally, C++ or Java runs faster than Matlab, thus it's unfair to compare methods with different language implementations. Compared with two recent methods (i.e., CPNL and MLFE), RBRL achieves a comparable time cost. For the last six datasets (corresponding to the linear model), RBRL is more efficient than Rank-SVM because our linear RBRL implementation can be viewed as an acceleration of kernel RBRL with linear kernel while Rank-SVM implementation doesn't employ this linear acceleration. Besides, in the first four datasets (corresponding to the RBF kernel model), RBRL is slower than Rank-SVM, and better algorithms can be considered to improve the training efficiency of RBRL.
	 		\begin{table}[!htbp]
				\scriptsize
				\renewcommand{\arraystretch}{1.5}
				\caption{The computational time cost (in seconds) for the various compared methods. The second row shows the programming language used to implement the corresponding methods. (``tr" indicates the training time cost. ``te" indicates the test time cost.)}
				\hskip-70pt
				\label{tab:multi_label_result_time_cost}
				\begin{tabular}{c|cc|cc|cc|cc|cc|cc|cc|cc|cc}
					\hline
					 & \multicolumn{2}{c|}{Rank-SVM} & \multicolumn{2}{c|}{Rank-SVMz} & \multicolumn{2}{c|}{BR} & \multicolumn{2}{c|}{ML-kNN} & \multicolumn{2}{c|}{CLR} & \multicolumn{2}{c|}{RAKEL} & \multicolumn{2}{c|}{CPNL} & \multicolumn{2}{c|}{MLFE} & \multicolumn{2}{c}{RBRL} \\ 
					\cline{2-19}
					& \multicolumn{2}{c|}{Matlab} & \multicolumn{2}{c|}{C++} & \multicolumn{2}{c|}{Java} & \multicolumn{2}{c|}{Java} & \multicolumn{2}{c|}{Java} & \multicolumn{2}{c|}{Java} & \multicolumn{2}{c|}{Matlab} & \multicolumn{2}{c|}{Matlab} & \multicolumn{2}{c}{Matlab} \\ 
					\cline{2-19}
					& tr & te & tr & te & tr & te & tr & te & tr & te & tr & te & tr & te & tr & te & tr & te \\
					\hline
					emotions & 27 & 0.2 & 9 & 0.1 & 1 & 0.5 & 0.2 & 0.1 & 2 & 1 & 3 & 1 & 38 & 0.2 & 13 & 0.2 & 199 & 0.2 \\
					\hline 
					image & 60 & 4 & 47 & 2 & 21 & 7 & 5 & 3 & 30 & 14 & 70 & 16 & 3919 & 11 & 1174 & 1 & 2118 & 11 \\
					\hline 
					scene & 102 & 7 & 91 & 3 & 22 & 7 & 6 & 4 & 32 & 19 & 86 & 23 & 6433 & 15 & 2253 & 0.4 & 3525 & 16 \\
					\hline 
					yeast & 1584 & 5 & 402 & 2 & 31 & 15 & 3 & 2 & 70 & 49 & 64 & 14 & 2480 & 4 & 353 & 0.8 & 5158 & 5 \\
					\hline 
					enron & 12217 & 12 & 2394 & 5 & 7 & 3 & 2 & 1 & 38 & 28 & 26 & 1 & 713 & 0 & 1883 & 2 & 9998 & 0 \\
					\hline 
					arts & 7794 & 35 & 5153 & 13 & 75 & 1 & 37 & 24 & 144 & 7 & 1103 & 2 & 280 & 0 & 1717 & 4 & 1495 & 0 \\
					\hline 
					education & 11390 & 57 & 7663 & 18 & 78 & 1 & 53 & 35 & 130 & 12 & 1294 & 2 & 285 & 0 & 1960 & 5 & 1788 & 0 \\
					\hline
					recreation & 4733 & 58 & 3700 & 10 & 59 & 1 & 46 & 31 & 145 & 8 & 1348 & 2 & 168 & 0 & 1794 & 4 & 1252 & 0 \\
					\hline
					science & 16435 & 80 & 11271 & 27 & 107 & 2 & 76 & 51 & 182 & 25 & 2467 & 2 & 398 & 0 & 2434 & 5 & 2759 & 0 \\
					\hline
					business & 3109 & 30 & 3650 & 15 & 53 & 1 & 47 & 30 & 136 & 9 & 621 & 1 & 295 & 0 & 1588 & 4 & 1585 & 0 \\
					\hline
				\end{tabular}
	 		\end{table}
	
	\section{Conclusions}
	\label{sec:conclusion}
		In this paper, we have presented a novel multi-label classification model, which joints Rank-SVM and Binary Relevance with robust Low-rank learning (RBRL). It incorporates the thresholding step into the ranking learning step of Rank-SVM via BR to train the model in only one step. Besides, the low-rank constraint is utilized to further exploit the label correlations. In addition, the kernelization RBRL is presented to obtain nonlinear multi-label classifiers. Moreover, two accelerated proximal gradient methods (APG) are employed to solve the linear and kernel RBRL efficiently. Extensive experiments against several state-of-the-art approaches have confirmed the effectiveness of our approach RBRL.
		
		In the future, it's interesting to combine the linear RBRL with deep neural networks (such as CNN) to learn from the raw multi-label data.
		
	\section*{Acknowledgements}
		This work has been partially supported by grants from:  Science and Technology Service Network Program of Chinese Academy of Sciences (STS Program, No. KFJ-STS-ZDTP-060), and National Natural Science Foundation of China (No. 71731009, 61472390, 71331005, 91546201).
	
	\section*{Conflict of interest}
	
	Declarations of interest: none.
	
	\appendix
	\section{Proof of Theorem \ref{theorem:linear_representer}}
	\label{appendix:1}
		\begin{proof}
			The problem \eqref{equation:kernel_model_rank} can be rewritten as
			\begin{equation}
			\label{equation:model_another}
			\begin{split}
			& \min \limits_\mathbf{W} \ \frac{1}{2} \sum_{i=1}^{n} \sum_{j=1}^{l} max(0, 1 - y_{ij} \langle \mathbf{w}^j, \phi(\mathbf{x}_i) \rangle )^2 + \frac{\lambda_1}{2} \| \mathbf{W} \|_F^2 + \\ 
			& \qquad \frac{\lambda_2}{2} \sum_{i=1}^{n} \frac{1}{|Y_i^+||Y_i^-|} \sum_{p \in Y_i^+} \sum_{q \in Y_i^-} max(0, 2 - \langle \mathbf{w}^p - \mathbf{w}^q, \phi(\mathbf{x}_i) \rangle)^2 \\
			& s.t. \ Rank(\mathbf{W}) \leq k
			\end{split}
			\end{equation}
			
			Denote the objective function in \eqref{equation:model_another} except the second term as $f(\mathbf{W}, \Phi(\mathbf{X}), \mathbf{Y})$. Hence, the optimization problem becomes 
			$\min \limits_{Rank(\mathbf{W}) \leq k} f(\mathbf{W}, \Phi(\mathbf{X}), \mathbf{Y}) + \frac{\lambda_1}{2} \| \mathbf{W} \|_F^2$.
			
			Let $\mathbf{W}_* = [\mathbf{w}_*^1, \mathbf{w}_*^2,...,\mathbf{w}_*^l]$ be an optimal solution of \eqref{equation:model_another}. Because $\mathbf{w}_*^j$ is an element of a Hilbert space for all $j$, we can rewrite $\mathbf{W}_*$ as
			\begin{equation}
			\mathbf{w}_*^j = \sum_{i=1}^{n} \alpha_{ij} \phi(\mathbf{x}_i) + \mathbf{u}^j = \Phi(\mathbf{X})^\top \alpha^j + \mathbf{u}^j, \ j = 1,...,l 
			\end{equation}
			or equivalently
			\begin{equation}
			\mathbf{W}_* = \Phi(\mathbf{X})^\top \mathbf{A} + \mathbf{U}
			\end{equation}
			where $\mathbf{U} = [\mathbf{u}^1, \mathbf{u}^2,...,\mathbf{u}^l]$, $\mathbf{A} = [\alpha_1,...,\alpha_l]$ and $\langle \mathbf{u}^j, \phi(\mathbf{x}_i) \rangle = 0$ for all $i,j$. Set $\mathbf{W} = \mathbf{W}_* - \mathbf{U}$. Clearly, $\|\mathbf{W}_*\|_F^2 = \|\mathbf{W}\|_F^2 + \|\mathbf{U}\|_F^2$, and $\lambda_1 > 0$, thus $\frac{\lambda_1}{2}\|\mathbf{W}\|_F^2 \leq \frac{\lambda_1}{2}\|\mathbf{W}_*\|_F^2$. Assume that there exsits $\mathbf{A}$ such that $Rank(\mathbf{A}) \leq k$, thus $Rank(\mathbf{W}) \leq k$. Additionally, for all $i,j$ we have that
			\begin{equation}
			\langle \mathbf{w}^j, \phi(\mathbf{x}_i) \rangle = \langle \mathbf{w}_*^j - \mathbf{u}^j, \phi(\mathbf{x}_i) \rangle = \langle \mathbf{w}_*^j, \phi(\mathbf{x}_i) \rangle 
			\end{equation}
			hence
			\begin{equation}
			f(\mathbf{W}, \Phi(\mathbf{X}), \mathbf{Y}) = f(\mathbf{W}_*, \Phi(\mathbf{X}), \mathbf{Y})
			\end{equation}
			It can be observed that the objective function of \eqref{equation:model_another} at $\mathbf{W}$ cannot be larger than the objective function at $\mathbf{W}_*$. Besides, $\mathbf{W}$ is also in the feasible region (i.e., $Rank(\mathbf{W}) \leq k$). Thus, $\mathbf{W} = \Phi(\mathbf{X})^\top \mathbf{A}, \ s.t. \ Rank(\mathbf{A}) \leq k$ is also an optimal solution.
		\end{proof}
	
	\section{Proof of Theorem \ref{theorem:linear_lipschitz_constant}}
		\subsection{Proof of the Lemma \ref{lemma:2}}
		\label{appendix:2}
			\begin{proof}
				$\forall \ a_i, a_j \in \mathbb{R}^m$, it always holds
				\begin{equation}
					2 \langle a_i, a_j \rangle \leq \| a_i \|^2 + \| a_j \|^2
				\end{equation}
				Thus, $\forall \ a_1, a_2, ... , a_n \in \mathbb{R}^m$, we have
				\begin{equation}
				\begin{split}
					& \| a_1 + a_2 + ... + a_n \|^2 \\
					& = \sum_{i=1}^{n} \| a_i \|^2 + \sum_i \sum_{j \neq i} 2 \langle a_i, a_j \rangle \\
					& \leq \sum_{i=1}^{n} \| a_i \|^2 + \sum_i \sum_{j \neq i} ( \| a_i \|^2 + \| a_j \|^2 ) \\
					& = n \sum_{i=1}^{n} \| a_i \|^2
				\end{split}
				\end{equation}
			\end{proof}
						
		\subsection{Proof of the Proposition \ref{proposition:linear_lipschitz_constant_part}}
		\label{appendix:3}			
			\begin{proof}
				Firstly, we compute the \emph{Lipschitz} constant for $\frac{\partial f_r}{\partial {\mathbf{w}^j}}, j=1,2,...,l$. $\forall \ \mathbf{u}_1, \mathbf{u}_2 \in \mathbb{R}^m$, we have
				\begin{equation}
				\begin{split}
				& \|\frac{\partial f_r(\mathbf{u}_1)}{\partial {\mathbf{w}^j}} - \frac{\partial f_r(\mathbf{u}_2)}{\partial {\mathbf{w}^j}} \|^2 \\
				& = \| \sum_{i=1}^{n} \frac{1}{|Y_i^+||Y_i^-|} \Big \{ [\![ j \in Y_i^{+} ]\!] \sum_{q \in Y_i^-} (| 2 - \langle \mathbf{u}_1 - \mathbf{w}^q, \mathbf{x}_i \rangle |_+ - |2 - \langle \mathbf{u}_2 - \mathbf{w}^q, \mathbf{x}_i \rangle |_+) (-\mathbf{x}_i) \\ 
				& \quad + [\![ j \in Y_i^{-} ]\!] \sum_{p \in Y_i^+} (| 2 - \langle \mathbf{w}^p - \mathbf{u}_1, \mathbf{x}_i \rangle |_+ - |2 - \langle \mathbf{w}^p - \mathbf{u}_2, \mathbf{x}_i \rangle |_+) \mathbf{x}_i \Big \} \|^2 \\
				& \leq \Big \{ \sum_{i=1}^{n} ([\![ j \in Y_i^{+} ]\!] |Y_i^-| + [\![ j \in Y_i^{-} ]\!] |Y_i^+|) \Big \} \sum_{i=1}^{n} \frac{1}{|Y_i^+|^2 |Y_i^-|^2} \Big \{ [\![ j \in Y_i^{+} ]\!] \sum_{q \in Y_i^-} \| (| 2 - \langle \mathbf{u}_1 - \mathbf{w}^q, \mathbf{x}_i \rangle |_+ \\ 
				& \quad - |2 - \langle \mathbf{u}_2 - \mathbf{w}^q, \mathbf{x}_i \rangle |_+) (-\mathbf{x}_i)  \|^2 + [\![ j \in Y_i^{-} ]\!] \sum_{p \in Y_i^+} \| (| 2 - \langle \mathbf{w}^p - \mathbf{u}_1, \mathbf{x}_i \rangle |_+ \\
				& \quad  - |2 - \langle \mathbf{w}^p - \mathbf{u}_2, \mathbf{x}_i \rangle |_+) \mathbf{x}_i  \|^2 \Big \} \\
				& \leq  \Big \{ \sum_{i=1}^{n} ([\![ j \in Y_i^{+} ]\!] |Y_i^-| + [\![ j \in Y_i^{-} ]\!] |Y_i^+|) \Big \} \sum_{i=1}^{n} \frac{1}{|Y_i^+|^2 |Y_i^-|^2} \Big \{ [\![ j \in Y_i^{+} ]\!] \sum_{q \in Y_i^-} \| \langle \mathbf{u}_2 - \mathbf{u}_1, \mathbf{x}_i \rangle \|^2 \\
				& \quad + [\![ j \in Y_i^{-} ]\!] \sum_{p \in Y_i^+} \| \langle \mathbf{u}_1 - \mathbf{u}_2, \mathbf{x}_i \rangle \|^2 \Big \} \| \mathbf{x}_i \|^2 \\
				& \leq \Big \{ \sum_{i=1}^{n} ([\![ j \in Y_i^{+} ]\!] |Y_i^-| + [\![ j \in Y_i^{-} ]\!] |Y_i^+|) \Big \} \sum_{i=1}^{n} \frac{[\![ j \in Y_i^{+} ]\!] |Y_i^-| + [\![ j \in Y_i^{-} ]\!] |Y_i^+|}{|Y_i^+|^2 |Y_i^-|^2} \| \mathbf{x}_i \|^2 \| \mathbf{x}_i \|^2 \| \mathbf{u}_1 - \mathbf{u}_2 \|^2 \\
				& = \Big \{ \sum_{i=1}^{n} ([\![ j \in Y_i^{+} ]\!] |Y_i^-| + [\![ j \in Y_i^{-} ]\!] |Y_i^+|) \Big \} \Big \{ \sum_{i=1}^{n} \frac{ ( [\![ j \in Y_i^{+} ]\!] |Y_i^-| + [\![ j \in Y_i^{-} ]\!] |Y_i^+| ) \| \mathbf{x}_i \|^4 }{|Y_i^+|^2 |Y_i^-|^2 } \Big \} \| \Delta \mathbf{u} \|^2 
				\end{split}
				\end{equation}
				where $\Delta \mathbf{u} = \mathbf{u}_1 - \mathbf{u}_2$. Thus, the \emph{Lipschitz} constant for $\frac{\partial f_r}{\partial {\mathbf{w}^j}}, j=1,2,...,l$ is as follows.
				\begin{equation}
				L_{f_r}^j = \sqrt{A_j * B_j}
				\end{equation}
				where $B_j = \sum_{i=1}^{n} \frac{ ( [\![ j \in Y_i^{+} ]\!] |Y_i^-| + [\![ j \in Y_i^{-} ]\!] |Y_i^+| ) \| \mathbf{x}_i \|^4 }{|Y_i^+|^2 |Y_i^-|^2}$, and $A_j = \sum_{i=1}^{n} ([\![ j \in Y_i^{+} ]\!] |Y_i^-| + [\![ j \in Y_i^{-} ]\!] |Y_i^+| ), j = 1,...,l$.
				
				Then, for $\nabla_\mathbf{W}{f_r(\mathbf{W})} = [\frac{\partial f_r}{\partial {\mathbf{w}^1}},\frac{\partial f_r}{\partial {\mathbf{w}^2}},...,\frac{\partial f_r}{\partial {\mathbf{w}^l}}]$, $\forall \ \mathbf{U}_1, \mathbf{U}_2 \in \mathbb{R}^{m \times l}$, we have
				\begin{equation}
				\begin{split}
				& \|\nabla{f_r(\mathbf{U}_1)} - \nabla{f_r(\mathbf{U}_2)} \|_F^2 \\
				& = \sum_{j=1}^{l} \| \frac{\partial f_r(\mathbf{u}_1^j)}{\partial {\mathbf{w}^j}} - \frac{\partial f_r(\mathbf{u}_2^j)}{\partial {\mathbf{w}^j}} \|^2 \\
				& \leq \sum_{i=1}^{l} (L_f^j)^2 \| \Delta \mathbf{u}^j \|^2 \\
				& = \sum_{i=1}^{l} A_j B_j \| \Delta \mathbf{u}^j \|^2 \\
				& \leq max\{ A_j B_j \}_{j=1,...,l} (\sum_{i=1}^{l} \| \Delta \mathbf{u}^j \|^2) \\
				& =  max\{ A_j B_j \}_{j=1,...,l} \| \Delta \mathbf{U} \|_F^2
				\end{split}
				\end{equation}
				where $\Delta \mathbf{U} = \mathbf{U}_1 - \mathbf{U}_2$, $l$ is the number of the labels, $B_j = \sum_{i=1}^{n} \frac{ ( [\![ j \in Y_i^{+} ]\!] |Y_i^-| + [\![ j \in Y_i^{-} ]\!] |Y_i^+| ) \| \mathbf{x}_i \|^4 }{|Y_i^+|^2 |Y_i^-|^2}$, and $A_j = \sum_{i=1}^{n} ([\![ j \in Y_i^{+} ]\!] |Y_i^-| + [\![ j \in Y_i^{-} ]\!] |Y_i^+| ), j = 1,...,l$.
			\end{proof}
			
		\subsection{Proof of the Theorem \ref{theorem:linear_lipschitz_constant}}
		\label{appendix:4}
			\begin{proof}	
				$\forall \ \mathbf{W}_1, \mathbf{W}_2 \in \mathbb{R}^{m \times l}$, we have
				\begin{equation}
				\label{equation:linear_model_lipschitz_induce}
				\begin{split}
				& \|\nabla{f(\mathbf{W}_1)} - \nabla{f(\mathbf{W}_2)} \|_F^2 \\
				& = \| \mathbf{X}^\top ( (|\mathbf{E} -\mathbf{Y} \circ (\mathbf{XW}_1)|_{+} - |\mathbf{E} -\mathbf{Y} \circ (\mathbf{XW}_2)|_{+}) \circ (-\mathbf{Y}) ) \\
				& \quad + \lambda_1 \Delta \mathbf{W} + \lambda_2 (\nabla{f_r(\mathbf{W}_1)} - \nabla{f_r(\mathbf{W}_2)}) \|_F^2 \\
				& \leq 3 \| \mathbf{X}^\top ( (|\mathbf{E} -\mathbf{Y} \circ (\mathbf{XW}_1)|_{+} - |\mathbf{E} -\mathbf{Y} \circ (\mathbf{XW}_2)|_{+}) \circ (-\mathbf{Y}) ) \|_F^2 \\
				& \quad + 3 \| \lambda_1 \Delta \mathbf{W} \|_F^2 + 3 \| \lambda_2 (\nabla{f_r(\mathbf{W}_1)} - \nabla{f_r(\mathbf{W}_2)}) \|_F^2 \\
				& \leq 3 \| \mathbf{X}^\top \|_F^2 \| |\mathbf{E} -\mathbf{Y} \circ (\mathbf{XW}_1)|_{+} - |\mathbf{E} -\mathbf{Y} \circ (\mathbf{XW}_2)|_{+} \|_F^2 \\
				& \quad + 3 \lambda_1^2 \| \Delta \mathbf{W} \|_F^2 + 3 (\lambda_2 L_{f_r})^2 \| \Delta \mathbf{W} \|_F^2 \\
				& \leq 3 \| \mathbf{X} \|_F^2 \| -\mathbf{Y} \circ (\mathbf{X} \Delta \mathbf{W}) \|_F^2 + 3 \lambda_1^2 \| \Delta \mathbf{W} \|_F^2 + 3 (\lambda_2 L_{f_r})^2 \| \Delta \mathbf{W} \|_F^2 \\
				& \leq 3 \| \mathbf{X} \|_F^2  \| \mathbf{X} \|_F^2 \| \Delta \mathbf{W} \|_F^2 + 3 \lambda_1^2 \| \Delta \mathbf{W} \|_F^2 + 3 (\lambda_2 L_{f_r})^2 \| \Delta \mathbf{W} \|_F^2 \\ 
				& = ( 3 (\| \mathbf{X} \|_F^2)^2 + 3 \lambda_1^2 + 3 (\lambda_2 L_{f_r})^2 ) \| \Delta \mathbf{W} \|_F^2
				\end{split}
				\end{equation}
				where $\Delta \mathbf{W} = \mathbf{W}_1 - \mathbf{W}_2$.
			\end{proof}

	\section*{References}
	\bibliography{cite_list}

\begin{thebibliography}{62}
\expandafter\ifx\csname natexlab\endcsname\relax\def\natexlab#1{#1}\fi
\expandafter\ifx\csname url\endcsname\relax
  \def\url#1{\texttt{#1}}\fi
\expandafter\ifx\csname urlprefix\endcsname\relax\def\urlprefix{URL }\fi

\bibitem[{Argyriou et~al.(2008)Argyriou, Evgeniou, and
  Pontil}]{argyriou2008convex}
Argyriou, A., Evgeniou, T., Pontil, M., 2008. Convex multi-task feature
  learning. Machine Learning 73~(3), 243--272.

\bibitem[{Boutell et~al.(2004)Boutell, Luo, Shen, and
  Brown}]{boutell2004learning}
Boutell, M.~R., Luo, J., Shen, X., Brown, C.~M., 2004. Learning multi-label
  scene classification. Pattern Recognition 37~(9), 1757--1771.

\bibitem[{Cai et~al.(2010)Cai, Cand{\`e}s, and Shen}]{cai2010singular}
Cai, J.-F., Cand{\`e}s, E.~J., Shen, Z., 2010. A singular value thresholding
  algorithm for matrix completion. SIAM Journal on Optimization 20~(4),
  1956--1982.

\bibitem[{Chang and Lin(2011)}]{CC01a}
Chang, C.-C., Lin, C.-J., 2011. {LIBSVM}: A library for support vector
  machines. ACM Transactions on Intelligent Systems and Technology 2,
  27:1--27:27.

\bibitem[{Clare and King(2001)}]{clare2001knowledge}
Clare, A., King, R.~D., 2001. Knowledge discovery in multi-label phenotype
  data. In: European Conference on Principles of Data Mining and Knowledge
  Discovery. Springer, pp. 42--53.

\bibitem[{Cortes and Mohri(2004)}]{cortes2004auc}
Cortes, C., Mohri, M., 2004. Auc optimization vs. error rate minimization. In:
  Advances in Neural Information Processing Systems 16. pp. 313--320.

\bibitem[{Dem{\v{s}}ar(2006)}]{demvsar2006statistical}
Dem{\v{s}}ar, J., 2006. Statistical comparisons of classifiers over multiple
  data sets. Journal of Machine Learning Research 7, 1--30.

\bibitem[{Dinuzzo(2012)}]{Dinuzzo2012The}
Dinuzzo, F., 2012. The representer theorem for hilbert spaces: a necessary and
  sufficient condition. In: Advances in Neural Information Processing Systems
  25. pp. 189--196.

\bibitem[{Elisseeff et~al.(2001)Elisseeff, Weston,
  et~al.}]{elisseeff2001kernel}
Elisseeff, A., Weston, J., et~al., 2001. A kernel method for multi-labelled
  classification. In: Advances in Neural Information Processing Systems 14. pp.
  681--687.

\bibitem[{Fan et~al.(2008)Fan, Chang, Hsieh, Wang, and Lin}]{fan2008liblinear}
Fan, R.-E., Chang, K.-W., Hsieh, C.-J., Wang, X.-R., Lin, C.-J., 2008.
  Liblinear: A library for large linear classification. Journal of Machine
  Learning Research 9, 1871--1874.

\bibitem[{Frank and Wolfe(1956)}]{frank1956algorithm}
Frank, M., Wolfe, P., 1956. An algorithm for quadratic programming. Naval
  Research Logistics Quarterly 3~(1-2), 95--110.

\bibitem[{Friedman(1937)}]{friedman1937use}
Friedman, M., 1937. The use of ranks to avoid the assumption of normality
  implicit in the analysis of variance. Journal of the American Statistical
  Association 32~(200), 675--701.

\bibitem[{F{\"u}rnkranz et~al.(2008)F{\"u}rnkranz, H{\"u}llermeier,
  Loza~Menc{\'\i}a, and Brinker}]{furnkranz2008multilabel}
F{\"u}rnkranz, J., H{\"u}llermeier, E., Loza~Menc{\'\i}a, E., Brinker, K.,
  2008. Multilabel classification via calibrated label ranking. Machine
  Learning 73~(2), 133--153.

\bibitem[{Ghamrawi and McCallum(2005)}]{ghamrawi2005collective}
Ghamrawi, N., McCallum, A., 2005. Collective multi-label classification. In:
  Proceedings of the 14th ACM International Conference on Information and
  Knowledge Management. pp. 195--200.

\bibitem[{Golub and Van~Loan(1996)}]{golub1996matrix}
Golub, G.~H., Van~Loan, C.~F., 1996. Matrix computations (3rd ed.). Johns
  Hopkins University, Press, Baltimore, MD, USA.

\bibitem[{Gopal and Yang(2010)}]{gopal2010multilabel}
Gopal, S., Yang, Y., 2010. Multilabel classification with meta-level features.
  In: Proceedings of the 33rd international ACM SIGIR Conference on Research
  and Development in Information Retrieval. ACM, pp. 315--322.

\bibitem[{Hou et~al.(2016)Hou, Geng, and Zhang}]{hou2016multi}
Hou, P., Geng, X., Zhang, M.-L., 2016. Multi-label manifold learning. In:
  Proceedings of the 30th {AAAI} Conference on Artificial Intelligence. {AAAI}
  Press, pp. 1680--1686.

\bibitem[{Hsu et~al.(2003)Hsu, Chang, Lin, et~al.}]{hsu2003practical}
Hsu, C.-W., Chang, C.-C., Lin, C.-J., et~al., 2003. A practical guide to
  support vector classification.

\bibitem[{Huang et~al.(2015)Huang, Li, Huang, and Wu}]{huang2015learning}
Huang, J., Li, G., Huang, Q., Wu, X., 2015. Learning label specific features
  for multi-label classification. In: 2015 IEEE International Conference on
  Data Mining. pp. 181--190.

\bibitem[{Huang et~al.(2018)Huang, Li, Huang, and Wu}]{huang2018joint}
Huang, J., Li, G., Huang, Q., Wu, X., 2018. Joint feature selection and
  classification for multilabel learning. IEEE Transactions on Cybernetics
  48~(3), 876--889.

\bibitem[{Jaggi(2013)}]{jaggi2013revisiting}
Jaggi, M., 2013. Revisiting frank-wolfe: Projection-free sparse convex
  optimization. In: Proceedings of the 30th International Conference on Machine
  Learning. pp. 427--435.

\bibitem[{Ji and Ye(2009)}]{ji2009accelerated}
Ji, S., Ye, J., 2009. An accelerated gradient method for trace norm
  minimization. In: Proceedings of the 26th International Conference on Machine
  Learning. pp. 457--464.

\bibitem[{Jiang et~al.(2008)Jiang, Wang, and Zhu}]{jiang2008calibrated}
Jiang, A., Wang, C., Zhu, Y., 2008. Calibrated rank-svm for multi-label image
  categorization. In: IEEE International Joint Conference on Neural Networks.
  pp. 1450--1455.

\bibitem[{Jing et~al.(2015)Jing, Yang, Yu, and Ng}]{jing2015semi}
Jing, L., Yang, L., Yu, J., Ng, M.~K., 2015. Semi-supervised low-rank mapping
  learning for multi-label classification. In: Proceedings of the IEEE
  Conference on Computer Vision and Pattern Recognition. pp. 1483--1491.

\bibitem[{Kivinen et~al.(2002)Kivinen, Smola, and
  Williamson}]{Kivinen2002Learning}
Kivinen, J., Smola, A., Williamson, R., 2002. Learning with kernels. MIT Press.

\bibitem[{Li et~al.(2012)Li, Chen, and Li}]{li2012error}
Li, H., Chen, N., Li, L., 2012. Error analysis for matrix elastic-net
  regularization algorithms. IEEE Transactions on Neural Networks and Learning
  Systems 23~(5), 737--748.

\bibitem[{McCallum(1999)}]{mccallum1999multi}
McCallum, A., 1999. Multi-label text classification with a mixture model
  trained by em. In: AAAI’99 workshop on text learning. pp. 1--7.

\bibitem[{Nesterov(2005)}]{nesterov2005smooth}
Nesterov, Y., 2005. Smooth minimization of non-smooth functions. Mathematical
  Programming 103~(1), 127--152.

\bibitem[{Qi et~al.(2007)Qi, Hua, Rui, Tang, Mei, and
  Zhang}]{qi2007correlative}
Qi, G.-J., Hua, X.-S., Rui, Y., Tang, J., Mei, T., Zhang, H.-J., 2007.
  Correlative multi-label video annotation. In: Proceedings of the 15th
  International Conference on Multimedia. pp. 17--26.

\bibitem[{Read et~al.(2011)Read, Pfahringer, Holmes, and
  Frank}]{read2011classifier}
Read, J., Pfahringer, B., Holmes, G., Frank, E., 2011. Classifier chains for
  multi-label classification. Machine Learning 85~(3), 333--359.

\bibitem[{Schapire and Singer(2000)}]{schapire2000boostexter}
Schapire, R.~E., Singer, Y., 2000. Boostexter: A boosting-based system for text
  categorization. Machine Learning 39~(2-3), 135--168.

\bibitem[{Sch{\"o}lkopf and Smola(2002)}]{scholkopf2002learning}
Sch{\"o}lkopf, B., Smola, A.~J., 2002. Learning with kernels: support vector
  machines, regularization, optimization, and beyond. MIT Press.

\bibitem[{Shalev-Shwartz and Ben-David(2014)}]{shalev2014understanding}
Shalev-Shwartz, S., Ben-David, S., 2014. Understanding machine learning: From
  theory to algorithms. Cambridge University Press.

\bibitem[{Trohidis et~al.(2008)Trohidis, Tsoumakas, Kalliris, and
  Vlahavas}]{trohidis2008multi}
Trohidis, K., Tsoumakas, G., Kalliris, G., Vlahavas, I.~P., 2008. Multi-label
  classification of music into emotions. In: {ISMIR} 2008, 9th International
  Conference on Music Information Retrieval. Vol.~8. pp. 325--330.

\bibitem[{Tsoumakas and Katakis(2006)}]{tsoumakas2006multi}
Tsoumakas, G., Katakis, I., 2006. Multi-label classification: An overview.
  International Journal of Data Warehousing and Mining 3~(3).

\bibitem[{Tsoumakas et~al.(2009)Tsoumakas, Katakis, and
  Vlahavas}]{tsoumakas2009mining}
Tsoumakas, G., Katakis, I., Vlahavas, I., 2009. Mining multi-label data. In:
  Data mining and knowledge discovery handbook. Springer, pp. 667--685.

\bibitem[{Tsoumakas et~al.(2011{\natexlab{a}})Tsoumakas, Katakis, and
  Vlahavas}]{tsoumakas2011random}
Tsoumakas, G., Katakis, I., Vlahavas, I., 2011{\natexlab{a}}. Random
  k-labelsets for multilabel classification. IEEE Transactions on Knowledge and
  Data Engineering 23~(7), 1079--1089.

\bibitem[{Tsoumakas et~al.(2011{\natexlab{b}})Tsoumakas, Spyromitros-Xioufis,
  Vilcek, and Vlahavas}]{tsoumakas2011mulan}
Tsoumakas, G., Spyromitros-Xioufis, E., Vilcek, J., Vlahavas, I.,
  2011{\natexlab{b}}. Mulan: A java library for multi-label learning. Journal
  of Machine Learning Research 12, 2411--2414.

\bibitem[{Ueda and Saito(2002)}]{ueda2003parametric}
Ueda, N., Saito, K., 2002. Parametric mixture models for multi-labeled text.
  In: Advances in Neural Information Processing Systems 15. pp. 721--728.

\bibitem[{Wang et~al.(2009)Wang, Huang, and Ding}]{wang2009image}
Wang, H., Huang, H., Ding, C., 2009. Image annotation using multi-label
  correlated green's function. In: {IEEE} 12th International Conference on
  Computer Vision. pp. 2029--2034.

\bibitem[{Wang et~al.(2016)Wang, Yang, Mao, Huang, Huang, and Xu}]{wang2016cnn}
Wang, J., Yang, Y., Mao, J., Huang, Z., Huang, C., Xu, W., 2016. Cnn-rnn: A
  unified framework for multi-label image classification. In: Proceedings of
  the IEEE Conference on Computer Vision and Pattern Recognition. pp.
  2285--2294.

\bibitem[{Wu et~al.(2018{\natexlab{a}})Wu, Tian, and Liu}]{wu2018cost}
Wu, G., Tian, Y., Liu, D., 2018{\natexlab{a}}. Cost-sensitive multi-label
  learning with positive and negative label pairwise correlations. Neural
  Networks 108, 411--423.

\bibitem[{Wu et~al.(2018{\natexlab{b}})Wu, Tian, and Zhang}]{wu2018unified}
Wu, G., Tian, Y., Zhang, C., 2018{\natexlab{b}}. A unified framework
  implementing linear binary relevance for multi-label learning. Neurocomputing
  289, 86--100.

\bibitem[{Wu et~al.(2016)Wu, Tan, Song, Chen, and Ng}]{wu2016mlforest}
Wu, Q., Tan, M., Song, H., Chen, J., Ng, M.~K., 2016. Ml-forest: A multi-label
  tree ensemble method for multi-label classification. IEEE Transactions on
  Knowledge and Data Engineering 28~(10), 2665--2680.

\bibitem[{Wu and Zhou(2017)}]{wu2017unified}
Wu, X.-Z., Zhou, Z.-H., 2017. A unified view of multi-label performance
  measures. In: Proceedings of the 34th International Conference on Machine
  Learning. pp. 3780--3788.

\bibitem[{Xing et~al.(2018)Xing, Yu, Domeniconi, Wang, and
  Zhang}]{xing2018multi}
Xing, Y., Yu, G., Domeniconi, C., Wang, J., Zhang, Z., 2018. Multi-label
  co-training. In: Proceedings of the 27th International Joint Conference on
  Artificial Intelligence. AAAI Press, pp. 2882--2888.

\bibitem[{Xu et~al.(2016)Xu, Liu, Tao, and Xu}]{xu2016local}
Xu, C., Liu, T., Tao, D., Xu, C., 2016. Local rademacher complexity for
  multi-label learning. IEEE Transactions on Image Processing 25~(3),
  1495--1507.

\bibitem[{Xu(2012)}]{xu2012efficient}
Xu, J., 2012. An efficient multi-label support vector machine with a zero
  label. Expert Systems with Applications 39~(5), 4796--4804.

\bibitem[{Xu(2018)}]{xu2018weighted}
Xu, J., 2018. A weighted linear discriminant analysis framework for multi-label
  feature extraction. Neurocomputing 275, 107--120.

\bibitem[{Xu et~al.(2014)Xu, Wang, Shen, Wang, and Chen}]{xu2014learning}
Xu, L., Wang, Z., Shen, Z., Wang, Y., Chen, E., 2014. Learning low-rank label
  correlations for multi-label classification with missing labels. In: 2014
  IEEE International Conference on Data Mining. pp. 1067--1072.

\bibitem[{Yu et~al.(2014)Yu, Jain, Kar, and Dhillon}]{yu2014large}
Yu, H.-F., Jain, P., Kar, P., Dhillon, I., 2014. Large-scale multi-label
  learning with missing labels. In: Proceedings of the 31th International
  Conference on Machine Learning. pp. 593--601.

\bibitem[{Yu et~al.(2005)Yu, Yu, and Tresp}]{yu2005multi}
Yu, K., Yu, S., Tresp, V., 2005. Multi-label informed latent semantic indexing.
  In: Proceedings of the 28th annual international ACM SIGIR Conference on
  Research and Development in Information Retrieval. ACM, pp. 258--265.

\bibitem[{Zhang et~al.(2015)Zhang, Li, and Liu}]{zhang2015towards}
Zhang, M.-L., Li, Y.-K., Liu, X.-Y., 2015. Towards class-imbalance aware
  multi-label learning. In: Proceedings of the 24th International Joint
  Conference on Artificial Intelligence. pp. 4041--4047.

\bibitem[{Zhang and Wu(2015)}]{zhang2015lift}
Zhang, M.-L., Wu, L., 2015. Lift: Multi-label learning with label-specific
  features. IEEE Transactions on Pattern Analysis and Machine Intelligence
  37~(1), 107--120.

\bibitem[{Zhang and Zhou(2006)}]{zhang2006multilabel}
Zhang, M.-L., Zhou, Z.-H., 2006. Multilabel neural networks with applications
  to functional genomics and text categorization. IEEE Transactions on
  Knowledge and Data Engineering 18~(10), 1338--1351.

\bibitem[{Zhang and Zhou(2007)}]{zhang2007ml}
Zhang, M.-L., Zhou, Z.-H., 2007. Ml-knn: A lazy learning approach to
  multi-label learning. Pattern Recognition 40~(7), 2038--2048.

\bibitem[{Zhang and Zhou(2014)}]{zhang2014review}
Zhang, M.-L., Zhou, Z.-H., 2014. A review on multi-label learning algorithms.
  IEEE Transactions on Knowledge and Data Engineering 26~(8), 1819--1837.

\bibitem[{Zhang et~al.(2018)Zhang, Zhong, and Zhang}]{zhang2018feature}
Zhang, Q., Zhong, Y., Zhang, M., 2018. Feature-induced labeling information
  enrichment for multi-label learning. In: Proceedings of the 32nd {AAAI}
  Conference on Artificial Intelligence. {AAAI} Press, pp. 4446--4453.

\bibitem[{Zhen et~al.(2018)Zhen, Yu, He, and Li}]{zhen2018multi}
Zhen, X., Yu, M., He, X., Li, S., 2018. Multi-target regression via robust
  low-rank learning. IEEE Transactions on Pattern Analysis and Machine
  Intelligence 40~(2), 497--504.

\bibitem[{Zhou(2012)}]{zhou2012ensemble}
Zhou, Z.-H., 2012. Ensemble methods: foundations and algorithms. CRC Press.

\bibitem[{Zhu et~al.(2005)Zhu, Ji, Xu, and Gong}]{zhu2005multi}
Zhu, S., Ji, X., Xu, W., Gong, Y., 2005. Multi-labelled classification using
  maximum entropy method. In: Proceedings of the 28th annual international ACM
  SIGIR Conference on Research and Development in Information Retrieval. ACM,
  pp. 274--281.

\bibitem[{Zhu et~al.(2018)Zhu, Kwok, and Zhou}]{zhu2018multi}
Zhu, Y., Kwok, J.~T., Zhou, Z.-H., 2018. Multi-label learning with global and
  local label correlation. IEEE Transactions on Knowledge and Data Engineering
  30~(6), 1081--1094.

\end{thebibliography}
	
	\end{spacing}
	
\end{document}